\documentclass[lettersize,journal]{IEEEtran}
\usepackage{amsmath,amsfonts}
\usepackage{algorithmic}
\usepackage{algorithm}
\usepackage{array}
\usepackage{textcomp}
\usepackage{stfloats}
\usepackage{url}
\usepackage[breaklinks=true]{hyperref}
\usepackage{verbatim}
\usepackage{graphicx}
\usepackage{cite}
\usepackage{xcolor}
\usepackage[process=all]{pstool}
\usepackage{siunitx}
\usepackage{soul}
\usepackage{subcaption}
\usepackage[subpreambles=false]{standalone} %
\newtheorem{proposition}{Proposition}

\newtheorem{remark}{Remark}
\newtheorem{proof}{Proof}
\newcommand{\diff}{\mathop{}\!\mathrm{d}}
\begin{document}

\title{Controlling Deformable Objects with Non-negligible Dynamics: a Shape-Regulation Approach to End-Point Positioning}

\author{Sebastien Tiburzio$^{\dag}$, Tomás Coleman$^{\dag}$, Daniel Feliu-Talegon$^{\dag}$,  Cosimo Della Santina$^{\dag, \ddag}$%
\thanks{The research was financially supported in part by the Dutch Research Foundation (NWO) through the VENI grant ROSES 20297, as part of Project EMERGE and in part by the European Union (ERC, RIPLEY, 101165078). Views and opinions expressed are however those of the author(s) only and do not necessarily reflect those of the European Union or the European Research Council. Neither the European Union nor the granting authority can be held responsible for them. We are also very grateful for this support. We also thank the company ROCSYS for sharing with us the electric cables used in the experimental validations of our method.}
\thanks{This paper has supplementary downloadable material available at http://ieeexplore.ieee.org, provided by the authors. This includes a video showing the performance of the proposed modeling and control strategies. Additional figures and video materials relevant to the results are available at https://github.com/sebtiburzio/PAC\_model\_matlab/tree/main/appendices}
\thanks{$^{\dag} $Department of Cognitive Robotics, Delft University of Technology, Delft, The Netherlands. 
        {\footnotesize c.dellasantina@tudelft.nl}}%
\thanks{$^{\ddag} $Institute of Robotics and Mechatronics, German Aerospace Center (DLR), 82234 Oberpfaffenhofen, Germany.}%
}

\maketitle

\begin{abstract}
Model-based manipulation of deformable objects has traditionally dealt with objects while neglecting their dynamics, thus mostly focusing on very lightweight objects at steady state. At the same time, soft robotic research has made considerable strides toward general modeling and control, despite soft robots and deformable objects being very similar from a mechanical standpoint.
In this work, we leverage these recent results to develop a control-oriented, fully dynamic framework of slender deformable objects grasped at one end by a robotic manipulator. We introduce a dynamic model of this system using functional strain parameterizations and describe the manipulation challenge as a regulation control problem. This enables us to define a fully model-based control architecture, for which we can prove analytically closed-loop stability and provide sufficient conditions for steady state convergence to the desired state.
The nature of this work is intended to be markedly experimental. We provide an extensive experimental validation of the proposed ideas, tasking a robot arm with controlling the distal end of six different cables, in a given planar position and orientation in space.

\end{abstract}

\section{Introduction}\label{sec:I-introduction}

Deformable objects are ubiquitous in our daily lives, but robots still find it extremely complex to manipulate them \cite{arriola2020modeling,yin2021modeling,zhu2022challenges,fiorini2022concepts}. A central reason for this difficulty is the need for infinite degrees of freedom (DoFs) to fully describe a deformable object's state, making the direct application of established strategies unfeasible \cite{mason1985robot,bicchi1994problem,dafle2014extrinsic,pang2023global}.

This is the case also for the simplest object geometries that we will be the focus of the present work: slender or deformable linear\footnote{It is worth stressing that \textit{linear} does not refer to the dynamics or kinematics of these systems being linear, but to the fact that these objects \textit{look like} a curved line. In fact, their dynamics are highly nonlinear, as we will discuss later in the paper. To avoid confusion, we sometimes employ the less common denomination of \textit{slender object}.} objects (DLOs). These are deformable bodies whose spatial configuration can be largely specified in one dominant spatial dimension.

\begin{figure}[t]
\centerline{\includegraphics[trim={0cm 0 0cm 0cm},width=0.4\textwidth,clip]{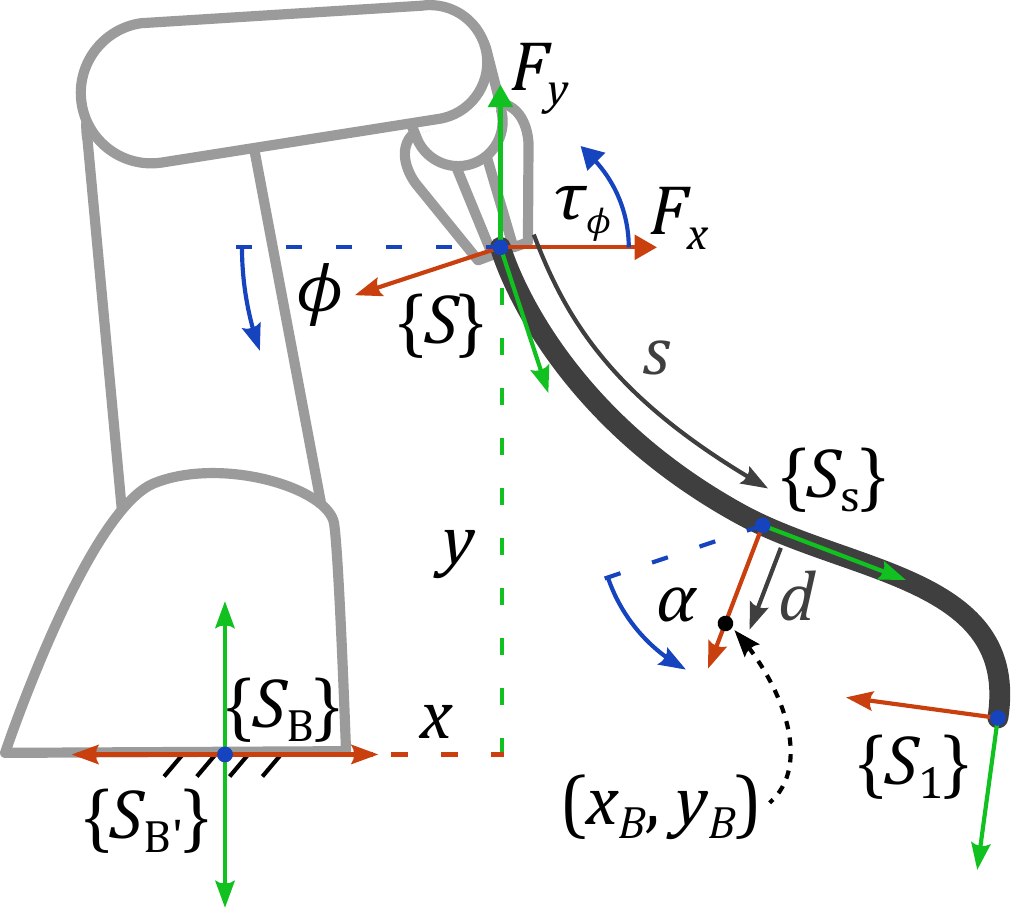}}
\caption{Pictorial representation of the task that we investigate in this work: the manipulation of slender deformable objects via a generic manipulator holding them at one of their ends by its end effector. We superimpose the important reference frames.}
\label{fig:frame_diagram}
\end{figure}
A substantial body of literature focuses on very lightweight, small, or slowly-moving DLOs (i.e., quasi-static regimes). These assumptions allow circumventing the challenge of formulating their dynamics by focusing on purely kinematic descriptions \cite{navarro2013model,zhu2018dual,lagneau2020active,zhu2021vision,shetab2022rigid}, usually relying on the deformation Jacobian paradigm and shape servoing. Interestingly, \cite{wakamatsu2004static} extends differential geometry to model linear object deformation—including flexure, torsion, and extension—for grasping and path planning. However, it remains limited to static cases, where inertia can be neglected. Machine learning-based variations of this general approach have also been considered \cite{yu2022shape,zhu2021vision}. 
Learning strategies are nowadays a common choice when it comes to relaxing the quasi-static hypothesis to move into the dynamic manipulation realm  \cite{nah_dynamic_2020,zhang_robots_2021, lin2021softgym, liu2023robotic,caporali2023weakly,kuroki2023generalizable, caporali2024deformable}. 
However, learning-based control comes with its own well-known limitations. Some of them are the lack of guarantees, the need for large amounts of data, and the potential safety concerns of the learning process.

This work aims to introduce and experimentally validate a fully control-oriented model-based perspective on manipulating slender deformable objects that does not require applying any quasi-static assumption on the object's behavior.
To achieve this goal, we put ourselves in the context of classical control of deformable mechanical systems \cite{wakamatsu2004static,lvDynamicModelingControl2022,macchelli2007port,tognon2018cooperative,do2018stabilization,mattioni2020modelling,gabellieri2023differential}. More precisely, we borrow from recent advancements in model-based control of soft robots \cite{della2023model}, which are systems entirely made of continuously deformable soft materials \cite{rus2015design,della2021soft}. 
The past few years have seen an exponential increase in control-oriented models of soft robots \cite{armanini_soft_2023,caasenbrood2024sorotoki}. These include shape\footnote{Their kinematic version have been independently investigated in deformable objects literature \cite{navarro2017fourier,qi2020adaptive,palli2020model}.} \cite{sadati2017control,wu2023towards} and - more recently - strain parametrizations \cite{grazioso2019geometrically,renda2020geometric,santina_soft_2020,stella_experimental_2022}.
In turn, the availability of finite-dimensional interpretable models has driven the flourishing of model-based control techniques \cite{george2018control,chang2020energy,franco2021adaptive,patterson2022robust,borja2022energy,renda2022geometrically,caasenbrood2022energy,stroppa2023shared,wang2023dynamic,renda2024dynamics} that combine improved performance with theoretical guarantees. 

To conclude, we propose a dynamic modeling framework for slender objects that have been pre-grasped at one of their ends by a generic robotic manipulator (see Fig. \ref{fig:frame_diagram}). The framework uses strain-based functional parameterization to model DLOs with a non-negligible dynamic response. We then cast the manipulation challenge as a regulation control problem and introduce a control architecture (see Fig. \ref{fig:block_diag}), for which we analytically prove closed-loop stability. 
Compared to recent works \cite{lv2022dynamic,tang2024learning,caporali2025robotic,aksoy2025planning} proposing control schemes within a dynamic context for single- and dual-arm DLO manipulation, our work: 1) proves closed-loop stability with explicit conditions, 2) models coupled robot-object dynamics with stability guarantees, and 3) includes gravity effects, analyzing behavior in gravity-influenced planes.

Finally, our principal goal is to showcase the possibilities that pursuing a model-based control route can yield. So, the key feature of this work is to extensively validate the proposed model in static and dynamic regimes and the control strategy when applied to object endpoint position and orientation regulation. Our setup (see Fig. \ref{fig:setup}) comprises a 7-DoF robot manipulator that grasps high-voltage electric cables normally used for charging electric vehicles.

In a broad sense, this paper can be considered the journal extension of the conference paper \cite{besselaar_one-shot_2022}. We preliminarily explored these ideas there, even if in a substantially simplified context and with different methods and goals.

\section{Strain-Based Dynamic Modeling of DLOs}
\label{sec:II-modelling}

Our goal is to introduce a fully dynamic model of a robot-object pair as described by Fig. \ref{fig:frame_diagram}. We start by introducing a dynamic model of the object, as it is the part that requires attention, and only at the end do we combine it with the model of the robot.

Compared to the model of a soft manipulator, the object part of the dynamic model introduced in this section presents the extra complexity of not having one of its ends rigidly fixed to the ground.
We can look at this as if the reference frame $\{S\}$ in Fig. \ref{fig:frame_diagram} is a floating base of the object.
We also define a fixed frame representing the manipulator base, $\{S_\mathrm{B}\}$, and the three floating base configuration variables $x \in \mathbb{R}$, $y \in \mathbb{R}$ and $\phi \in [-\pi, \pi)$. We consider the nominal state of the object to be hanging vertically downwards from its base, so it is convenient to define an intermediate frame $\{S_{B'}\}$ coincident with $\{S_\mathrm{B}\}$ but rotated $\pi$ radians in the $xy$ plane. When $x$, $y$ and $\phi$ are all 0, the object base frame $\{S\}$ is coincident with $\{S_{B'}\}$.

\subsection{Additional assumptions}
Here, we consider a few assumptions for the sake of simplicity of derivations but without loss of generality.
We considered the object inextensible and not experiencing shear strain in the plane. We also neglect to model any out-of-plane offset curvature and torsional strain, as the gravitational forces dominate in the range of end-effector orientations we validate the model in. We focus on the polynomial curvature functional parameterization \cite{della2019control}, which is representative of the general strain case discussed in \cite{renda2020geometric}.
Moreover, we propose derivations in the case of planar motions of the object, as this is the case we consider in our experimental validation, and - again - including full motions in 3D only makes derivations more complex without any substantial shift in the core ideas introduced here \cite{stella_piecewise_2022,baaij2023learning}.
Furthermore, the control algorithms that we propose later in this manuscript are agnostic to these simplifying assumptions and can be applied to the general case without modifications.

\subsection{Object's Backbone Kinematics}

We describe the DLO shape through its backbone (i.e., its central axis) as the spatial curve $(x_\mathrm{B},y_\mathrm{B},\alpha_\mathrm{B}):[0,1] \rightarrow \mathrm{SO}(2) \simeq \mathbb{R}^3$, where $s \in [0,1]$ is the local abscissa along the object. For example, $(x_\mathrm{B},y_\mathrm{B},\alpha_\mathrm{B})(0)$ represents the configuration of the grasped location, $(x_\mathrm{B},y_\mathrm{B},\alpha_\mathrm{B})(1)$ of the one of the other end, and $(x_\mathrm{B},y_\mathrm{B},\alpha_\mathrm{B})(0.5)$ the one of the mid-point.

Using standard results in differential geometry of curves \cite{do2016differential}, we express the orientation of each point of the backbone as 
\begin{equation}
    \alpha_\mathrm{B}(s) = (\pi + \phi) + \int_0^s \kappa(v) \mathrm{d}v,
\end{equation}
where $\kappa:[0,1] \rightarrow \mathbb{R}$ is the curvature strain (well-defined thanks to the Lie structure of $\mathrm{SO}(2)$), and $\phi \in S^1 \simeq [-\pi,\pi)$ is the orientation of the grasping location. 

Again, relying on the differential geometry of curves \cite{do2016differential}, the Cartesian coordinates of the central axis are evaluated by integrating components in the $x$- and $y$-directions of $\{S_v\}$ as $\alpha_\mathrm{B}$ varies up to $s$
\begin{equation}\label{eq:fk__backbone_floating}
    \begin{bmatrix}
        x_\mathrm{B}(s) \\
        y_\mathrm{B}(s)
    \end{bmatrix}
    =
    \begin{bmatrix}
        x\\ y
    \end{bmatrix}
    + 
    L \int_0^s \begin{bmatrix}
        \sin(\alpha_\mathrm{B}(v))\\-\cos(\alpha_\mathrm{B}(v))
    \end{bmatrix}\mathrm{d}v
\end{equation}
where $L$ is the total length of the object, and $x,y \in \mathbb{R}$ are two extra parameters. It is immediate to verify that $(x_\mathrm{B}, y_\mathrm{B}, \alpha_{\mathrm{B}})(0) = (x,y, \pi + \phi)$ thus giving an immediate interpretation to the three parameters as the configuration of the robot's end effector.

Thus, the complete configuration of the object can be reformulated as $(x,y,\phi,\kappa(\cdot))$. Note that, up to this point, the description is exact and infinite dimensional - as $\kappa$ is a function. We move from infinite to finite domain by approximating the curvature with a Taylor expansion
\begin{equation}\label{eq:curvature}
    \kappa(s) \simeq \sum_{i=0}^\infty \theta_i s^i,
\end{equation}
which can be truncated to degree $n$ to obtain a finite object configuration vector $\Theta \in \mathbb{R}^{n+1}$ approximating the real object to a desired precision. 
We use this to express the orientation of the central axis $\alpha$ (relative to the object base frame $\{S\}$) at a normalized distance $s \in [0,1]$ along the object's length as the integral of the curvature function
\begin{equation}
    \alpha(s, \Theta) = \int_0^s \sum_{i=0}^n \theta_i v^i \mathrm{d}v
\end{equation}

The configuration vector for the floating base model combines the curvature and manipulator's end effector variables: $q_{\mathrm{O}}=(\Theta, x, y, \phi) \in \mathbb{R}^{n+4}$.

\subsection{Object's Full Kinematics Under Cosserat Assumption}

So far, we have derived the object's kinematic model under the implicit working assumption that it is infinitely thin.
This can be readily relaxed under the Cosserat rod description \cite{altenbach2013cosserat} - i.e., we represent the object as nondeforming outside of its central backbone and assume that an undeformable slice of a point mass is rigidly connected at each point of the backbone.

Adding an offset of normalized distance $d \in [-\frac{1}{2},\frac{1}{2}]$ perpendicular to the central axis, we can then write the forward kinematics for a point on the object, scaling to its physical length $L$ and width $D$
\begin{equation}\label{eq:fk_floating}
    \begin{bmatrix}
        x_\mathrm{B}(s,d) \\
        y_\mathrm{B}(s,d)
    \end{bmatrix}
    =
    \begin{bmatrix}
        x \\
        y
    \end{bmatrix}
    +
    \begin{bmatrix}
        L\int_0^s\sin(\alpha_\mathrm{B}(s))\diff v - Dd\cos(\alpha_\mathrm{B}(s)) \\
        L\int_0^s-\cos(\alpha_\mathrm{B}(s))\diff v - Dd\sin(\alpha_\mathrm{B}(s))
    \end{bmatrix}.
\end{equation}
It is immediate to see that, for $d = 0$, this expression reduces itself to the kinematics of the backbone.

\subsection{Object's Dynamics}

Now that the finite-dimensional forward kinematics describing the complete infinite-dimensional shape of the object is defined, we can proceed to establish the dynamics using the Euler-Lagrange methodology. These steps are only briefly summarized here, as they are very similar to the ones detailed in \cite{santina_soft_2020,della2019control}. 
The inertia matrix is constructed by consideration of the kinetic energy of each infinitesimal mass element in the object:
\begin{equation}\label{eqn:B}
    B_{\mathrm{O}}(q_{\mathrm{O}}) = \int_0^1\int_{-\frac{1}{2}}^{\frac{1}{2}} \rho(s,d)J_{s,d}^TJ_{s,d}^{} \,\diff d\,\diff s.
\end{equation}
Here $\rho(s,d)$ is the mass density distribution, and $J_{s,d}$ is the Jacobian matrix of the forward kinematics function \eqref{eq:fk_floating} with respect to $q_{\mathrm{O}}$.
The Coriolis and centrifugal matrix $C$ from the inertia matrix is evaluated by standard procedures - e.g., Christoffel symbols.

The gravitational force field is derived by differentiation of the gravitational potential energy of the infinitesimal masses, with the definition:
\begin{equation}\label{eqn:G}
    G_{\mathrm{O}}(q_{\mathrm{O}}) = g \nabla_{q_{\mathrm{O}}}\int_0^1\int_{-\frac{1}{2}}^{\frac{1}{2}} \rho(s,d) y_\mathrm{B}\,\diff d\,\diff s
\end{equation}
The direction of the gravitational field is considered directed along $-y$ in the $\{S_\mathrm{B}\}$ frame.

We model the object's internal elastic and dissipative forces as discussed in \cite{della2019control}. An addition compared to that model is the relaxation of the assumption that the minimum of the elastic potential is in the straight configuration. Indeed, while this is a reasonable assumption for soft robots, it is not for deformable objects - as we will observe in the experimental validation of our results. This can be simply obtained by introducing an offset $\Bar{\Theta}$ in the elastic force term - which is the configuration in which the object would rest when not immersed in a gravitational field.

Combining all these expressions yields the dynamic equations of motion for the floating base system:
\begin{equation}\label{eqn:dynamics}\small
\begin{split}
    B_{\mathrm{O}}(q_{\mathrm{O}})\Ddot{q}_{\mathrm{O}} &+ C_{\mathrm{O}}(q_{\mathrm{O}},\Dot{q}_{\mathrm{O}})\Dot{q}_{\mathrm{O}} +G_{\mathrm{O}}(q_{\mathrm{O}}) \\ + &\begin{bmatrix}
        kH & 0_{n+1\times3}\\
        0_{3\times n+1} & 0_{3\times3} 
    \end{bmatrix}
    \begin{bmatrix}
        \Theta - \Bar{\Theta}\\
        x\\
        y\\
        \phi
    \end{bmatrix} \\
    + &\begin{bmatrix}
        \beta H & 0_{n+1\times3}\\
        0_{3\times n+1} & 0_{3\times3} 
    \end{bmatrix}
    \Dot{q}_{\mathrm{O}}
    =
    \begin{bmatrix}
        0_{n\times1}\\
        F_\mathrm{x}\\
        F_\mathrm{y}\\
        \tau_\phi
    \end{bmatrix},
\end{split}
\end{equation}
where $F_\mathrm{x}$, $F_\mathrm{y}$, and $\tau_{\phi}$ are generalized forces (two forces and torque) representing the action of the robot on the object, $H$ is the $n\times n$ Hankel matrix, and the internal stiffness and damping coefficients of the object averaged along its length are $k$ and $\beta \succ 0$.

\subsection{Complete Dynamical Model}

An object dynamics in the form expressed by \eqref{eqn:dynamics} is extremely convenient when - as for our experimental setup - a force/torque sensor collocated at the gripper location is available that can directly measure $F_\mathrm{x}, F_\mathrm{y}, \tau_\phi$. 

However, we want to be able to design controllers and assess stability when considering the complete robot-object system dynamics. First, we express the object's forward kinematics as a function of the robot configuration $q_{\mathrm{r}} \in \mathbb{R}^{n_{\mathrm{r}}}$. This is achieved by plugging $(x,y,\phi) = h_{\mathrm{r}}(q_{\mathrm{r}})$ in \eqref{eq:fk_floating} with $h_{\mathrm{r}}:\mathbb{R}^{n_{\mathrm{r}}} \rightarrow \mathrm{SO}(3)$ being the forward kinematics of the robot's end effector.

Then, the full-order dynamics is readily derived by differentiating separately w.r.t. $q_{\mathrm{r}}$ and $\Theta$ the new kinetic and potential energies so defined
\begin{equation}\label{eq:complete_dynamics}\small
\begin{split}
    &\begin{bmatrix}
        B_{\mathrm{r},\mathrm{r}}(q_{\mathrm{r}},\Theta) &B_{\mathrm{r},\Theta}(q_{\mathrm{r}},\Theta)\\
        B_{\mathrm{r},\Theta}^{T}(q_{\mathrm{r}},\Theta) &B_{\Theta, \Theta}(\Theta)
    \end{bmatrix}
    \begin{bmatrix}
        \Ddot{q}_{\mathrm{r}}\\
        \Ddot{\Theta}
    \end{bmatrix}
    + C(q_{\mathrm{r}},\Theta,\dot{q}_{\mathrm{r}},\dot{\Theta})
    \begin{bmatrix}
        \dot{q}_{\mathrm{r}}\\
        \dot{\Theta}
    \end{bmatrix} \\
    & + 
     \begin{bmatrix}
        0_{n_{\mathrm{r}}\times3} & 0_{n+1\times n_{\mathrm{r}}}\\
        0_{n_{\mathrm{r}}\times n+1} & \beta H
    \end{bmatrix}\begin{bmatrix}
        \dot{q}_{\mathrm{r}}\\
        \dot{\Theta}
    \end{bmatrix} + \begin{bmatrix}
        G_{\mathrm{r}}(q_{\mathrm{r}},\Theta)\\
        G_{\Theta}(q_{\mathrm{r}},\Theta)
    \end{bmatrix} \\
    & + 
     \begin{bmatrix}
        0_{n_{\mathrm{r}}\times3} & 0_{n+1\times n_{\mathrm{r}}}\\
        0_{n_{\mathrm{r}}\times n+1} & kH
    \end{bmatrix}\begin{bmatrix}
        {q}_{\mathrm{r}}\\
        (\Theta-\bar{\Theta})
    \end{bmatrix} = \begin{bmatrix}
        \tau \\
        0_{n+1\times 1}
    \end{bmatrix}.
\end{split}
\end{equation}
where $(q_{\mathrm{r}},\Theta) \in \mathbb{R}^{n_{\mathrm{r}} + n + 1}$ is the full order counterpart of $q_{\mathrm{O}}$ and complete configuration of the robot-object system. The other elements of the equation have their meaning in accordance with previous definitions.

Two facts about \eqref{eq:complete_dynamics} are worth stressing here. First, the object and the robot are coupled purely through the potential field and the inertia forces. No elastic or dissipative coupling between the two dynamics is present.

Second, writing the operational space dynamics of the robot from \eqref{eq:complete_dynamics} and equating it to \eqref{eqn:dynamics} allows us to prove that
\begin{equation}\label{eq:zero_dyn_trick}
    (q_{\mathrm{r}},\dot{q}_{\mathrm{r}}) \equiv 0 \Rightarrow (F_\mathrm{x}, F_\mathrm{y}, \tau_\phi) = \begin{bmatrix}
        0_{3 \times n +1} & I_{3 \times 3}
    \end{bmatrix}G_{\mathrm{O}}.
\end{equation}

\begin{figure*}[t]
\centerline{\includegraphics[trim={0cm 0 0cm 0cm},width=\textwidth,clip]{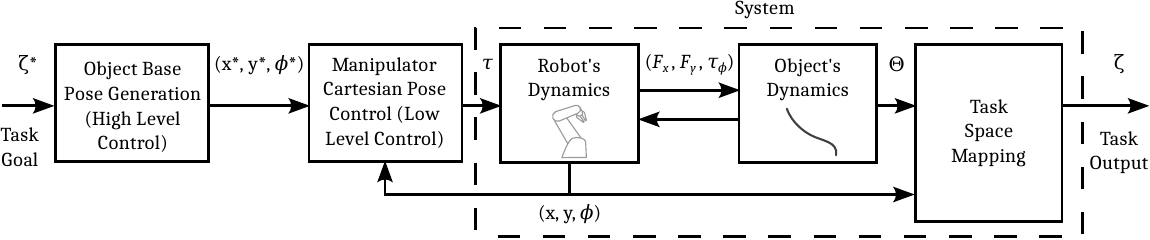}}
\caption{A block diagram outline of the control strategy. The input is a task goal $\zeta^*$ which can be achieved by regulation of the object's shape. We use a nonlinear model-constrained optimization to map this to a Cartesian object base pose $(x^*, y^*, \phi^*)$, providing a control input to the coupled manipulator-object system. }
\label{fig:block_diag}
\end{figure*}

\section{Manipulation as Closed-loop Control}

The general control architecture proposed in this work is presented in Fig. \ref{fig:block_diag}. In the following, we present its main components step by step. In the process, we discuss the stability and convergence properties of a broad class of controllers that can be applied to this setting.

\subsection{Task Goal}
Thanks to the effort to establish a complete model of the robot-object pair in Sec. \ref{sec:II-modelling}, we can now define our manipulation goal as a shape regulation task. More precisely, we wish to define a policy that specifies the control action $\tau$ in \eqref{eq:complete_dynamics} such that
\begin{equation}\label{eq:control_goal}
    \lim_{t \rightarrow \infty} \zeta(x,y,\phi,\Theta) = \zeta^*.
\end{equation}
Here, $\zeta:\mathbb{R}^{n+4} \rightarrow \mathbb{R}^{m}$ is a generic representation of a object-centric task. In other words, the configuration of the robot affects the fulfillment of the task only to the extent to which a change of $q_{\mathrm{r}}$ reflects a change in the robot's state.
This can, for example, be the position and orientation of one or more points along the object structure (we consider this case in the experimental validation), their relative position to an external target (e.g., insertion task), or directly the object shape $\Theta$. 

\subsection{Low-level Control}

At this stage, we use the control inputs $\tau$ to regulate the collocated configuration variables $q_{\mathrm{r}}$. Our goal for this section is to analyze the closed-loop control problem and propose a simple but provably stable control strategy. This layer will then be used by the high-level controller to ultimately solve the control goal stated above.

\subsubsection{Stability of the zero dynamics}
Combining \eqref{eqn:dynamics} and \eqref{eq:zero_dyn_trick} yields the zero dynamics
    \begin{equation}\label{eq:zd}
        \begin{split}
            B_{\Theta, \Theta}(\Theta)\Ddot{\Theta}
            + C_{\Theta, \Theta}(\Theta,\dot{\Theta}) \dot{\Theta} + \beta H
                \dot{\Theta} &+  G_{\Theta}(\Theta,\phi^*) \\ &+ kH{(\Theta-\bar{\Theta})} =  0.
        \end{split}
    \end{equation}
    Where $\phi^*$ is the end effector orientation associated with the fixed robot configuration $q^*_{\mathrm{r}}$.
    This equation describes the behavior of the object when the robot is fixed in some configuration, and analyzing its stability is a fundamental step in designing a robot-side controller and assessing its performance. We do this in the following proposition.
\begin{proposition} \label{pr:zd}
    The state $(\Theta^*, 0) \in \mathbb{R}^{2n}$ is an asymptotically stable equilibrium of (\ref{eq:zd}) if an open neighborhood $\mathcal{N}(\Theta)$ $\subseteq \mathbb{R}^{2}$ of $\Theta^*$  exists such that $\forall \Theta \in \mathcal{N}(\Theta^*)/\{\Theta^*\}$ 
       \begin{equation}\label{eq:Condit}
    \begin{split}
        (U_{G_{\Theta}}(\Theta,\phi^*)+ & U_{K}(\Theta))  > (U_{G_{\Theta}}(\Theta^*,\phi^*)+U_{K}(\Theta^*)) \\
        +  & \left(\frac{\partial}{\partial \Theta}(U_{G_{\Theta}}(\Theta,\phi^*)+U_{K}(\Theta)) \right)\bigg{|}^{T}_{\Theta = \Theta^*} (\Theta - \Theta^*)
        \end{split}
    \end{equation} 
    \noindent where $U_{G_{\Theta}}$ and $U_{K}$ represent the potential energy associated with gravity and elasticity of the DLO, respectively.
    \end{proposition}

\begin{proof}
    Taking inspiration from the soft robotics literature, we propose to use a modified version of the Lyapunov function in \cite[Theorem 1]{della2023model}, which is
           \begin{equation}
    \begin{split}
        V = &  \frac{1}{2}\dot{\Theta}^{T}B_{\Theta, \Theta}\dot{\Theta} + U_{G_{\Theta}}(\Theta,\phi^*) - U_{G_{\Theta}}(\Theta^*,\phi^*) + U_{K}(\Theta) \\
         - & U_{K}(\Theta^*) +  \left(kH{(\Theta^*-\bar{\Theta})}+G_{\Theta}(\Theta^*,\phi^*) \right) (\Theta^* - \Theta).
        \end{split}
    \end{equation} 
    Taking into account condition (\ref{eq:Condit}) and the fact that the kinetic energy is always strictly positive definite in $\dot{\Theta}$, it follows that $V$ is positive definite. Next, we analyze the sign of its time derivative, which is
    \begin{equation}\label{ec:deriv_V}
    \begin{split}
        \dot{V} = & \frac{1}{2}\dot{\Theta}^{T}\dot{B}_{\Theta, \Theta}\dot{\Theta}+ \dot{\Theta}^{T}B_{\Theta, \Theta}\Ddot{\Theta} + \dot{\Theta}^{T}(kH{(\Theta-\bar{\Theta})}+G_{\Theta}(\Theta,\phi^*)) \\
        -  & \dot{\Theta}^{T}(kH{(\Theta^*-\bar{\Theta})}+G_{\Theta}(\Theta^*,\phi^*)) \\ 
        = & \frac{1}{2}\dot{\Theta}^{T}\dot{B}_{\Theta, \Theta}\dot{\Theta} + \dot{\Theta}^{T}(kH{(\Theta-\bar{\Theta})}+G_{\Theta}(\Theta,\phi^*)) \\   
        & +  \dot{\Theta}^{T} \left( -C_{\Theta, \Theta}\dot{\Theta} - kH{(\Theta-\bar{\Theta})}-G_{\Theta}(\Theta,\phi^*) - \beta H \dot{\Theta}\right) \\
         =  & -\dot{\Theta}^{T}\beta H\dot{\Theta} \leq 0,
        \end{split}
    \end{equation} 
    \noindent where we leverage the system's passivity, as expressed by $\dot{B}_{\Theta, \Theta}-2 C_{\Theta, \Theta} = 0$, and the equilibrium condition $kH{(\Theta^*-\bar{\Theta})}+G_{\Theta}(\Theta^*,\phi^*)=0$. Consequently, equation (\ref{ec:deriv_V}) is shown to be negative semidefinite, but thanks to LaSalle's principle, we can guarantee that the system converges to the set $(\Theta^*, 0)$, provided that the equilibrium configuration $\Theta^*$ is the only configuration in $\mathcal{N}(\Theta^*)$ satisfying $\Ddot{\Theta} = 0$ for $\dot{\Theta} = 0$.  
    \hfill $\square$
\end{proof}

\vspace{0.2cm}

    Then, the collocated zero dynamics (\ref{eq:zd}) is such that
    \begin{equation}
        \lim_{t \rightarrow \infty} \dot{\Theta} = 0
    \end{equation}
    if $\beta > 0$.

\subsubsection{Control strategy}

The stability of the zero dynamics enables direct feedback linearization to regulate the robot's states. However, we prefer here a simpler regulator that extends controllers commonly used in industrial robots.
\begin{equation}\label{eq:control_ll}
    \tau = K_{\mathrm{P}} (q^*_{\mathrm{r}} - q_{\mathrm{r}}) + K_{\mathrm{D}} (-\dot{q}_{\mathrm{r}}) + G_{\mathrm{r}}(q_{\mathrm{r}},\Theta).
\end{equation}
where $K_{\mathrm{P}},K_{\mathrm{D}}\succ 0$ are two positive definite control gain matrices. The only condition necessary for the closed loop stability is that $\beta > 0$. 

{
\begin{proposition} 
    There always exist a large-enough proportional gain $K_{\mathrm{P}}$ such that the trajectories of the closed-loop system (\ref{eq:control_ll}), (\ref{eq:complete_dynamics}) are bounded and converge asymptotically to the equilibrium state $(q_{r}^{*}, \Theta^*, 0, 0)$,  where $\Theta^*$ is a solution of
\begin{equation}\label{eq:ss_zd}
     G_{\Theta}(\Theta^*,\phi^*) + kH{(\Theta^*-\bar{\Theta})} = 0.
\end{equation}
\end{proposition}

\begin{proof}
    The proof follows directly from Proposition \ref{pr:zd}, following the same steps of the proof of Theorem 1 in the soft robotics control paper \cite{pustina2022feedback}. 
    \hfill $\square$
\end{proof}

Furthermore, we can provide even stronger convergence properties for the closed loop under slightly stronger hypotheses, as discussed in the following proposition.

\begin{proposition} 
The closed loop (\ref{eq:control_ll}), (\ref{eq:complete_dynamics}) converges asymptotically to the unique solution of \eqref{eq:ss_zd} in the region of attraction $\mathcal{N}$, if
\begin{equation}
        \left( kH + \frac{\partial G_{\Theta}(\Theta,\phi^{*})}{\partial \Theta} \right) \succ 0, \quad \forall \Theta \in \mathcal{N}.
       \label{eq:elastic_dominance}
  \end{equation} 
  \end{proposition}

\begin{proof}

    The statement has two parts. The first one is that the solution in $\mathcal{N}$ is unique. This follows by the application of Hadamard's global inverse function theorem since the left-hand side of \eqref{eq:elastic_dominance} is the Jacobian of the left-hand side of \eqref{eq:ss_zd}. Also, the left-hand side of \eqref{eq:elastic_dominance} is proper because it is radially unbounded as \( \|\Theta\| \to \infty \), the linear term \( kH(\Theta - \bar{\Theta}) \) dominates because it grows without bound. This allows to state the property globally if $\mathcal{N}$ coincides with the whole configuration space.

    The second part is about asymptotic convergence. This one can be proven by relying on a variation of the Lyapunov candidate that we proposed in the proof of Proposition \ref{pr:zd}
    \begin{equation}
    \begin{split}
        V & =  \frac{1}{2}
        \begin{bmatrix}
        \dot{q}_{\mathrm{r}}\\
        \dot{\Theta}
    \end{bmatrix}^{\top}
\begin{bmatrix}
        B_{\mathrm{r},\mathrm{r}}(q_{\mathrm{r}},\Theta) &B_{\mathrm{r},\Theta}(q_{\mathrm{r}},\Theta)\\
        B_{\mathrm{r},\Theta}^{T}(q_{\mathrm{r}},\Theta) &B_{\Theta, \Theta}(\Theta)
    \end{bmatrix}
    \begin{bmatrix}
        \dot{q}_{\mathrm{r}}\\
        \dot{\Theta}
    \end{bmatrix} 
     \\ & + U_{G_{\Theta}}(\Theta,\phi) - U_{G_{\Theta}}(\Theta^*,\phi^*) + U_{K}(\Theta) - U_{K}(\Theta^*) \\ & +  \left(kH{(\Theta^*-\bar{\Theta})}+G_{\Theta}(\Theta^*,\phi^*) \right) (\Theta^* - \Theta) \\ 
     & + \frac{1}{2} (q^{*}_r-q_{r})^T K_{p}(q^{*}_r-q_{r}).
    \end{split}
    \end{equation}
    The proof then follows similar steps, with the only addition of noticing that \eqref{eq:elastic_dominance} implies \eqref{eq:Condit}.
    \hfill $\square$
\end{proof}

In the soft robotics literature, \eqref{eq:elastic_dominance} is called elastic dominance, and its goal is to ensure that the physical compliance of the robot (or, in this case, of the object) is high-enough when compared to gravity.

\begin{remark} 
The control strategy (\ref{eq:control_ll}) simplifies to the standard PD control with gravity compensation under the hypothesis that the mass of the
object is significantly smaller than the robot’s mass. This becomes evident when considering that the gravity terms associated with the robot are the sum of the contributions from the robot's own mass and the effect of the DLO on the robot, expressed as $G_{\mathrm{r}}(q_{\mathrm{r}},\Theta) = G_{\mathrm{rr}}(q_{\mathrm{r}})+G_{\mathrm{or}}(q_{\mathrm{r}},\Theta)$. If we assume that the mass of the object is significantly smaller than the robot's mass, i.e., $||B_{\Theta,\Theta}|| << ||B_{\mathrm{r,r}}||$, it follows that $G_{\mathrm{r}}(q_{\mathrm{r}},\Theta) \simeq G_{\mathrm{rr}}(q_{\mathrm{r}})$. This can be further formulated as:     
    ${\partial G_{\mathrm{r}}}/{\partial \Theta} \simeq 0$,
which is commonly observed in practice, as it reflects the typical scenario where the object's mass is negligible compared to the robot's mass.
\end{remark}

\subsection{High-level Control}

Employing the low-level controller \eqref{eq:control_ll} allows to freely specify a configuration $q^*_{\mathrm{r}}$ for the robot. Combining this capability with standard Jacobian-based Cartesian controllers, we can use this capability to freely specify a configuration of the robot's gripper $(x^*,y^*,\phi^*)$.

To avoid destabilizing the low-level control loop, we cast the high-level component as pure feedforward, as outlined in the block diagram of Fig. \ref{fig:block_diag}. We propose a model-constrained nonlinear optimization as an object pose generation algorithm. We use the manipulator's grasping point to control the equilibrium shape $\Theta^*$ of the object via the steady state zero dynamics equation \eqref{eq:ss_zd}. Note that we can do so because of the pure dependency of $G_{\Theta}$ on $(x^*,y^*,\phi^*)$ rather than on the complete configuration $q^*_{\mathrm{r}}$.

Our desired object configuration is one such that the control goal \eqref{eq:control_goal} is verified. To find such a configuration, we define a cost based on a distance measure between desired and modeled positions and orientations, $d(\zeta^*,\zeta)$. 

Thus, the optimization problem can then be written as:
\begin{equation}\label{eqn:general_NLO}
\begin{split}
    &\min_{\Theta^* \in \mathbb{R}^{n+1},(x^*,y^*,\phi^*)\in\mathbb{F}}
    \quad d(\zeta^*,\zeta(x^*,y^*,\phi^*,\Theta^*)) \\
    &\mathrm{s.t.} \quad G_{\Theta}(\Theta^*,\phi^*) + kH{(\Theta^*-\bar{\Theta})} = 0.
\end{split}
\end{equation}
Here, we also impose a feasible set $\mathbb{F}$ on the floating base coordinates to represent the constrained reachable workspace of the manipulator - thus avoiding the risk of generating unfeasible references to the Cartesian control. This set can also be used to represent environmental constraints.

By solving \eqref{eqn:general_NLO}, we retrieve a base configuration $(x^*,y^*,\phi^*)$ minimizing the difference between the desired and modeled object shape, thus completing the control architecture in Fig. \ref{fig:block_diag}.

\section{Experimental Validation of the Model}\label{sec:III-model-validation}

\subsection{Experimental Setup}

\subsubsection{Test Objects}

The DLOs used in our experimental investigations are heavy gauge electrical cables, chosen particularly due to their significant inertia and moderate stiffness. Different lengths, gauges (diameter and material composition) and the weight of the endpoint mass provide variation in the object properties leading to a range of different dynamic and equilibrium behaviours. Objects OB1-OB4 are made from the same, thicker cable material, and have variations of both a short or long length, and a significant or minimal endpoint mass. Objects OB5 and OB6 are made from a more flexible material, and only a long, weighted variation and a short, unweighted variation are considered. The full set of objects is pictured in Fig. \ref{fig:objs_photo}.

\begin{table}[t]
\caption{Object Parameter Variations}
\begin{center}
\begin{tabular}{ccccc}
\hline
ID & $m_{\mathrm{L}} [kg]$ & $m_1 [kg]$ & L [m]    & D [m]     \\ \hline
OB1  & 0.6   & 0.23  & 0.6  & 0.02  \\
OB2  & 0.6   & 0.03  & 0.6  & 0.02  \\
OB3  & 0.42  & 0.23  & 0.4  & 0.02  \\
OB4  & 0.42  & 0.03  & 0.4  & 0.02  \\
OB5  & 0.4   & 0.23  & 0.75 & 0.015 \\
OB6  & 0.25  & 0.03  & 0.45 & 0.015 \\
\hline
\end{tabular}
\label{tab:objs}
\end{center}
\end{table}

\begin{figure}[t]
\centerline{\includegraphics[width=0.4\textwidth]{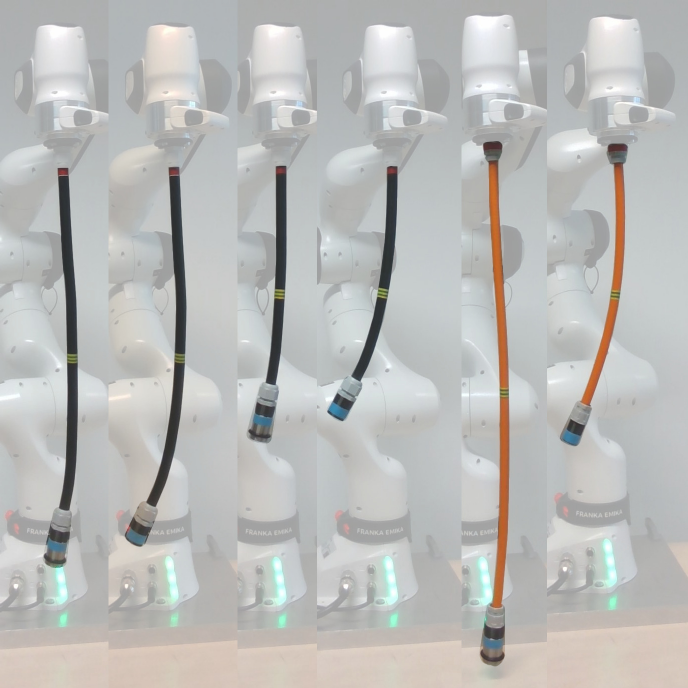}}
\caption{The test objects used in the experiments, OB1-OB6 from left to right. These are high-voltage electric cables used in the context of the electric car industry. The cables end with a plug connector. }
\label{fig:objs_photo}
\end{figure}

The objects considered here are assumed to have constant density and cross-section along their length; however, instead of using the exact continuous distribution of $\rho(s,d)$ when evaluating \eqref{eqn:B} and \eqref{eqn:G}, we simplify using an approximation of six discrete lump masses. These are equally spaced along the length of the object at positions $s = \{\frac{1}{12},\frac{3}{12},\frac{5}{12},\frac{7}{12},\frac{9}{12},\frac{11}{12}\}$, each with a mass of $\frac{m_{\mathrm{L}}}{6}$ distributed across the total diameter. Different masses $m_0$ and $m_1$ are also included at $s=0$ and $s=1$, to account for separate masses concentrated at the base and end of the object. In terms of unmodeled non-idealities, approximately 0.05-0.1m at the end of the objects is rigid to facilitate the attachment of the endpoint mass. Table \ref{tab:objs} summarises the properties of the objects. 

\subsubsection{Test Equipment}

A Franka Emika FR3 7-DoF manipulator arm was used to provide actuation of the object's base in the experiments. When required, a Bota Systems SenseOne 6-axis force-torque sensor was assembled in between to measure the actuation applied. An Intel Realsense D435 camera was used to capture RGB images at 30Hz, from which the object state could be extracted. All hardware control and communication were managed through a PC running Ubuntu 20.04 and the ROS Noetic middleware. Fig. \ref{fig:setup} shows an overview of the test equipment setup.

\begin{figure}[t]
    \centering
    \includegraphics[trim = {100 0 0 0}, clip, width = \columnwidth]{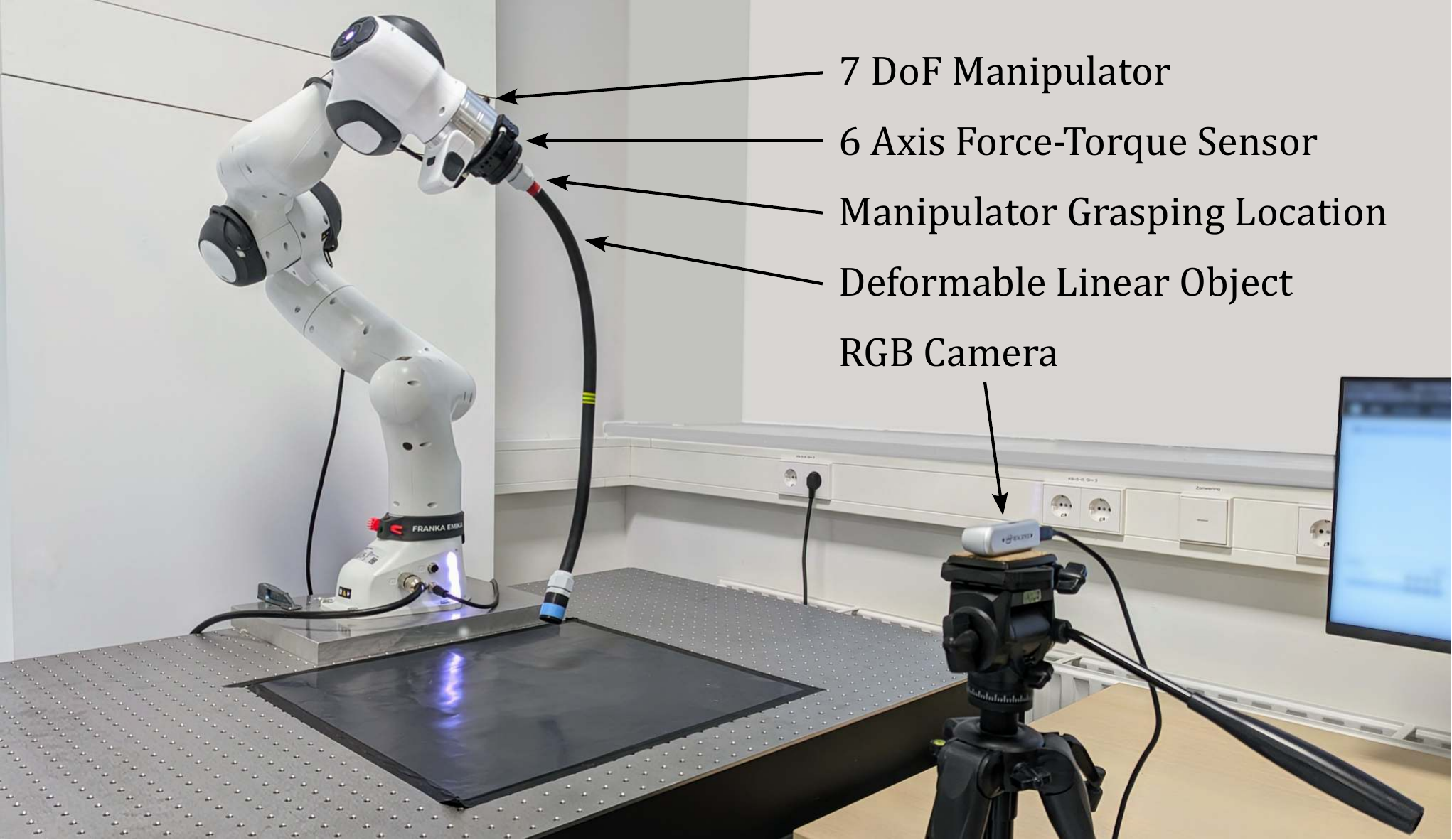}
    \caption{The experimental setup used for validation of the proposed model and control architecture. Relevant components are highlighted. Note that for simplicity and reproducibility, we are connecting the object directly to the robot's end effector instead of grasping it via a gripper.}
    \label{fig:setup}
\end{figure}

\subsection{Extraction of the Object State}\label{sec:III-Extraction}

Points on the object at $s=\{0,0.5,1\}$ and $d=0$ were identified with coloured markers, facilitating the detection of their pixel locations via HSV thresholding. Under the assumption that the object stays confined to the plane on which the manipulator endpoint moves, the 3D marker positions were determined by intersecting this plane with the line projected through the associated pixel coordinate. The positions of the base, mid-point, and endpoint of the DLO are extracted for two purposes: to estimate the object's curvature and to present results and validate our approach at different points along the DLO.

\begin{figure}[b]
    \begin{center}
        \includegraphics[width=0.45\textwidth]{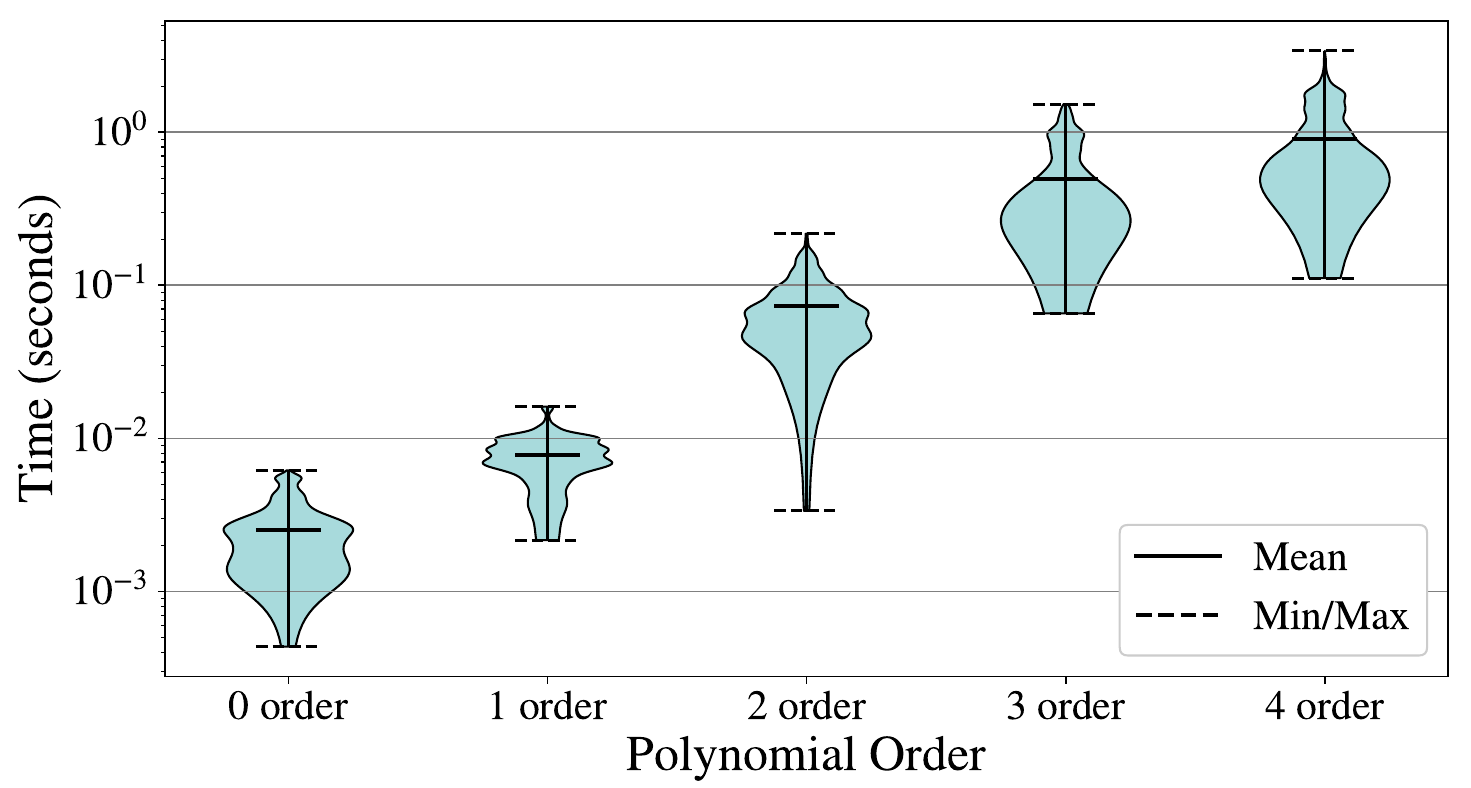}
    \end{center}
    \caption{Distributions of computation times for the inverse kinematics algorithm \ref{alg:num_IK}, for various order of approximation. The assumed DoFs for the object are equal to the polynomial order plus one. The time taken to converge to a solution was calculated for 1000 values randomly selected throughout the $\theta$-space.}
    \label{fig:ik_benchmark}
\end{figure}

By taking the recovered 3D marker positions and discarding the third coordinate (in the direction normal to the assumed object plane) we are left with the planar model 2D Cartesian coordinates of the start-, mid- and endpoints of the central axis in $\{S_B\}$: $p_{s,B}$, $p_{m,B}$ and $p_{e,B}$. Note that hereafter we will omit the subscript $B$, so that any Cartesian coordinates mentioned can be assumed measured in $\{S_B\}$ unless indicated otherwise. 

The floating base system configuration best matching the observed points on the object can be extracted via the algorithm \ref{alg:num_IK}. The algorithm has the form of a CLIK (closed-loop inverse kinematics) strategy with an over-constrained task space, as the number of target outputs (measurements of the current shape) is higher than the configurations. There, $q_{\mathrm{O},init}$ is an initial guess, $p$ are the measured coordinates of the three marked points, and $h(q_{\mathrm{O}})$ and $J(q_{\mathrm{O}})^+$ evaluate the forward kinematics and Jacobian pseudoinverse respectively at the same locations. Note that the pseudoinverse is here to be intended as a minimum MSE solution. Finally,  $\epsilon$ and $\Delta$ are tuned to a desired accuracy and convergence rate.
The calculation time of the Inverse Kinematics algorithm, which was outlined in Algorithm \ref{alg:num_IK} was tested using the parameters identified for OB1. The tests were conducted using polynomial models of order 0 to 4, with an initial guess of $\theta = 0$. The input values of $p_{\mathrm{m}}$ and $p_{\mathrm{e}}$ were taken from the DLO and generated over the range $-\pi \geq \theta_i \geq \pi$. The results of the benchmarked computation times for this algorithm are shown in Fig. \ref{fig:ik_benchmark}. The tests were run on an Intel 12th Gen i7-12700H (20) @ 4.600GHz CPU. These results demonstrate that the calculation can feasibly be integrated into a real-time control system, with both accuracy and computation time adjustable by selecting the appropriate polynomial order to suit specific application requirements.

\begin{algorithm}
\caption{Iterative Estimation of the Shape (CLIK-like)}
\label{alg:num_IK}
\begin{algorithmic}
\STATE $q_{\mathrm{O}} \leftarrow q_{\mathrm{O},init}$
\WHILE{$i \leq i_{max}$}
    \STATE $e = \|p-h(q_{\mathrm{O}})\|_2$
    \IF{$e < \epsilon$}
        \RETURN $q_{\mathrm{O}}$
        \ELSE
        \STATE $q_{\mathrm{O}} \leftarrow q_{\mathrm{O}} - \Delta J(q_{\mathrm{O}})^+e$
    \ENDIF
\ENDWHILE
\end{algorithmic}
\end{algorithm}

\subsection{Parameter Identification}\label{subsec:param_id}

The model contains several object-specific parameters, some of which are not easy to directly measure. Measurable parameters are the object length $L$, diameter $D$, total mass of the object body $m_{\mathrm{L}}$, object base mass $m_0$, and endpoint mass $m_1$. Immeasurable parameters are the object stiffness $k$, object damping $\beta$, and curvature offset $\Bar{\Theta}$. To determine the unknown parameters, an identification procedure was developed in two parts, separating out those that could be determined from the static case and those that require dynamic analysis.

For the static case, the equilibrium state of the object is extracted at a series of discrete base orientation angles $\phi$. This data is used to construct a set of linear equations, from which a pseudoinverse solution can be used to determine the values for $k$ and $\Bar{\Theta}$, providing a least squares best fit. 
Equation \eqref{eqn:param_ID_static} shows the development of the zero dynamics in \eqref{eq:ss_zd} into the form used to obtain the pseudoinverse solution. This represents a system of $n+1$ equations corresponding to a single $\phi$, in practice this is extended with additional rows for each $\phi$ and object state $\Theta^*$ measured, before the pseudoinversion step.

\begin{equation}\label{eqn:param_ID_static}
\begin{split}
    G_{\Theta}(\Theta^*,\phi^*) + kH{(\Theta^*-\bar{\Theta})} &= 0 \\
    \begin{bmatrix}H\Theta^* & -H\end{bmatrix} \begin{bmatrix} k\\ k\Bar{\Theta} \end{bmatrix} &= -G_{\Theta}(\Theta^*,\phi^*) \\
    \begin{bmatrix} k\\ k\Bar{\Theta} \end{bmatrix} &= \begin{bmatrix}H\Theta^* & -H\end{bmatrix}^+(-G_{\Theta}(\Theta^*,\phi^*))
\end{split}
\end{equation}

With the values of $k$ and $\Bar{\Theta}$ estimated as outlined above, identification of the final unknown parameter $\beta$ can subsequently be done using data taken from a dynamic evolution of the object. The approach is similar, using the zero dynamics \eqref{eq:zd} to construct an overdetermined system of equations and calculate a pseudoinverse solution as shown in \eqref{eqn:paramID_dynamic}.

\begin{equation}\label{eqn:paramID_dynamic}
\begin{split}
\beta = (H \dot{\Theta})^+ (-B_{\Theta, \Theta}(\Theta)\Ddot{\Theta}
            - C_{\Theta, \Theta}(\Theta,\dot{\Theta}) \dot{\Theta} 
            &- G_{\Theta}(\Theta,\phi^*)\\ 
            &- kH(\Theta-\bar{\Theta})).
\end{split}
\end{equation}

\subsection{Comparative analysis of polynomial curvature across different orders}\label{Comparative_study}

In the proposed approach, the configuration of the slender object is parametrized using polynomial curvature functions (\ref{eq:curvature}). While this
representation is exact in the infinite-dimensional case, it can be truncated to a polynomial of degree $n$ to yield a finite-dimensional object
configuration suitable for real-world applications. In this section, we compare the use of constant, linear, and quadratic curvature models for describing the shape of a real object. For this comparative analysis, we use the object OB1 and identify the equivalent object parameters for the three cases. %
With a first-order curvature model, the object configuration can be uniquely identified from two points (Section \ref{subsec:param_id}). For higher-order polynomials, multiple solutions exist, but only one satisfies the equilibrium. We recover parameters by minimizing the error between measured and predicted marker positions over 23 configurations spanning base orientations in [$-\pi$, $\pi$], following strategies similar to \cite{feliu2024dynamic}.

The resulting parameters for the three cases are as follows: 1) $k = 0.1726$ and $\bar{\Theta}=-0.0005$, 2) $k = 0.197$ and $\bar{\Theta}=(0.142, 1.24)$ and 3) $k = 0.1695$ and $\bar{\Theta}=(0.1081, 1.5017, 0.0022)$. Fig. \ref{fig:Comparative analysis} presents the results obtained after parameter identification for the three cases. The lower part of Fig. \ref{fig:Comparative analysis} shows the errors at the middle point ($\Delta p_{m}$) and the end-effector of the object ($\Delta p_{e}$), while the upper part illustrates the object’s shape for some configurations. The average errors with the standard deviations at the middle point for the three cases are  $11 \pm 5.5$ cm, $3.5 \pm 1.6$ cm and $1.2 \pm 0.4$ cm, while at the endpoint they are $11 \pm 6.7$ cm, $2.9 \pm 1.5$ cm and $1.5 \pm 0.5$ cm, respectively. These results effectively demonstrate that increasing the order of the polynomial significantly improves accuracy, particularly in capturing the object's middle point. However, this improvement comes at the cost of increasing the polynomial order, which, in turn, increases the dimensions of the matrices describing the system.

\begin{figure}[t]
\centerline{\includegraphics[width=0.5\textwidth]{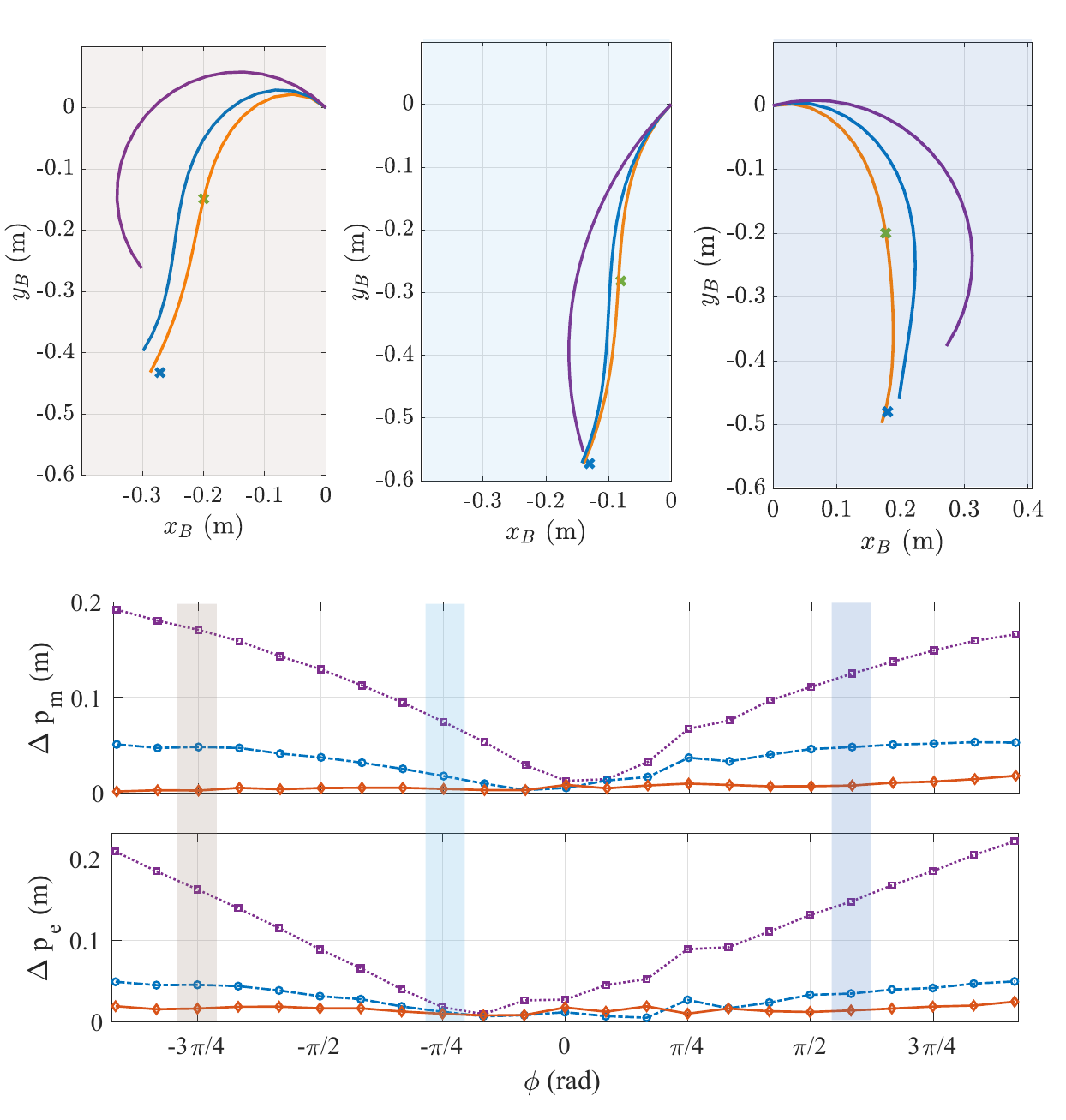}}
\caption{Comparative analysis of polynomial curvature across different orders. The results are obtained using constant (purple line), linear (blue line), and quadratic curvature (orange line). In the upper part, we show the shapes for three different orientations with colored backgrounds, which correspond to the colored regions in the figure below.}
\label{fig:Comparative analysis}
\end{figure}

\subsection{Model Validation Results}

A polynomial of degree $n=1$ was used for the experimental validation: an affine curvature strain parametrization \cite{della2020soft} of the object shape, where $\Theta = (\theta_0, \theta_1)$. The parameter identification procedures laid out in Section \ref{subsec:param_id} were carried out using this model. The ultimately determined values are summarised in Table \ref{tab:params}, while the rest of this subsection will detail the practical considerations and intermediate results.

\subsubsection{Steady State}\label{subsec:results_param_id_static}

Static equilibrium datasets were collected for each of the objects at 23 orientations of $\phi$, angled at evenly spaced $\frac{\pi}{12}$ rad increments excluding $-\pi$. To minimize the influence of hysteresis on the object's shape between measurements, they were returned to $\phi=0$ and shaken loose before moving to each measurement orientation. Configurations of $\Theta$ corresponding to the measured $p_s$, $p_{\mathrm{m}}$ and $p_{\mathrm{e}}$ were then extracted to use as input to the parameter identification formulation for $k$ and $\Bar{\Theta}$ as described in Section \ref{subsec:param_id}. Note that the affine curvature model exhibits parameter-dependent multi-stability \cite{trumic_stability_2023}. Here, it is generally safe to assume that the object will settle to an expected equilibrium if it begins in a nominal static state at $\phi=0$ and $\Theta$ close to $(0,0)$, and is rotated slowly within $-\pi \leq \phi < \pi$. 

\begin{table}[H]
\caption{Identified Object Parameters}
\begin{center}
\resizebox{6.5cm}{!} 
{ 
\begin{tabular}{ccccc}
\hline
ID & $k$ & $\bar{\theta}_0$ & $\bar{\theta}_1$ & $\beta$ \\ \hline
OB1  & 0.197 & 0.142  & 1.24 & 0.0347 \\
OB2  & 0.226 & -0.0358 & 2.06 & 0.0311 \\
OB3  & 0.253 & 0.830  & -0.452 & 0.0359 \\
OB4  & 0.360 & 0.748  & 0.0711 & 0.0454 \\
OB5  & 0.0761 & 0.354  & 0.636 & 0.0343 \\
OB6  & 0.269 & 0.479  & 0.621 & 0.0397 \\
\hline
\end{tabular}
\label{tab:params}}
\end{center}
\end{table}

Following identification for the static case, we check how well the resulting model predicts the equilibrium behaviour of the real objects. Fig. \ref{fig:static_id_composite_and_assessment}a shows the results of this process for OB1, plotted for $\phi=\{0,\pm\frac{\pi}{4},\pm\frac{\pi}{2},\pm\frac{3\pi}{4}\}$ (7 of the 23 datapoints used). Here, the coloured crosses indicate the measured positions of $p_{\mathrm{m}}$ and $p_{\mathrm{e}}$, and the light orange curves represent the configurations of $\Theta$ from inverse kinematics, used in \eqref{eqn:param_ID_static} to identify the parameters. The dark orange curves are the steady state of the model obtained by forward simulation using the identified parameters. A composite image of the actual object shapes at these angles is shown in Fig. \ref{fig:static_id_composite_and_assessment}b. Equivalent figures for the other objects are not included for the sake of space, but are available in the extra appendix materials. 

\begin{figure}[b]
\centering
    \begin{subfigure}{0.635\columnwidth}
        \includegraphics[trim=8.8cm 0cm 10cm 1cm,width=\textwidth,clip=true]{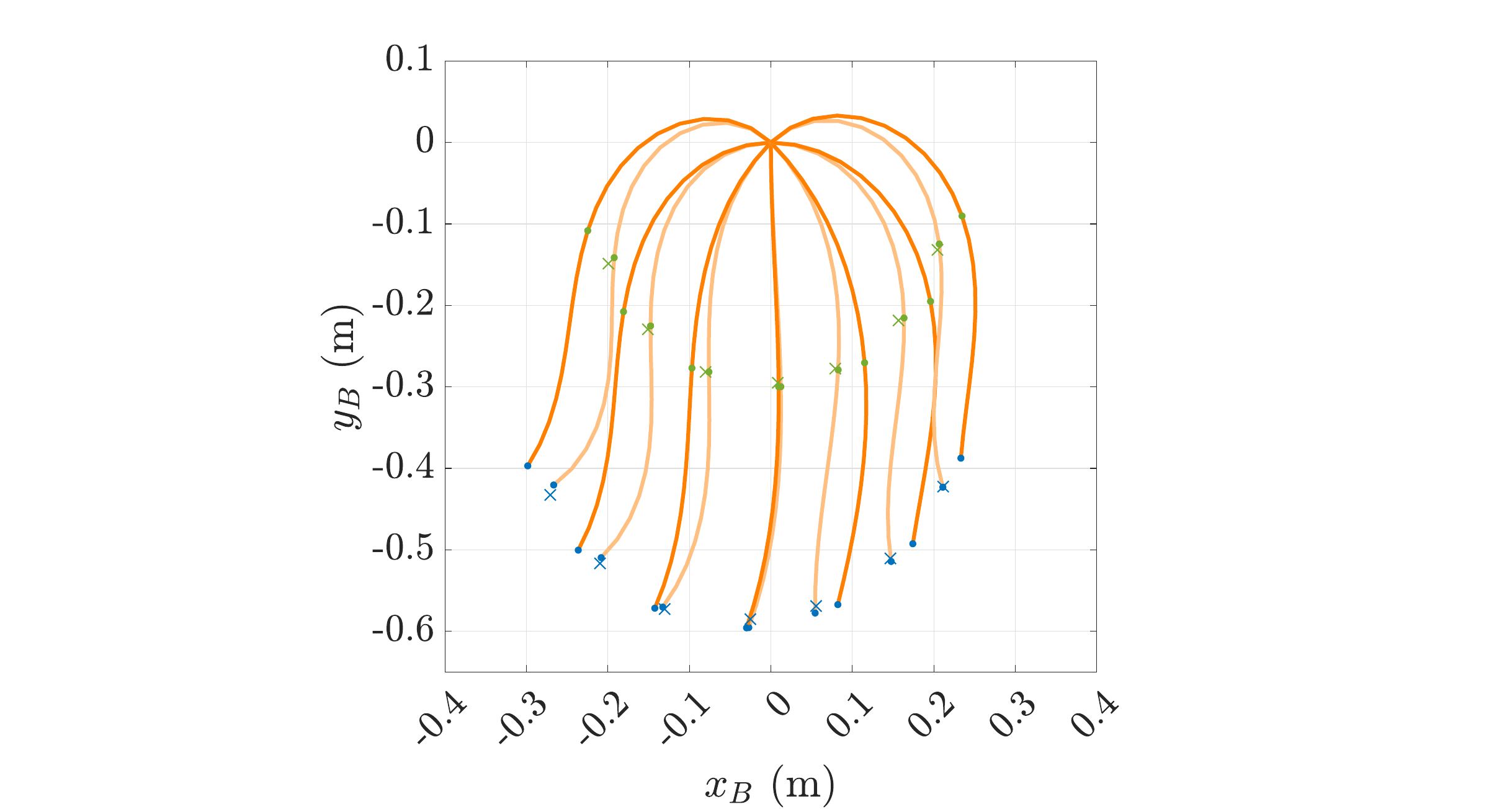}
        \caption{}
    \end{subfigure}
    \hfill
    \begin{subfigure}{0.345\columnwidth}
        \includegraphics[width=\textwidth]{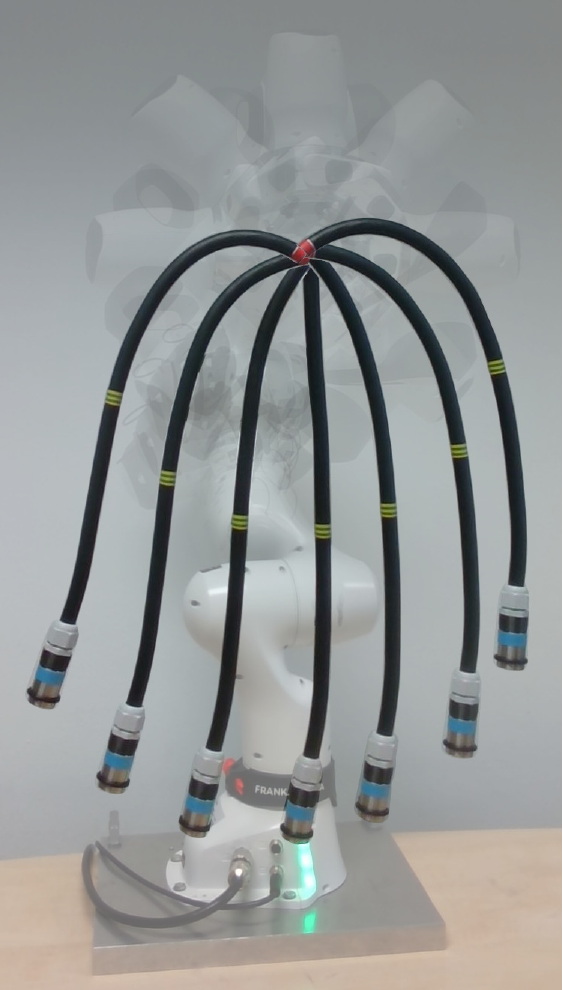}
        \caption{}
    \end{subfigure}
\caption{Comparison of the measured and modeled object equilibrium shapes for OB1. a) Here the light orange curves represent the configurations with $\Theta$ extracted from the measurements (colored crosses), and the dark orange are those obtained from modeling after parameter identification. b) Composite image showing the corresponding real object shapes.}
\label{fig:static_id_composite_and_assessment}
\end{figure}

\begin{figure*}[h]
\centering

\begin{subfigure}{0.315\textwidth}
        \includegraphics[width=\textwidth,trim= 0 0 0 0,clip=true]{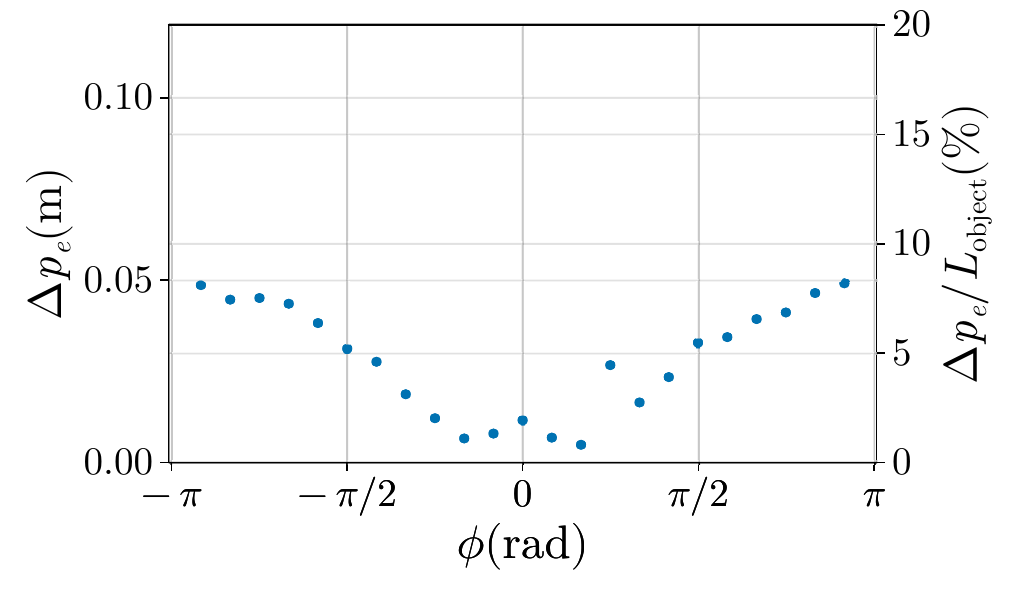}
        \caption{OB1}
    \end{subfigure}
    \hfill
\begin{subfigure}{0.315\textwidth}
        \includegraphics[width=\textwidth,trim= 0 0 0 0,clip=true]{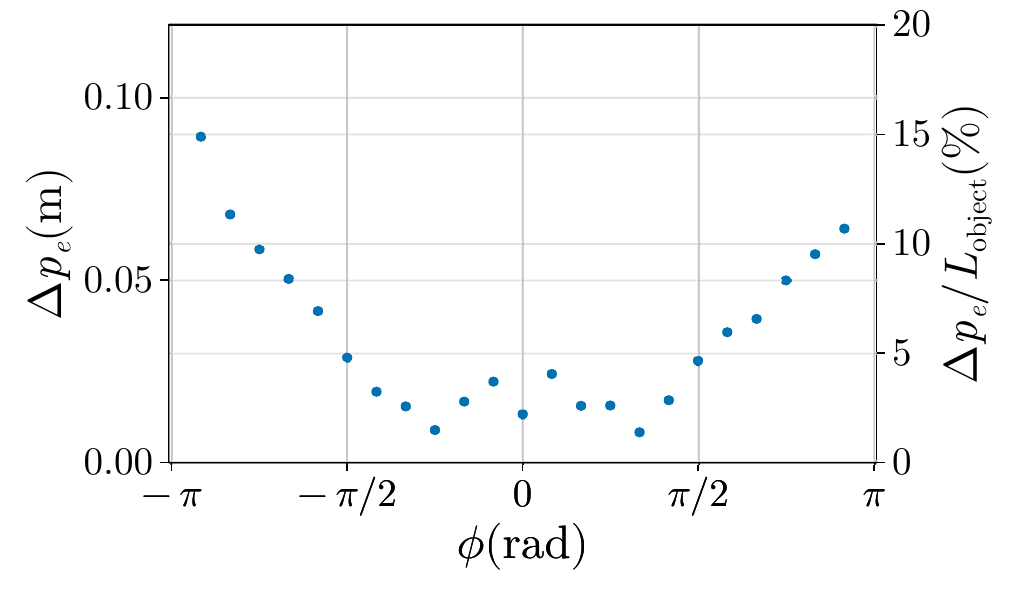}
        \caption{OB2}
    \end{subfigure}
    \hfill
\begin{subfigure}{0.315\textwidth}
        \includegraphics[width=\textwidth,trim= 0 0 0 0,clip=true]{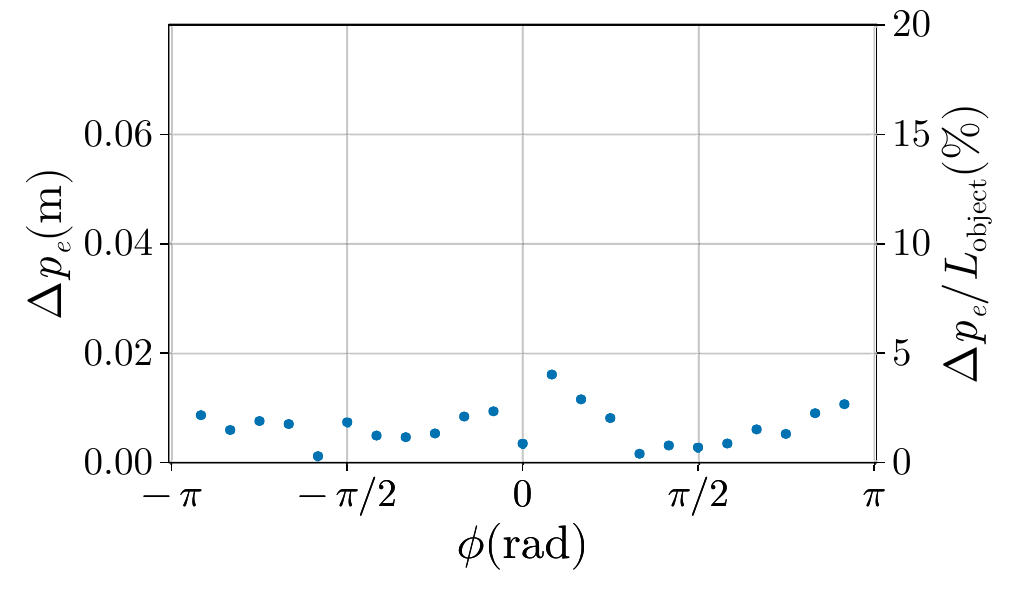}
        \caption{OB3}
    \end{subfigure}
    \hfill
\begin{subfigure}{0.315\textwidth}
        \includegraphics[width=\textwidth,trim= 0 0 0 0,clip=true]{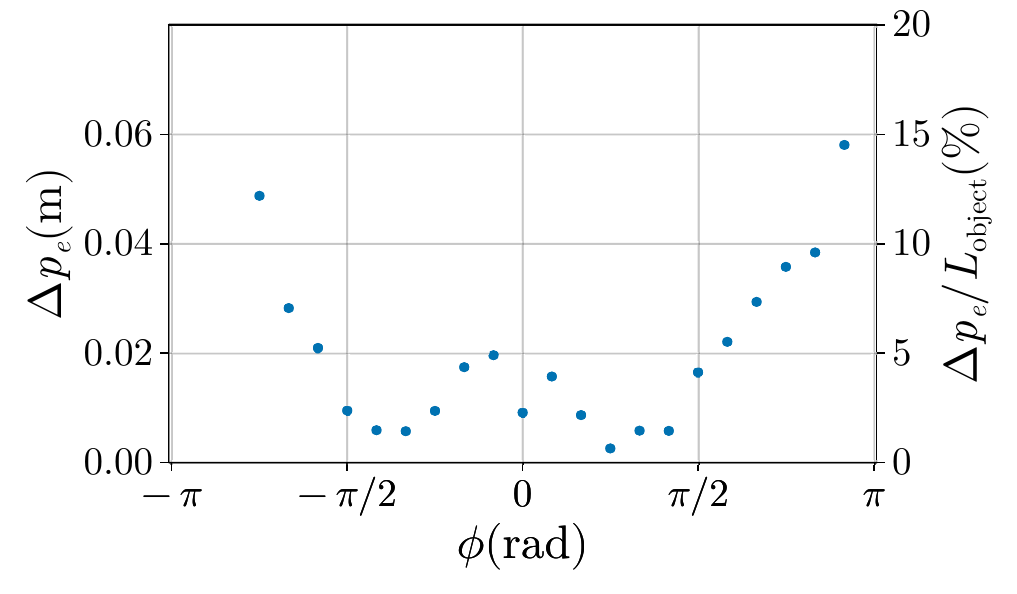}
        \caption{OB4}
    \end{subfigure}
    \hfill
\begin{subfigure}{0.315\textwidth}
        \includegraphics[width=\textwidth,trim= 0 0 0 0,clip=true]{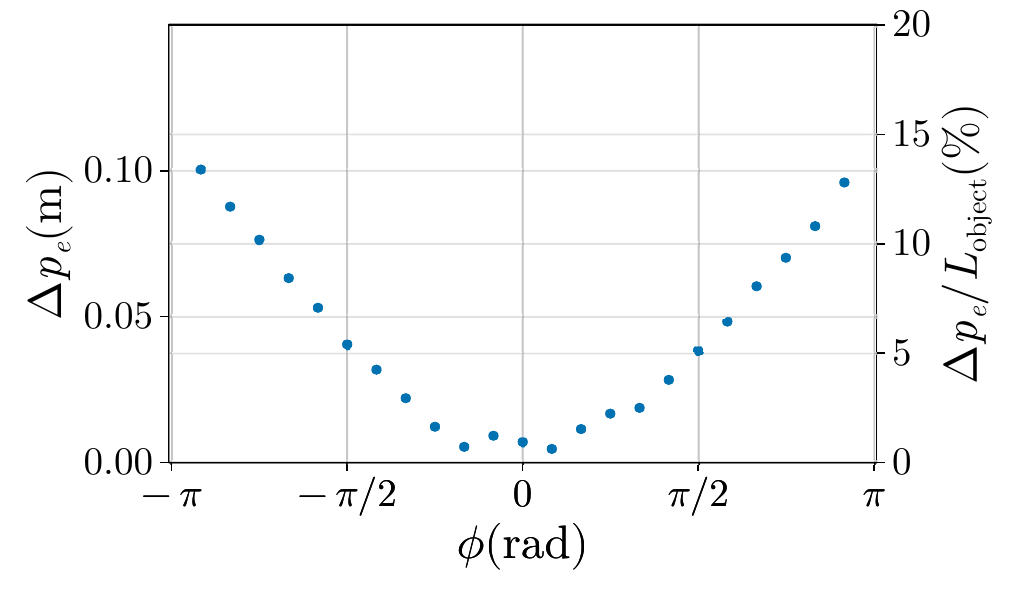}
        \caption{OB5}
    \end{subfigure}
    \hfill
\begin{subfigure}{0.315\textwidth}
        \includegraphics[width=\textwidth,trim= 0 0 0 0,clip=true]{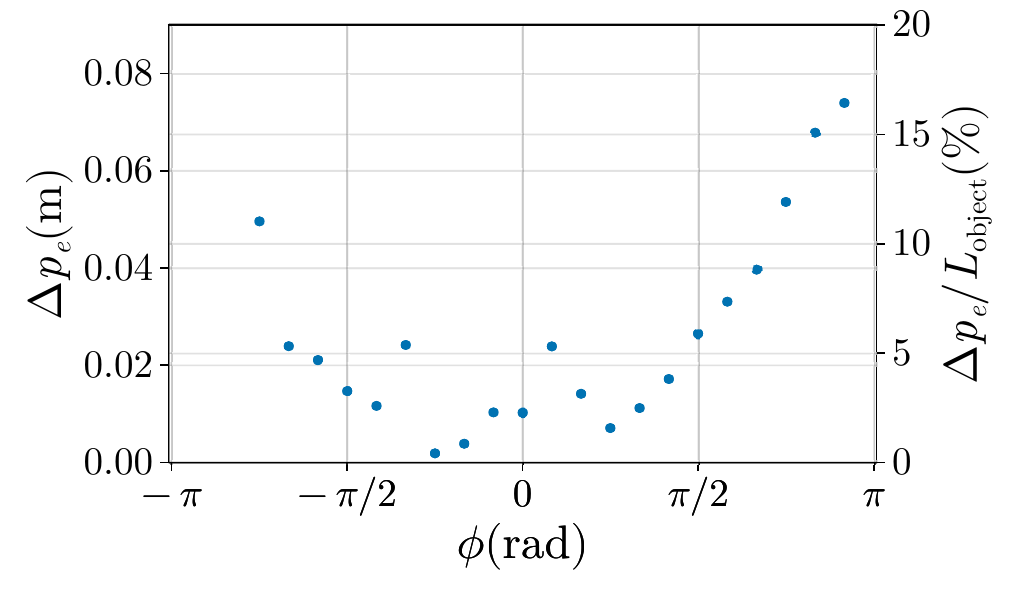}
        \caption{OB6}
    \end{subfigure}
\caption{Error between measured and modeled object endpoint $p_{\mathrm{e}}$ at equilibrium, when the object base is moved to each angle $\phi$ in the steady state model validation of Section \ref{subsec:results_param_id_static}.}
\label{fig:static_endpt_error_plot}
\end{figure*}

The error between the measured and modeled object end effector position, $\Delta p_{\mathrm{e}}$, and relative error $\Delta p_{\mathrm{e}}/L$, are plotted against the angle $\phi$ in Fig. \ref{fig:static_endpt_error_plot} for the each object, providing a measure of the modeling accuracy. The results show that the static equilibrium model provides quite an accurate estimate for the endpoint position in the range $\frac{-3\pi}{4} < \phi < \frac{3\pi}{4}$ for all object types, with an error below 0.05m (8\%) in almost all cases. Up to $\phi = \pm \pi$, most types continue to only suffer relatively mild deterioration in accuracy, with some notable exceptions. Most egregiously, OB4 and OB6 display a sudden spike in error for $\phi < \frac{-3\pi}{4}$; on closer inspection, the model with these properties is no longer multi-stable at this point, with the sole equilibrium point biased to the right-hand side of the $xy$ plane due to the offset curvature. This result shows that the in-plane assumption holds for a large range of end effector angles. It also highlights the effect out-of-plane motion of the object has on positioning accuracy. Larger out-of-plane movement is seen at higher angles of $\phi$, due to the gravitational forces no longer keeping a sufficient proportion of the object mass below the grasp point, removing this stabilizing effect.

Another limitation of the model is highlighted by OB5, with its worst-case combination of length, stiffness, and endpoint mass resulting in high curvature near the base followed by straightening out along the length. The affine curvature model cannot accurately describe this over a single segment, as will be shown later.

\subsubsection{Dynamic Response}\label{subsec:results_param_id_dynamic}

The data for dynamic identification was collected from a single evolution of a pendulum motion for each object with the base fixed. In these experiments, the robot was set to $\phi = 0$, starting from a deformed position far from equilibrium to observe the evolution of the states of the
object. Fig. \ref{fig:dyn_id_comparison}a shows a time lapse of the images captured for OB1, along with the detected positions of $p_s$, $p_{\mathrm{m}}$ and $p_{\mathrm{e}}$, as well as the $\Theta$ configurations from inverse kinematics projected into the camera frame. Note that this shows only the first swing, the full sets of data run until the oscillations cease. This data was used to estimate the value of $\beta$, and the resulting fully identified model was then simulated from the same initial conditions. In Fig. \ref{fig:dyn_id_comparison}b this simulated evolution is then projected back onto the same captured images to show how the simulated object shape compares to the real object in Cartesian space, with only every second frame included for clarity. The time lapse images were not created for each object, but videos are available with the extra materials.

\begin{figure}[h]
\centering
    \begin{subfigure}{0.485\columnwidth}
        \includegraphics[width=\textwidth]{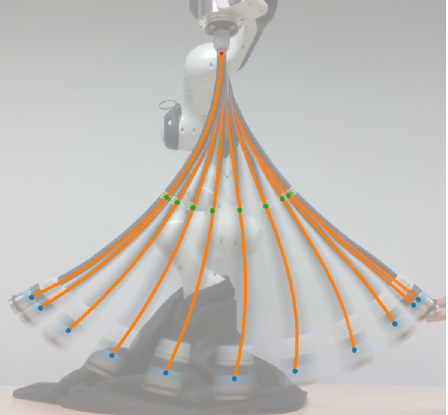}
        \caption{}
    \end{subfigure}
    \hfill
    \begin{subfigure}{0.49\columnwidth}
        \includegraphics[width=\textwidth]{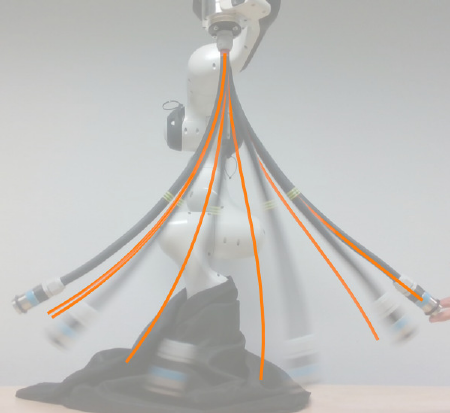}
        \caption{}
    \end{subfigure}
\caption{Partial time lapse comparison of the dynamic evolution of $\Theta$ for OB1. The configurations extracted from the images are shown in (a), and the simulated evolution, after using this data for identification, is projected over the captured images in (b), for every second frame only.}
\label{fig:dyn_id_comparison}
\end{figure}

\begin{figure*}[t]
\centering
\begin{subfigure}{0.325\textwidth}
        \includegraphics[trim=2.6cm 0.3cm 3.8cm 1.2cm,width=\textwidth,clip=true]{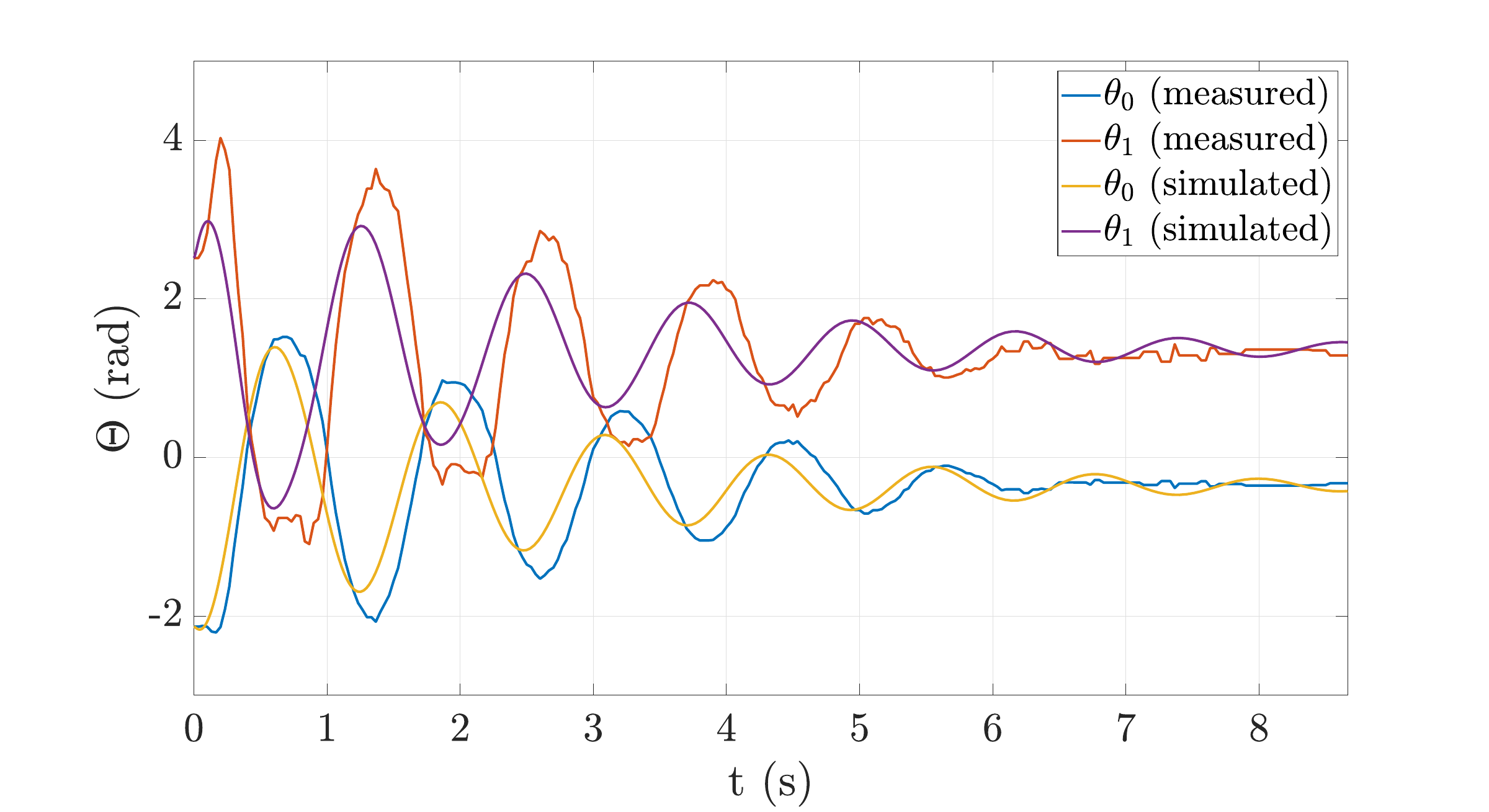}
        \caption{OB1}
    \end{subfigure}
    \hfill
\begin{subfigure}{0.325\textwidth}
        \includegraphics[trim=2.6cm 0.3cm 3.8cm 1.2cm,width=\textwidth,clip=true]{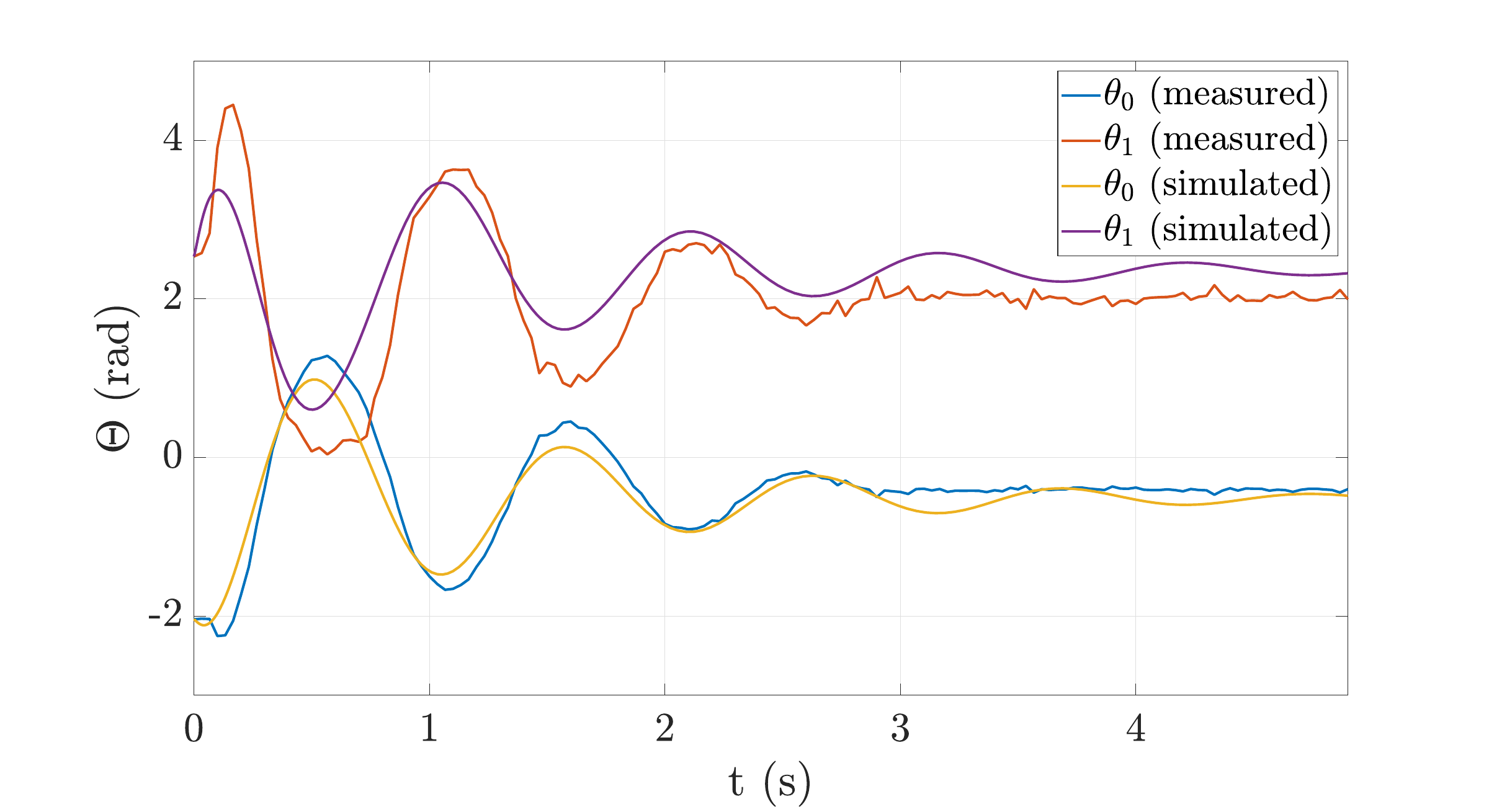}
        \caption{OB2}
    \end{subfigure}
    \hfill
\begin{subfigure}{0.325\textwidth}
        \includegraphics[trim=2.6cm 0.3cm 3.8cm 1.2cm,width=\textwidth,clip=true]{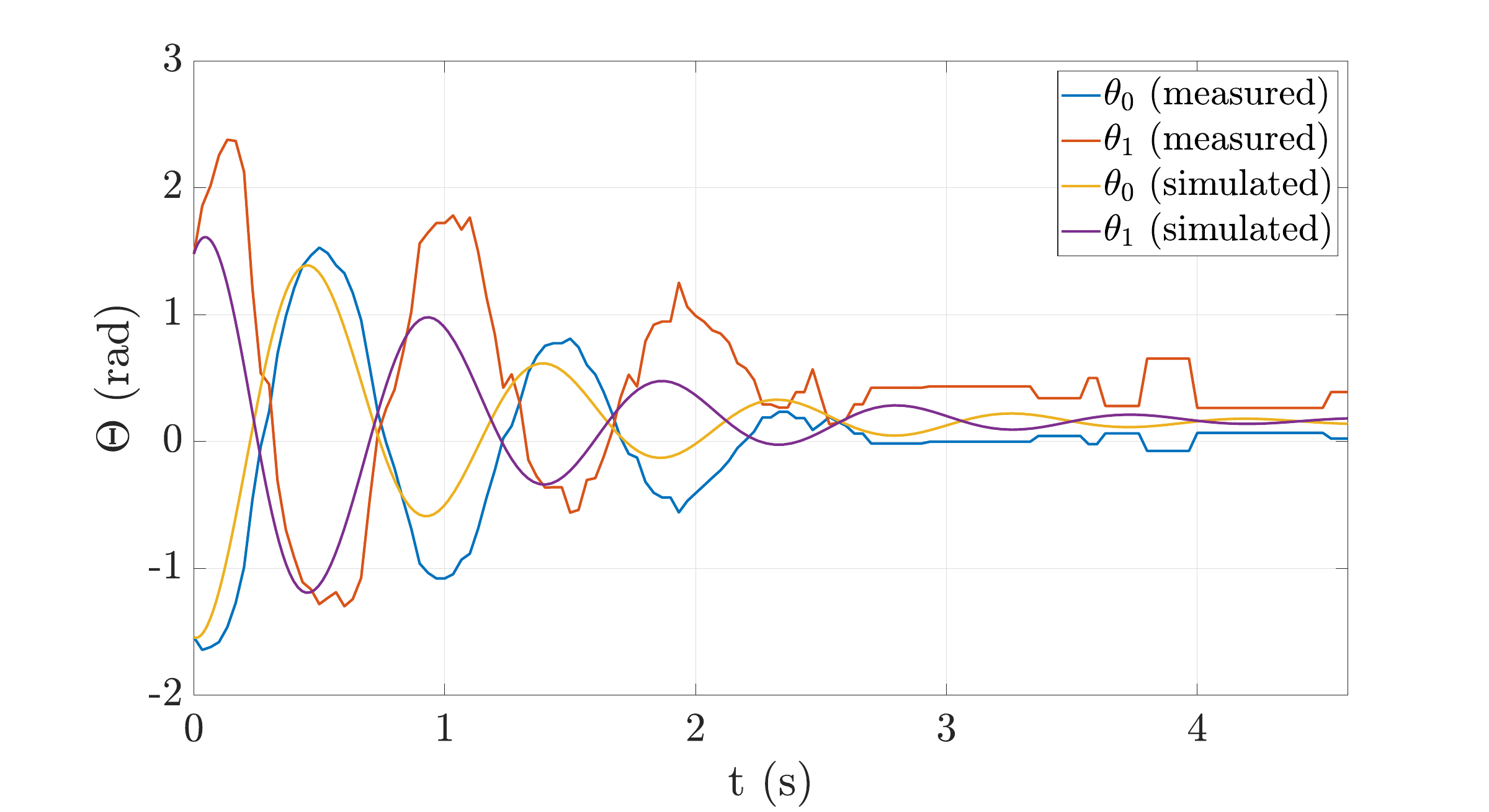}
        \caption{OB3}
    \end{subfigure}
    \hfill
\begin{subfigure}{0.325\textwidth}
        \includegraphics[trim=2.6cm 0.3cm 3.8cm 1.2cm,width=\textwidth,clip=true]{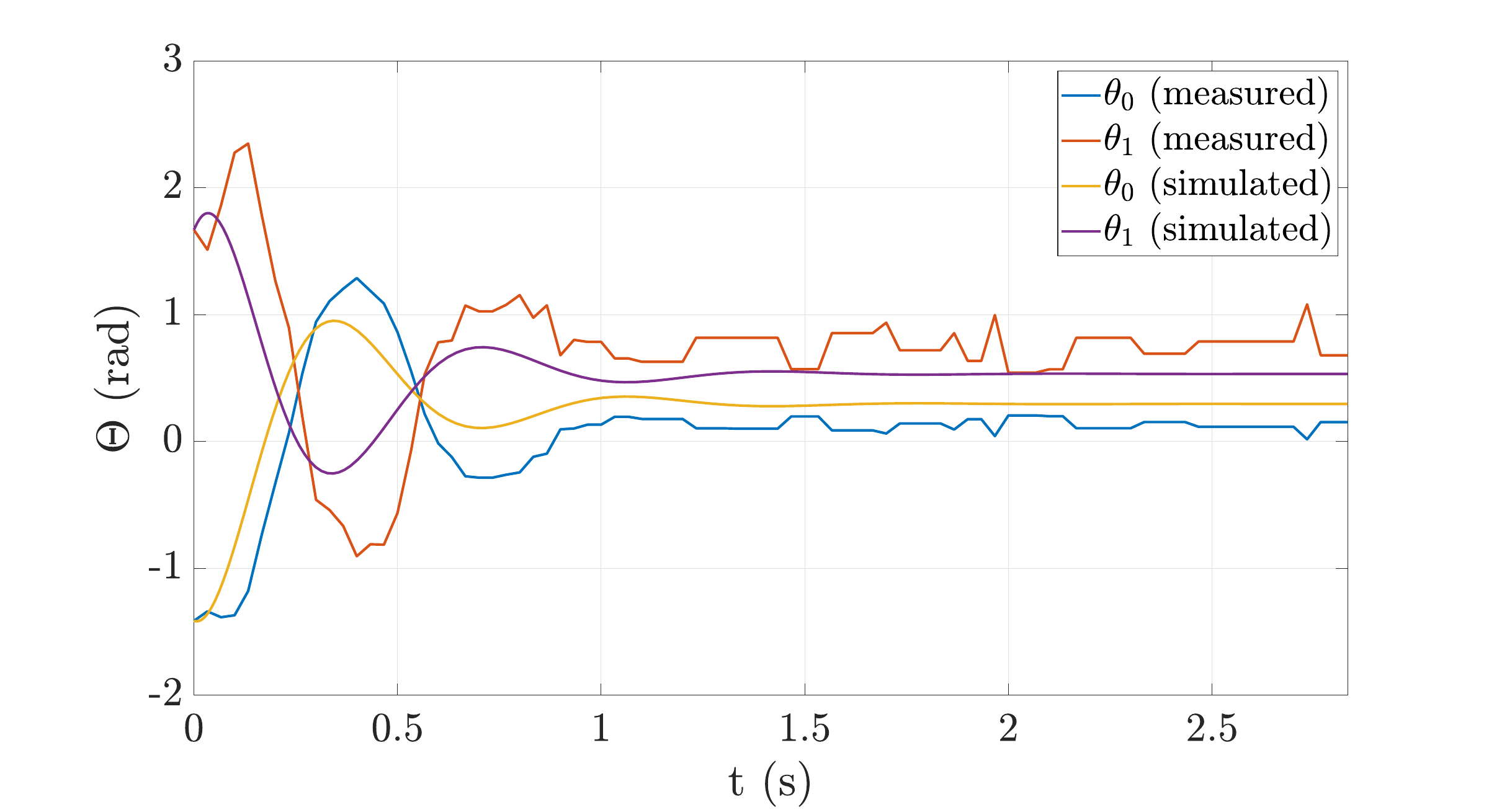}
        \caption{OB4}
    \end{subfigure}
    \hfill
\begin{subfigure}{0.325\textwidth}
        \includegraphics[trim=2.6cm 0.3cm 3.8cm 1.2cm,width=\textwidth,clip=true]{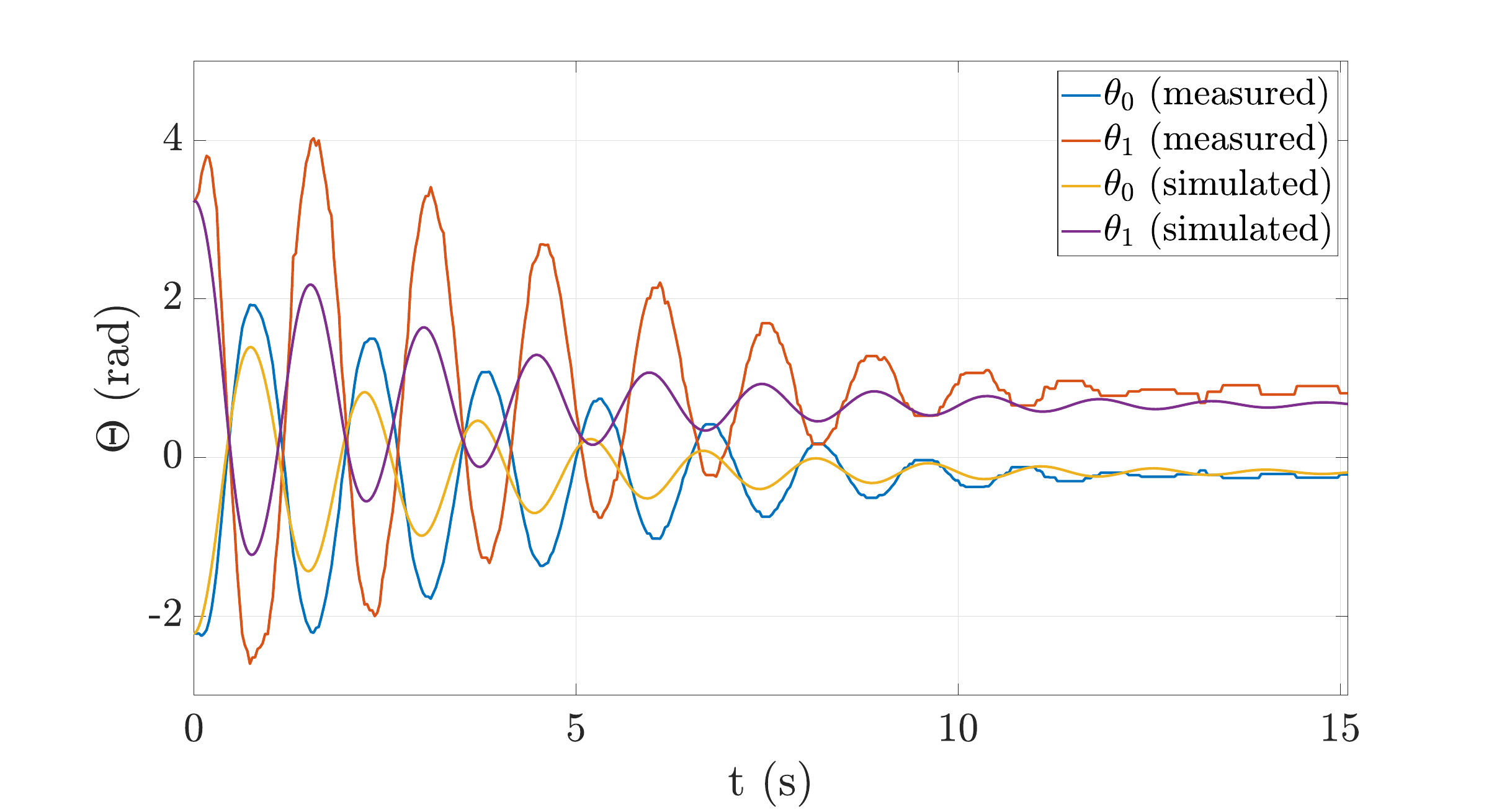}
        \caption{OB5}
    \end{subfigure}
    \hfill
\begin{subfigure}{0.325\textwidth}
        \includegraphics[trim=2.6cm 0.3cm 3.8cm 1.2cm,width=\textwidth,clip=true]{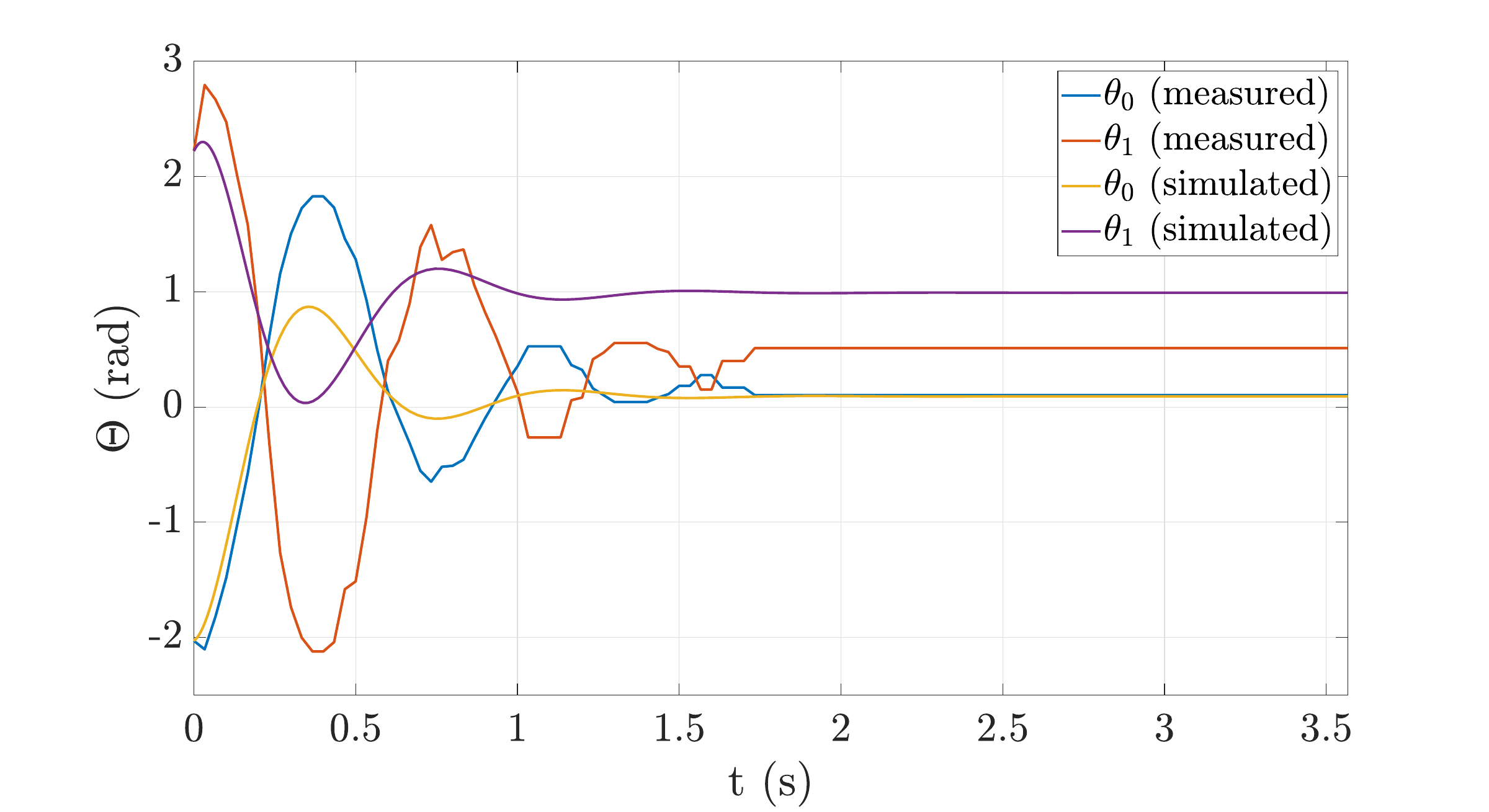}
        \caption{OB6}
    \end{subfigure}
\caption{Comparison between $\Theta$ state evolutions undergoing pendulum motion from the dynamic model validation of Section \ref{subsec:results_param_id_dynamic}. The measured data in the graphs is extracted from the image frames, and the simulated traces are generated from the model using the same initial $\Theta$ and with $\dot{\Theta}(t=0)=0$.}
\label{fig:dyn_id_evolution_comparison}
\end{figure*}

Plots comparing the evolution of $\Theta$ measured from the experiment and generated by the simulation are shown in Fig. \ref{fig:dyn_id_evolution_comparison}. Several of the objects demonstrate quite closely matching responses, though there are some discrepancies for others. This is most obvious when comparing some of the amplitudes, however it is worth noting that as the two components of $\Theta$ generally occur with opposite phase, these plots can be a misleading indication of how the object shapes in Cartesian space compare. The discrepancies in the steady-state results are attributed to the hysteresis effect of the DLO, which tends towards the offset configuration, $\Bar{\Theta}$. The hysteresis effect causes slight alterations in this configuration producing the errors seen observed in the figure. The sensitivity of our model to the hysteresis effect is analyzed in a subsequent section. Conversely, the phase lead in these plots that tends to appear fairly minor for most objects, is more clearly visible in Fig. \ref{fig:dyn_id_comparison}b. The phase lead results from a slight mismatch in the initial state guess used during the dynamic simulation, as observed in Fig. \ref{fig:dyn_id_evolution_comparison}. This discrepancy is attributed to minor variations in the parameters of the DLO. These factors make it particularly challenging to establish an accurate initial guess for validating the model. To better evaluate the dynamic evolution of the system, we compare the time evolution of the mid-point and end-effector coordinates of the DLO as observed in the experiment with those generated by our model in simulation, as shown in Fig. \ref{fig:dyn_pos_comparison}.

\subsection{Influence of hysteresis in our model}

In this section, we have analyzed the effect of hysteresis for a specific object (OB1) to quantify this effect in our model. Our observations indicate that the hysteresis effect of DLO can be effectively captured by adjusting the offset $\bar{\Theta}$ in the elastic force term. In the model, this parameter was identified to minimize the error across various shapes and was kept constant.
However, we observed that if this parameter $\bar{\Theta}$ is identified separately for each case, its value can vary in the range [$-2\bar{\theta}_{i}$, \ $2\bar{\theta}_{i}$] considering that $\bar{\theta}_{i}$ is the nominal value (see Table \ref{tab:params}).

\begin{figure}[h]
\centering
    \includegraphics[width=0.98\columnwidth]{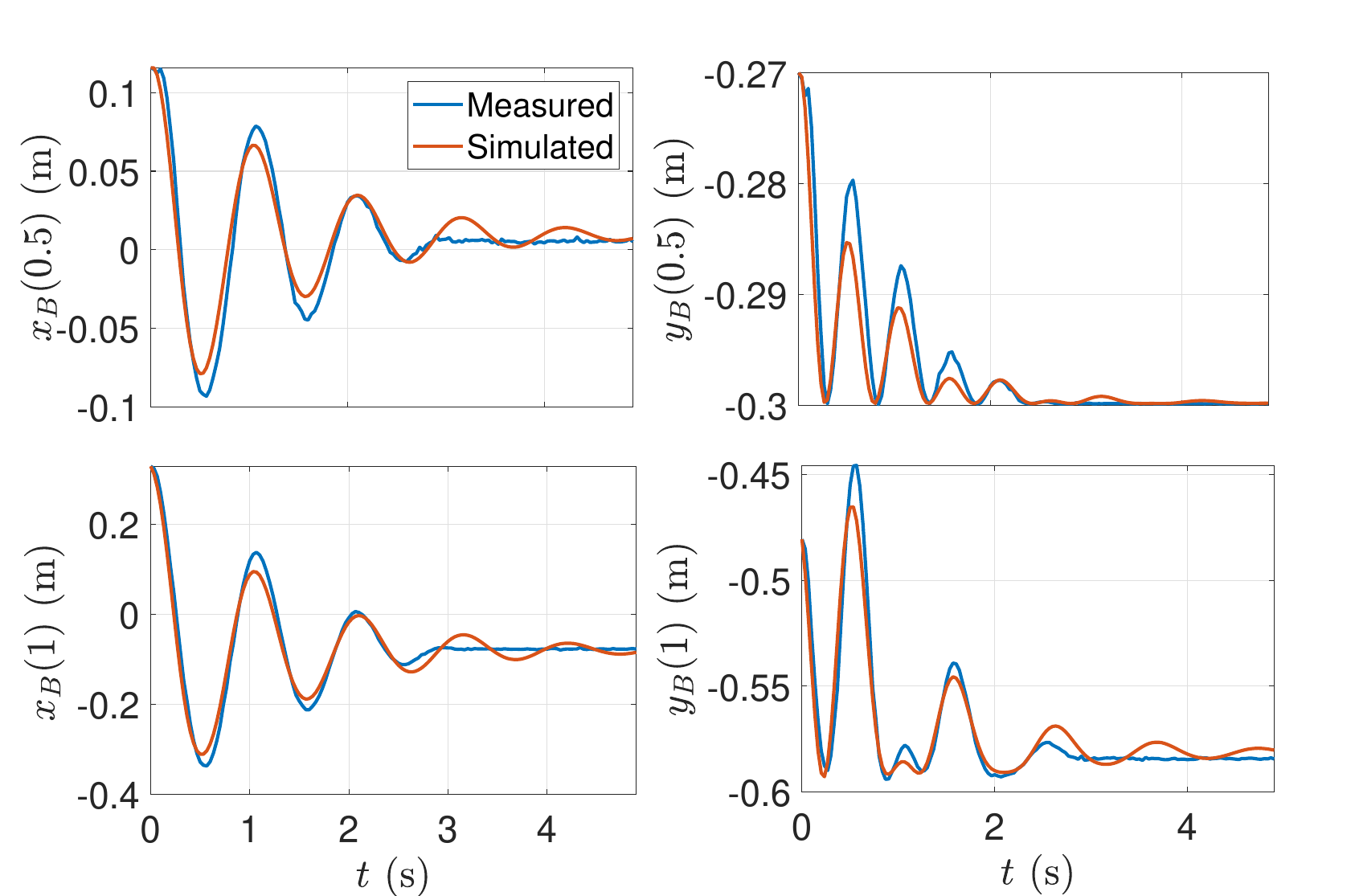}
\caption{Comparison of the end-effector coordinates of the DLO to evaluate the dynamic evolution of the system for OB2.}
\label{fig:dyn_pos_comparison}
\end{figure}

\begin{figure}[h]
\centerline{\includegraphics[width=0.49\textwidth]{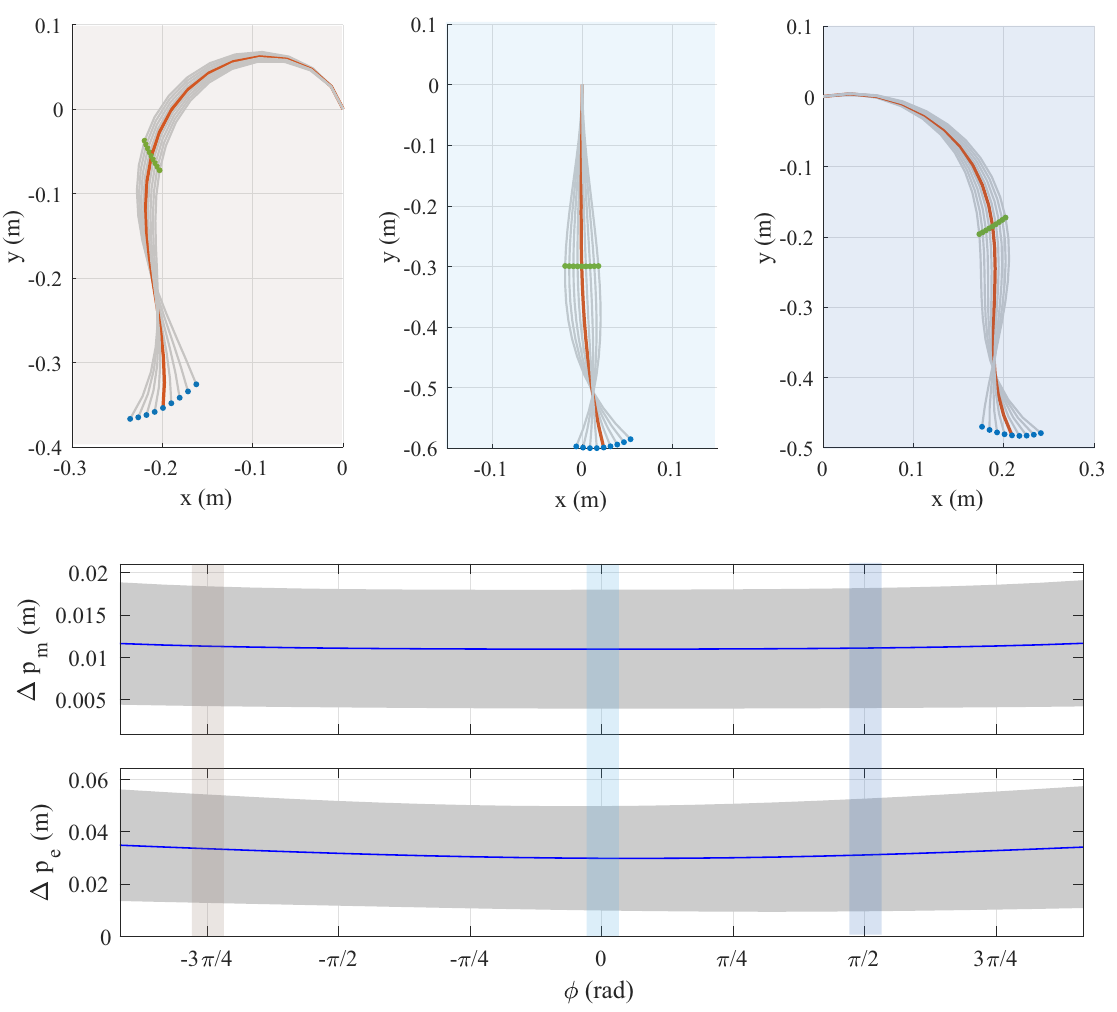}}
\caption{Influence of hysteresis for different configurations within the observed experimental range. The nominal case is plotted with an orange line in the upper part, while the different shapes using hysteresis are shown in grey.}
\label{fig:nfluence_hysteresis}
\end{figure}

Taking this into account, we performed a sensitivity analysis to examine how the shapes and errors change within this range for different configurations. The results of this analysis are presented in Fig. \ref{fig:nfluence_hysteresis} where we analyze the sensitivity of our model to the hysteresis effect under different orientations of the robot's end-effector. This effect results in an average error of 5 $\%$ and 7.5 $\%$ of the DLO's length at the midpoint and endpoint, respectively. The error can be significantly reduced by loosening the DLO through shaking prior to conducting an experiment.

\section{Experimental Validation of the Controller}\label{sec:V-controller-validation}

The control framework was experimentally validated, focusing on control of the object endpoint as a practical goal specification. Due to the underactuated nature of DLO, not all configurations are attainable, as they must satisfy the static equilibrium condition. Therefore, we propose solving the kinematic inversion of our model for a given desired task coordinate that satisfies the static equilibrium condition through an optimization process.

\subsection{Position}\label{subsec:validation_position}

In the first case we look at positioning the object endpoint at a goal $p^*$ only, so that $\zeta = p_e$, and take the Euclidean distance error as the optimization cost. We use a circular constraint on the floating base coordinates to represent the workspace of the manipulator, specifically a circle of radius 0.5m centred at $(0,0.333)$ in $\{S_B\}$, which approximates the reach of the FR3. Since the modeling accuracy and planar assumption start to deteriorate at higher base orientations, it was also decided to limit the range of angles to $\frac{-3\pi}{4} \leq \phi \leq \frac{3\pi}{4}$. The optimization problem is therefore:
\begin{equation}\label{eqn:position_NLO}
\begin{split}
    &\min_{\Theta^* \in \mathbb{R}^{n+1},(x^*,y^*,\phi^*)\in\mathbb{F}}
    \quad \|p^*-p_e(x^*,y^*,\phi^*,\Theta^*)\|_2\\
    &\mathrm{s.t.} \quad G_{\Theta}(\Theta^*,\phi^*) + kH{(\Theta^*-\bar{\Theta})} = 0\\
    \mathbb{F} = \{x^2&+(y-0.333)^2-0.5^2 \leq 0\} \times \{\frac{-3\pi}{4} \leq \phi \leq \frac{3\pi}{4}\}.
\end{split}
\end{equation}

We used the controller to position the object endpoint by solving \eqref{eqn:position_NLO} at each point of a grid of endpoint goals, with 0.1m spacing. The grid covers an area in $\{S_B\}$ of $-0.7 \leq x \leq 0.7$ m and $0.05 \leq y \leq 0.75$ m, although depending on the object type it may not be necessary to cover the full height in the $y$-direction once it exceeds the reachable range. The MATLAB \texttt{fmincon} solver was used to obtain the solutions, and results were obtained for OB1, OB2, OB3 and OB5, as their properties cover a wide range of high deformation behaviour. 
Additionally, a 'model-free' reference was evaluated to compare our method with alternative strategy for positioning the DLO's end-effector, without considering its curvature model. This strategy involves constraining $\phi=0$ and positioning the manipulator at a constant offset to $p^*$ (or as close as possible, given the manipulator's workspace constraint), which is derived from the measured steady state $p_e$.

Fig. \ref{fig:opt_sols} shows the manipulator solutions and corresponding equilibrium configurations of the modeled object for each row of the workspace validation in both cases for OB1. Lines are drawn pairing the goals with the achieved modeled endpoints, and the red circle indicates the constraint placed on the manipulator endpoint position.

\begin{figure}[t]
\includegraphics[trim={3.4cm 3.9cm 5.7cm 0.7cm},width=0.525\columnwidth,clip]{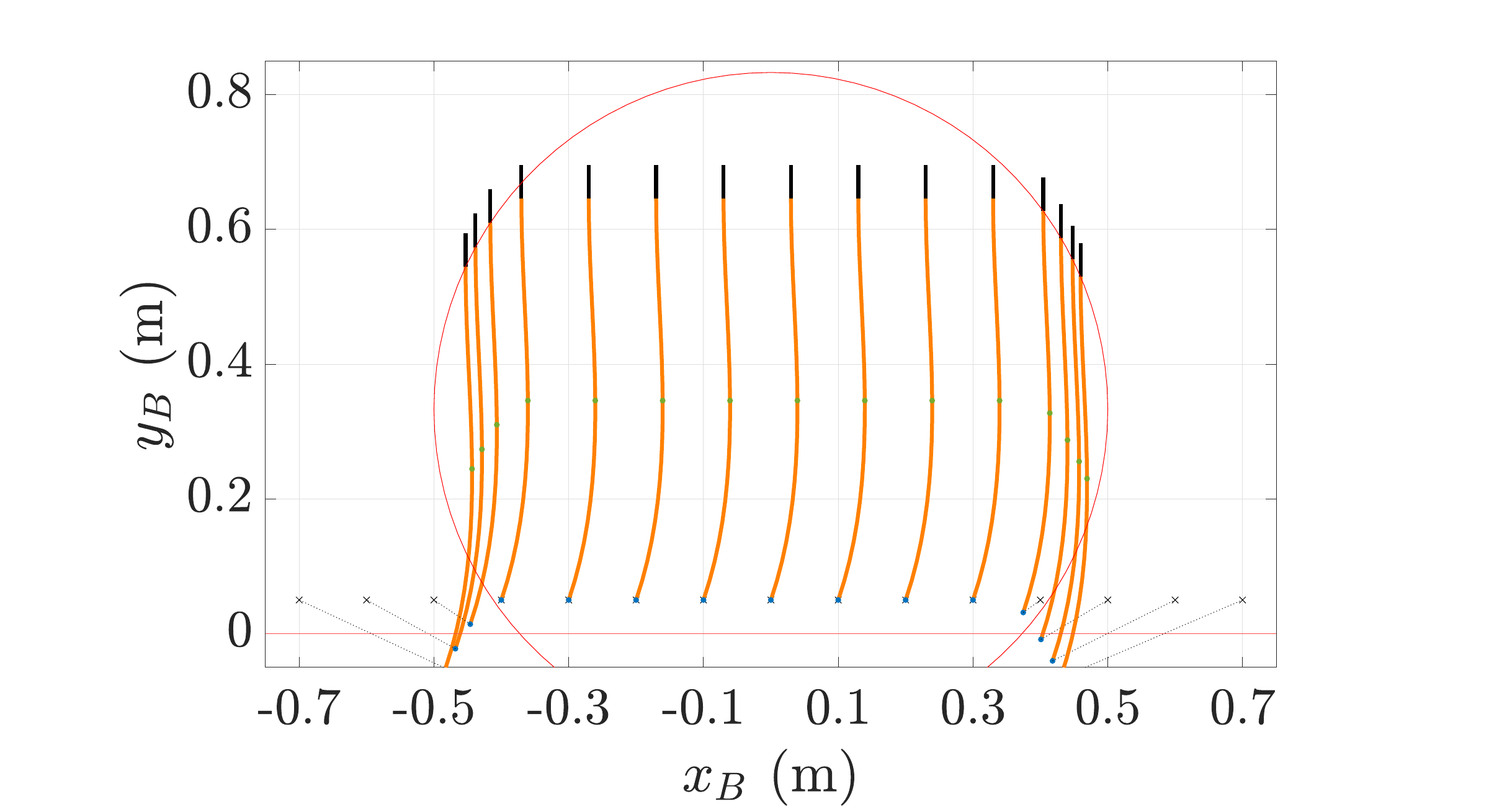} 
\hfill
\includegraphics[trim={7.2cm 3.9cm 5.7cm 0.7cm},width=0.46\columnwidth,clip]{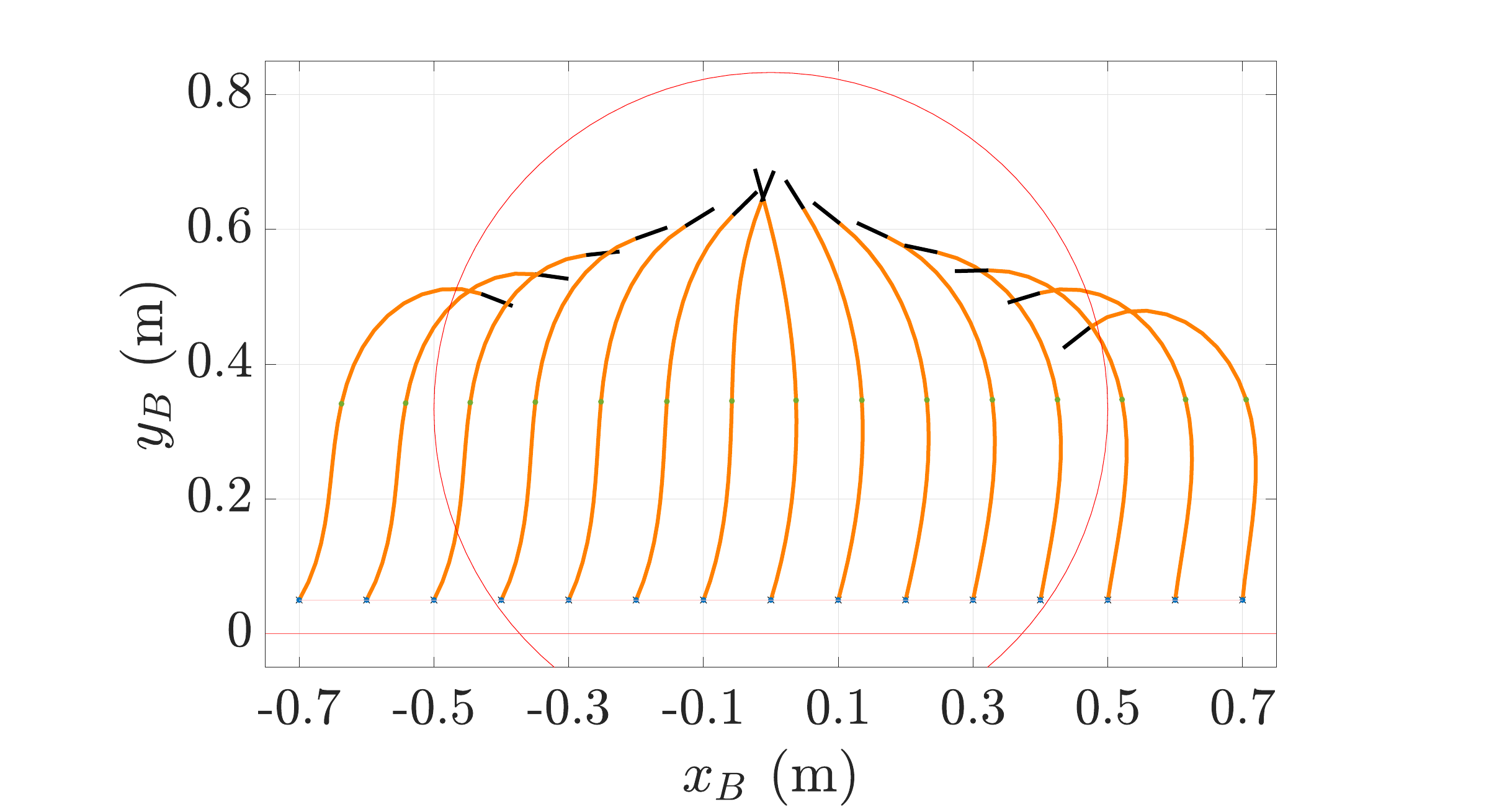}
\includegraphics[trim={3.4cm 3.9cm 5.7cm 0.7cm},width=0.525\columnwidth,clip]{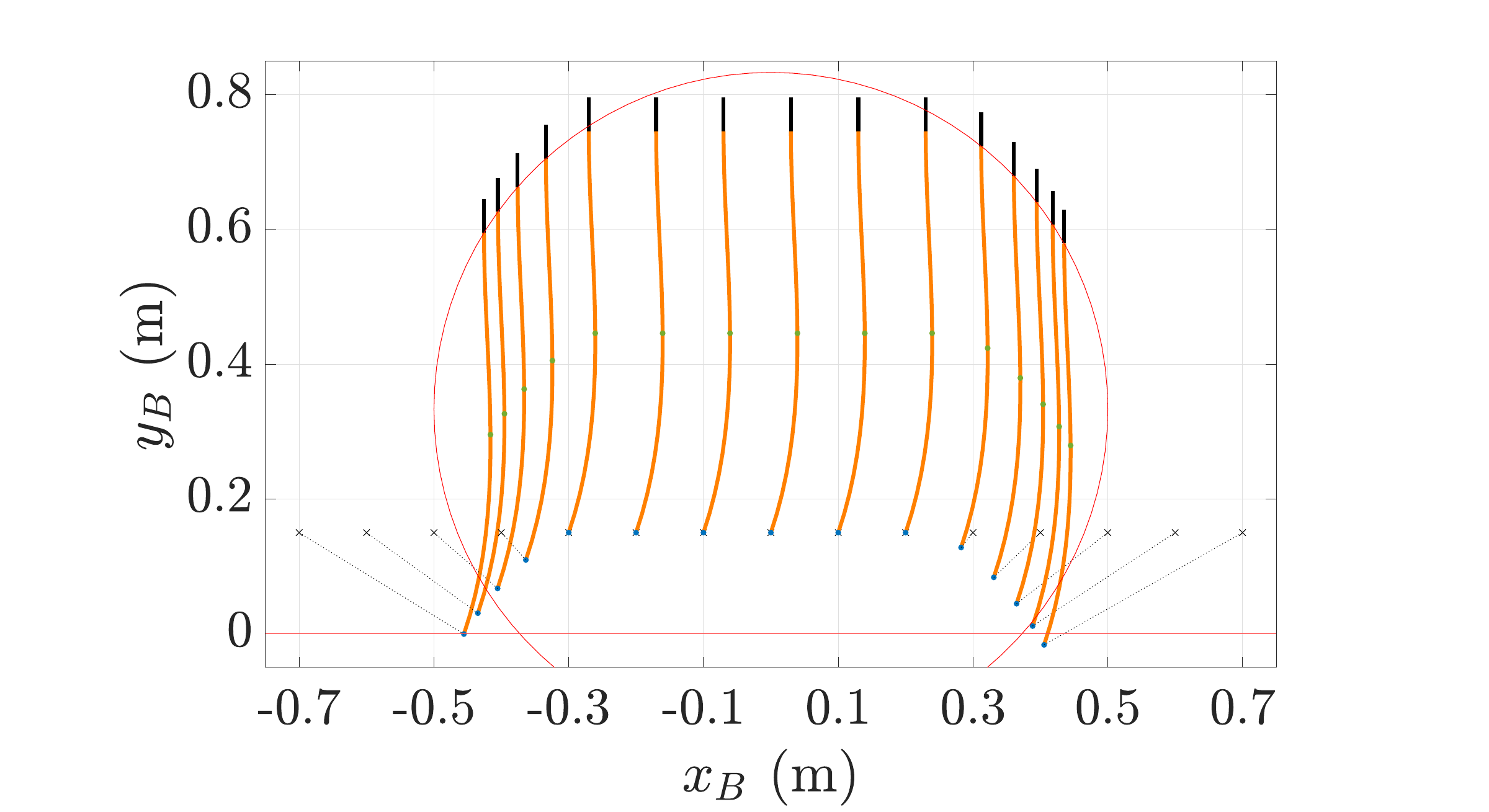} 
\hfill
\includegraphics[trim={7.2cm 3.9cm 5.7cm 0.7cm},width=0.46\columnwidth,clip]{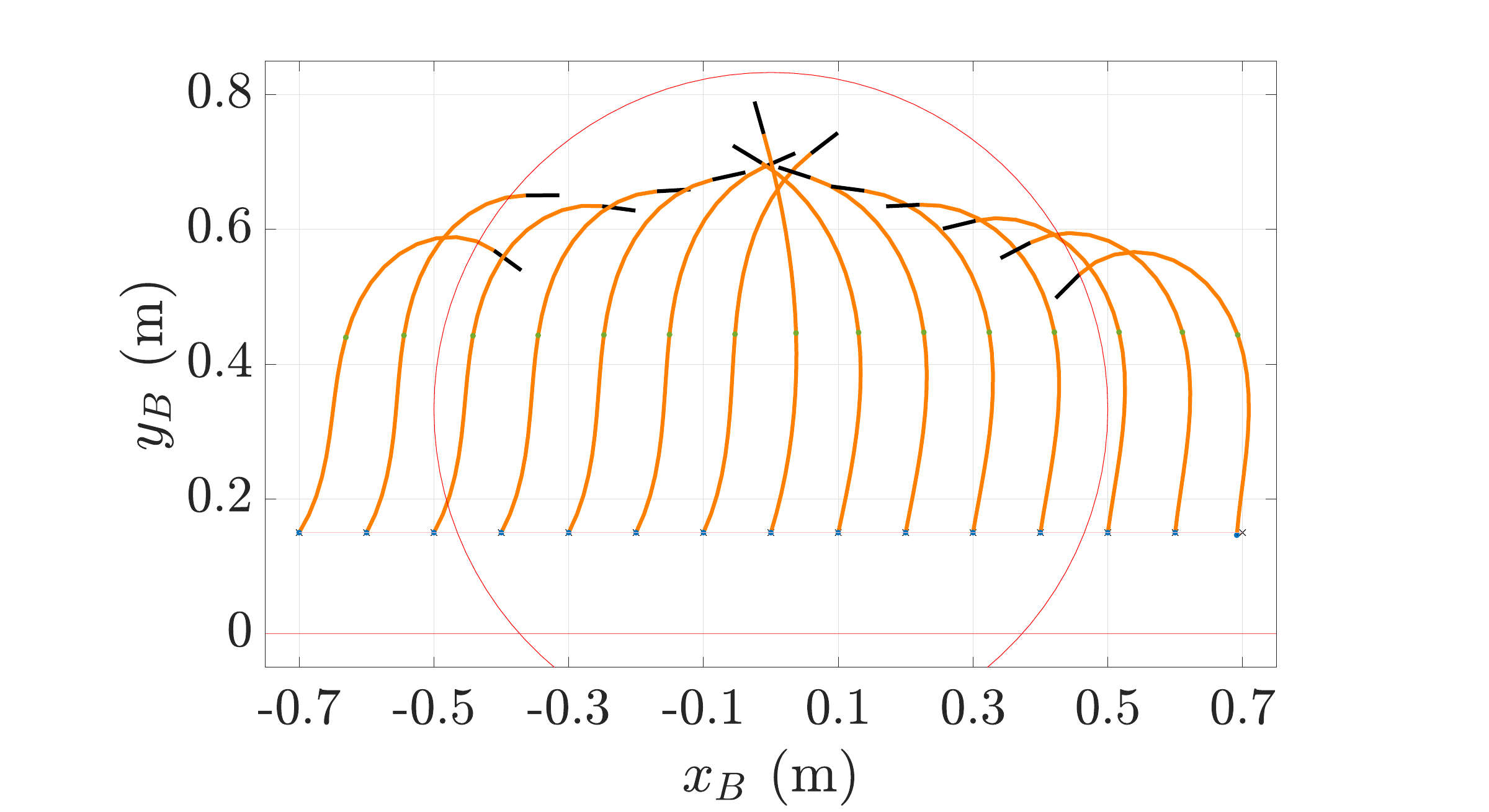}
\includegraphics[trim={3.4cm 3.9cm 5.7cm 0.7cm},width=0.525\columnwidth,clip]{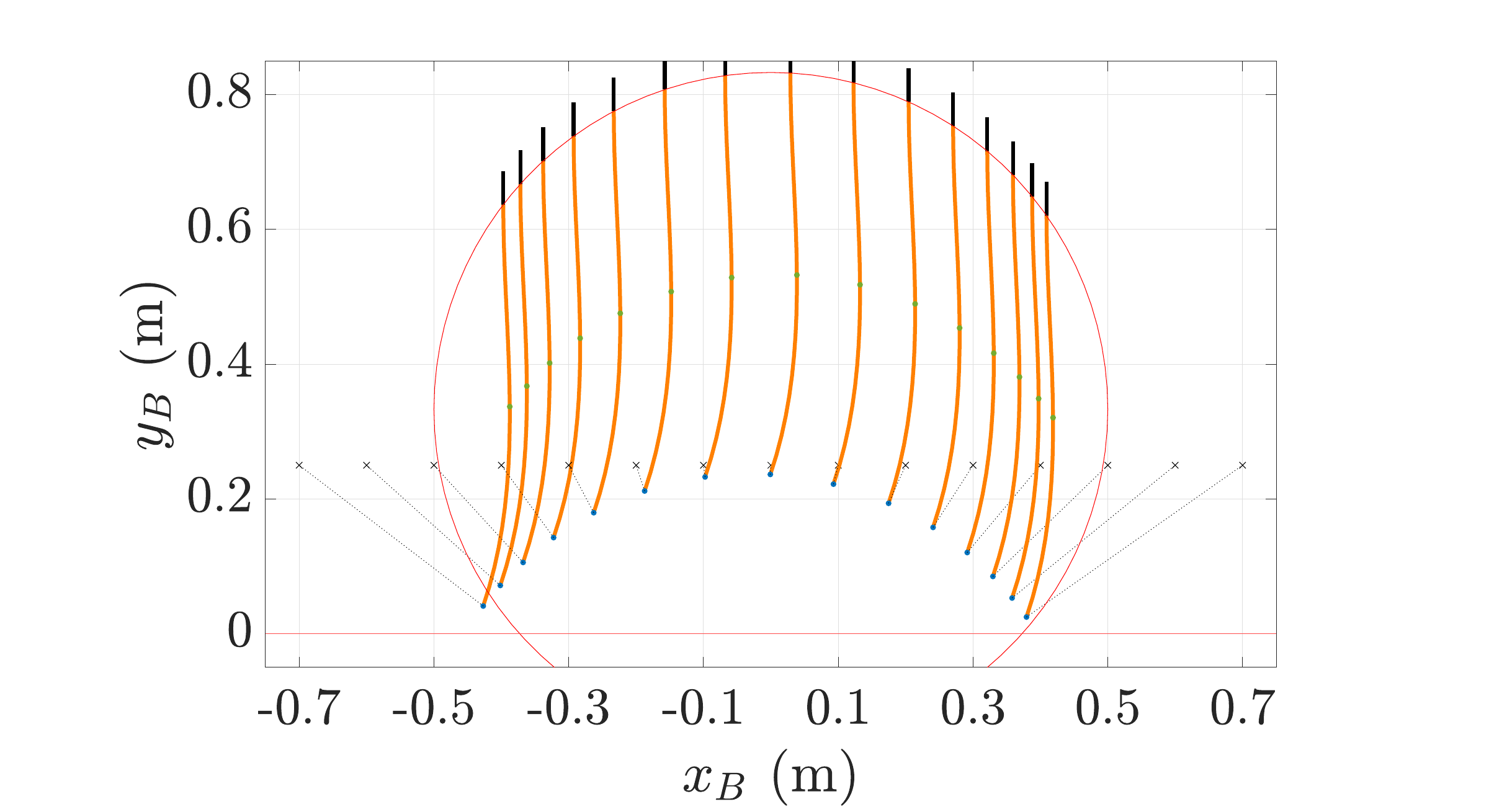} 
\hfill
\includegraphics[trim={7.2cm 3.9cm 5.7cm 0.7cm},width=0.46\columnwidth,clip]{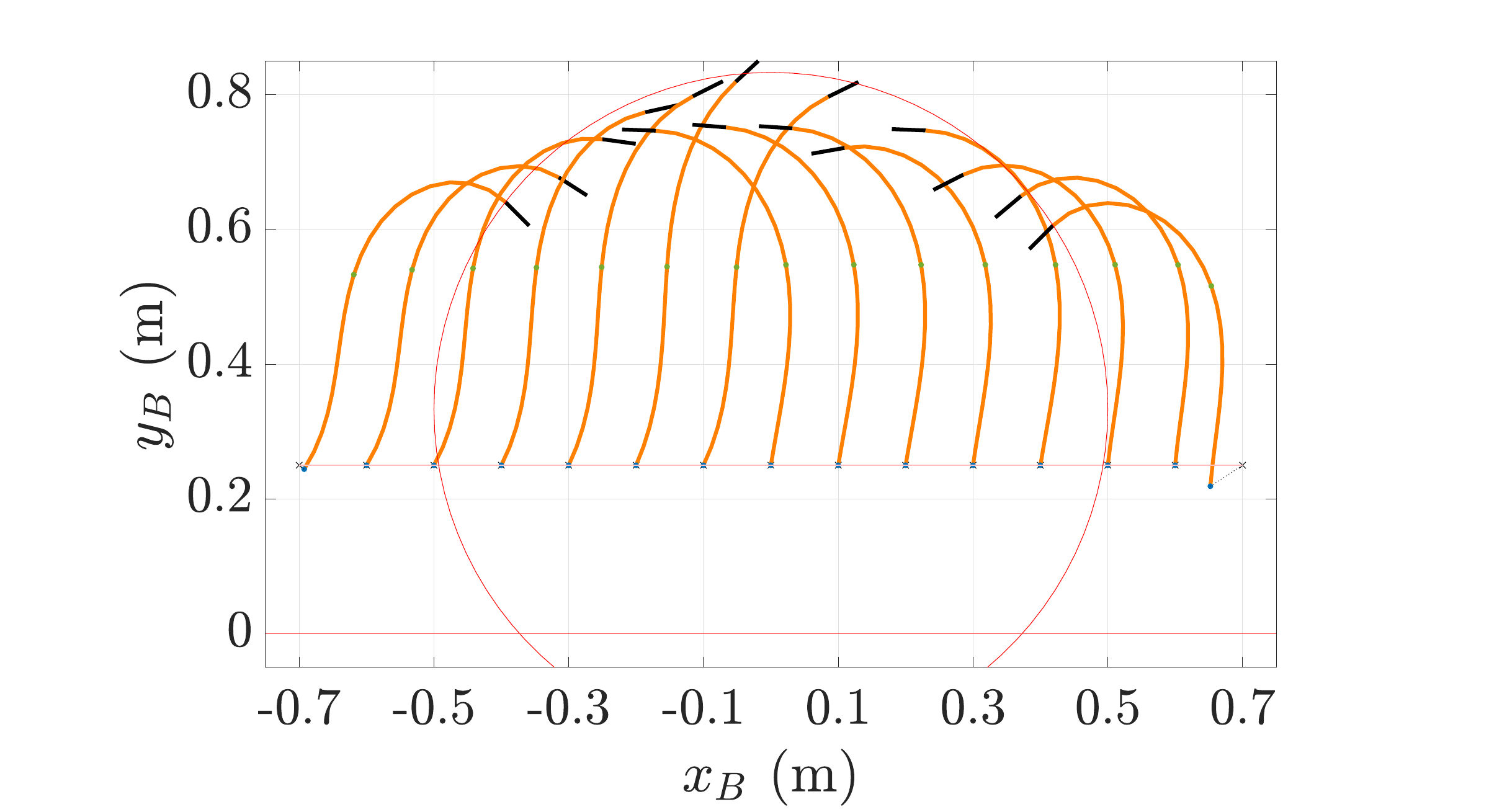}
\includegraphics[trim={3.4cm 3.9cm 5.7cm 0.7cm},width=0.525\columnwidth,clip]{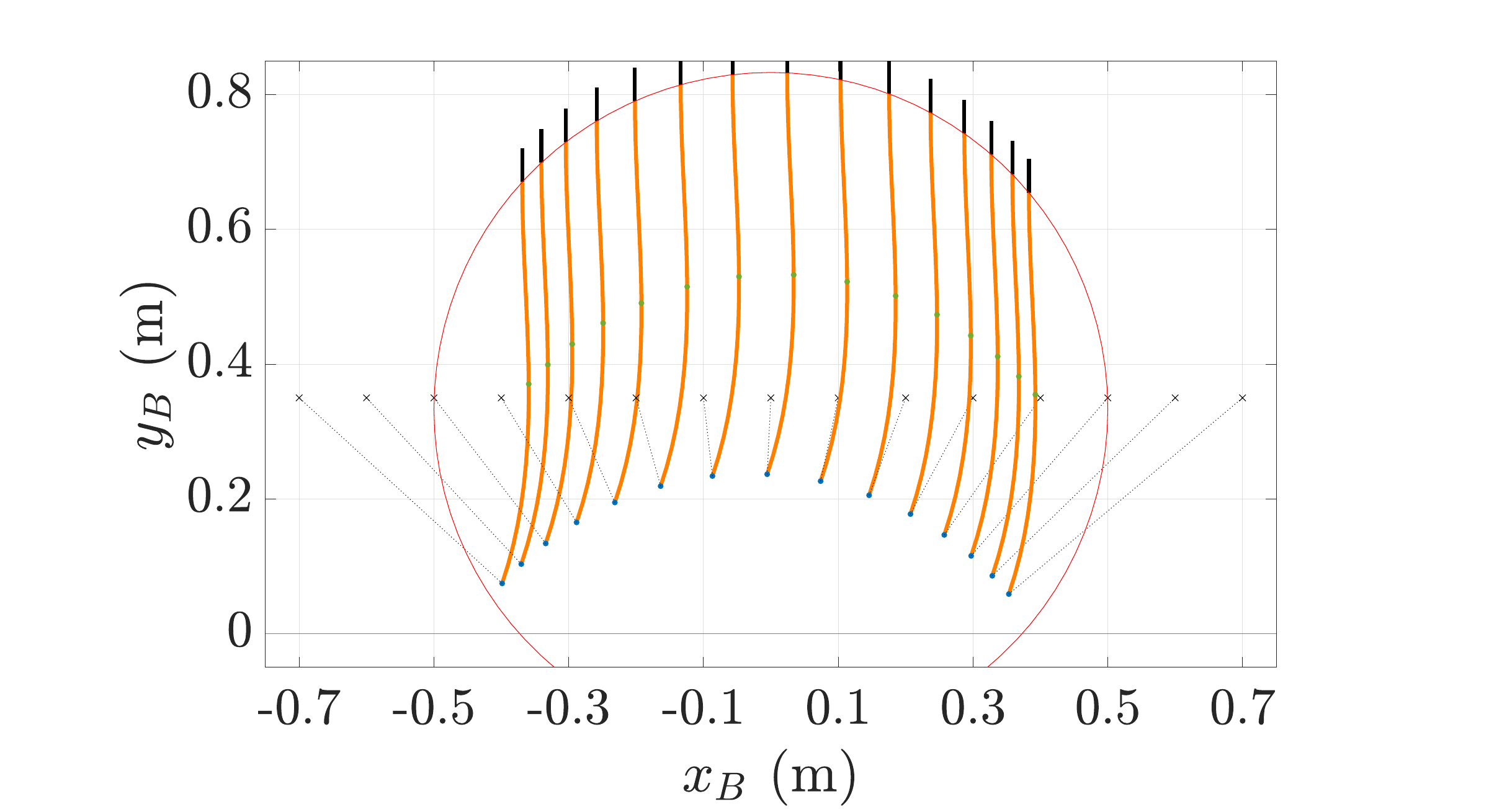} 
\hfill
\includegraphics[trim={7.2cm 3.9cm 5.7cm 0.7cm},width=0.46\columnwidth,clip]{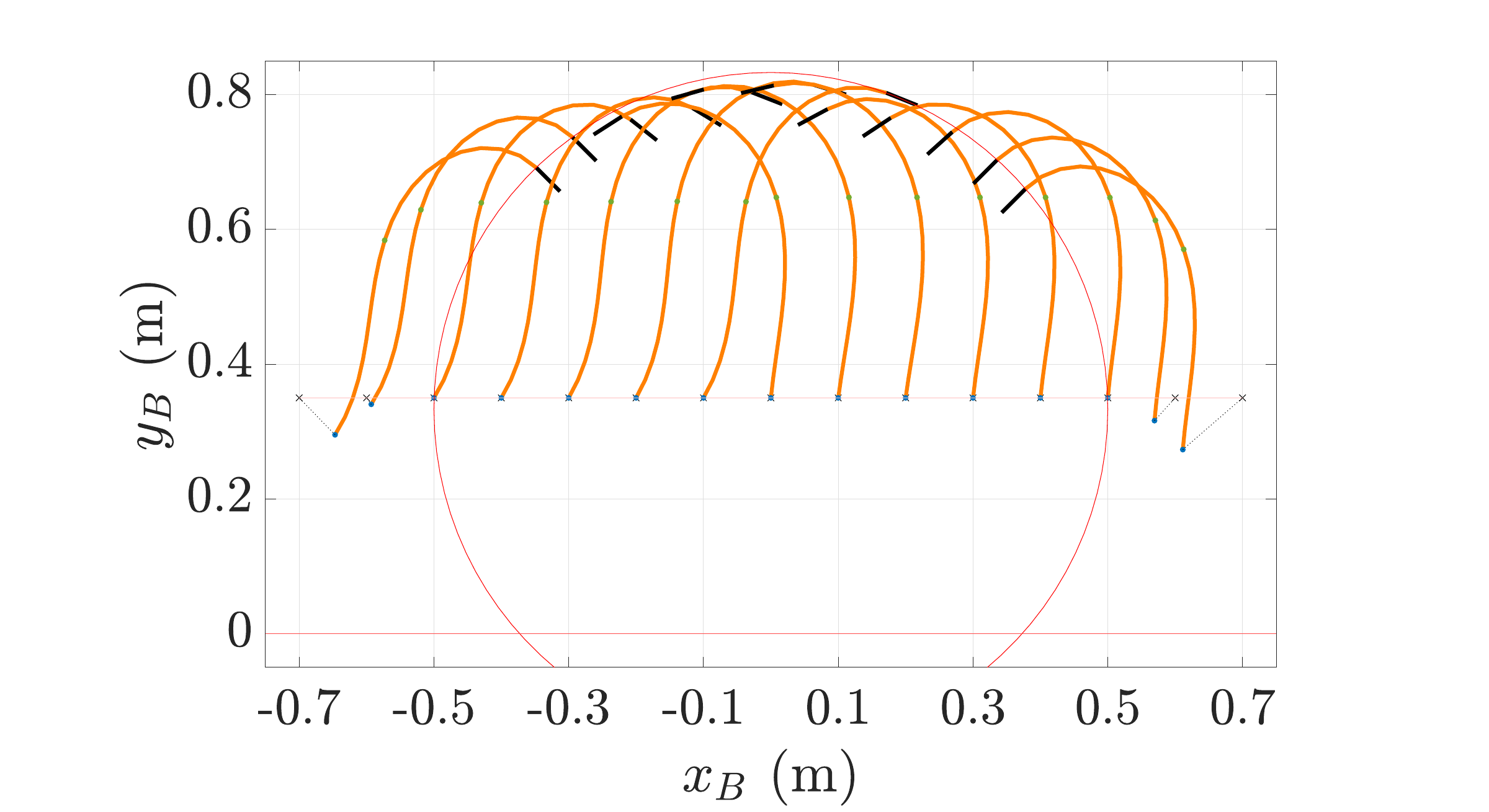}
\includegraphics[trim={3.4cm 3.9cm 5.7cm 0.7cm},width=0.525\columnwidth,clip]{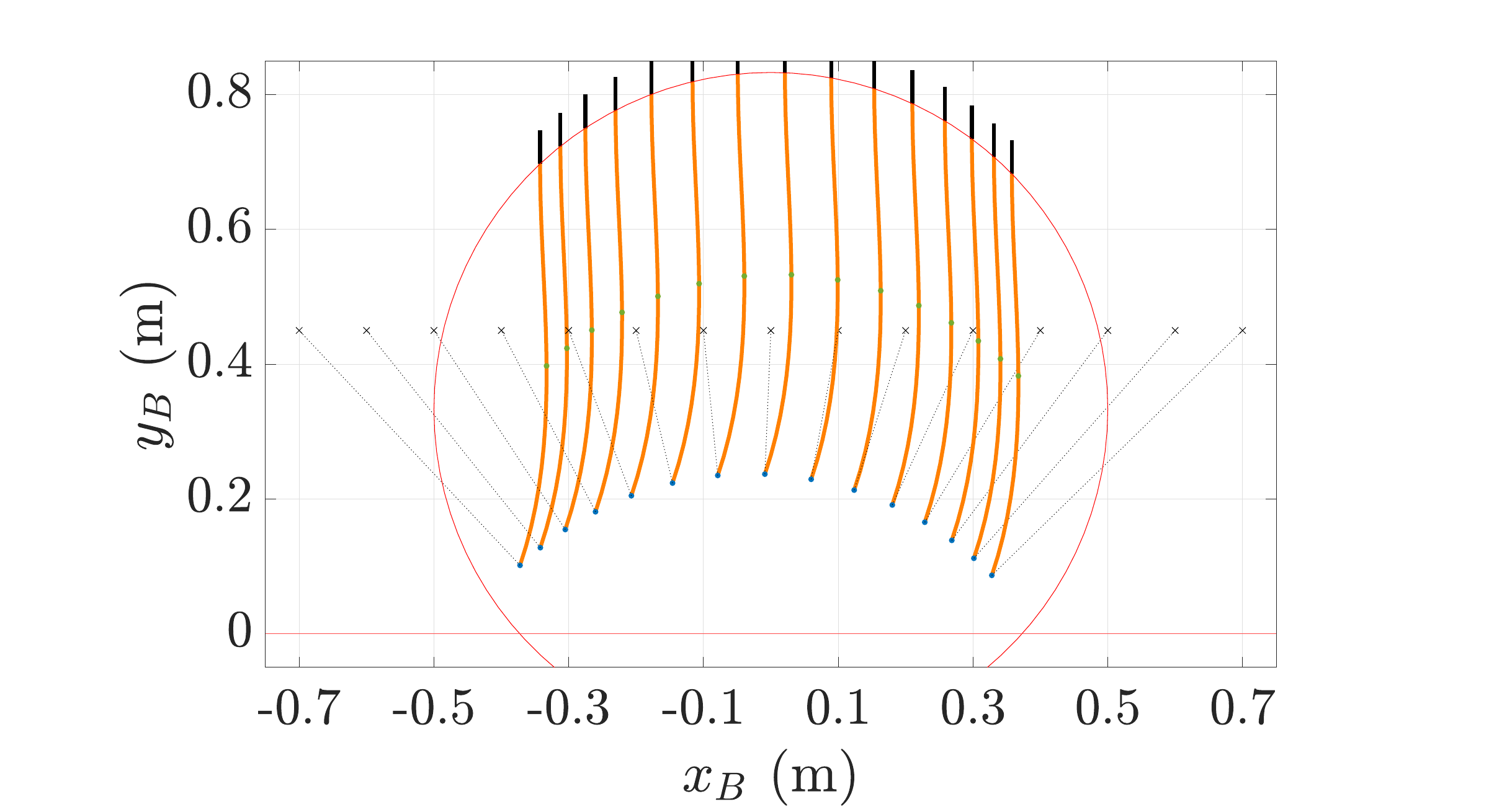} 
\hfill
\includegraphics[trim={7.2cm 3.9cm 5.7cm 0.7cm},width=0.46\columnwidth,clip]{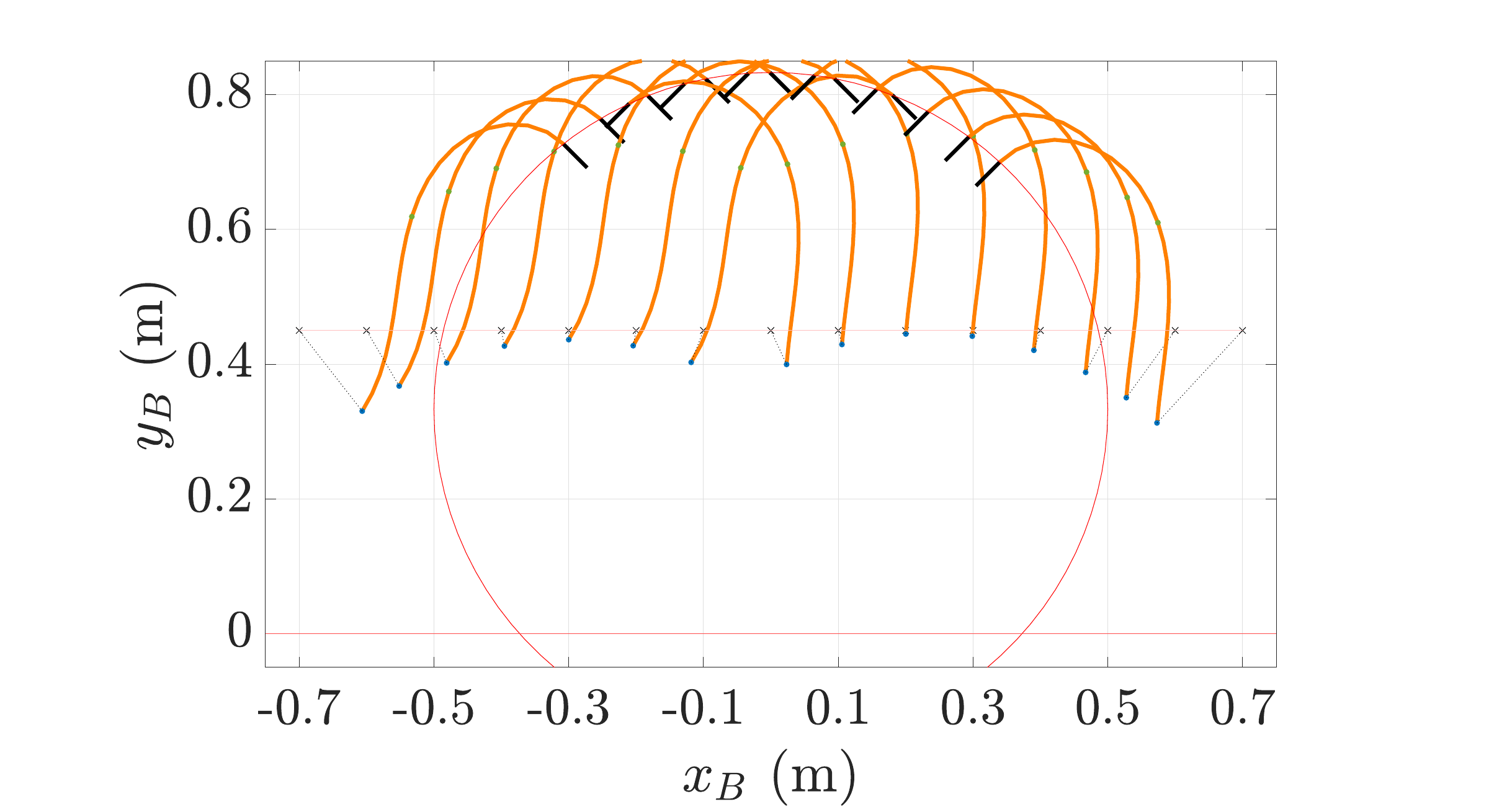}
\includegraphics[trim={3.4cm 0cm 5.7cm 0.7cm},width=0.525\columnwidth,clip]{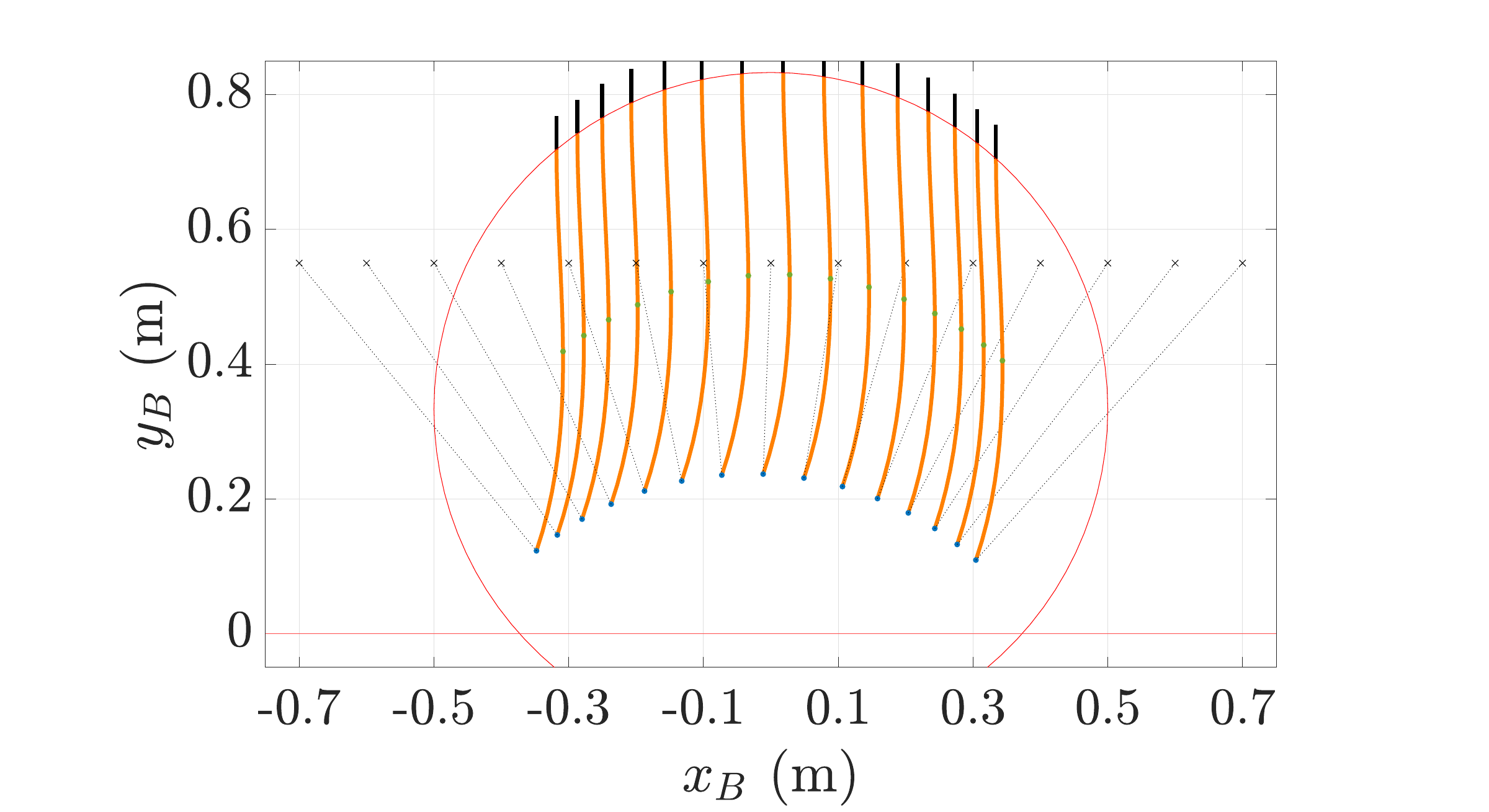} 
\hfill
\includegraphics[trim={7.2cm 0cm 5.7cm 0.7cm},width=0.46\columnwidth,clip]{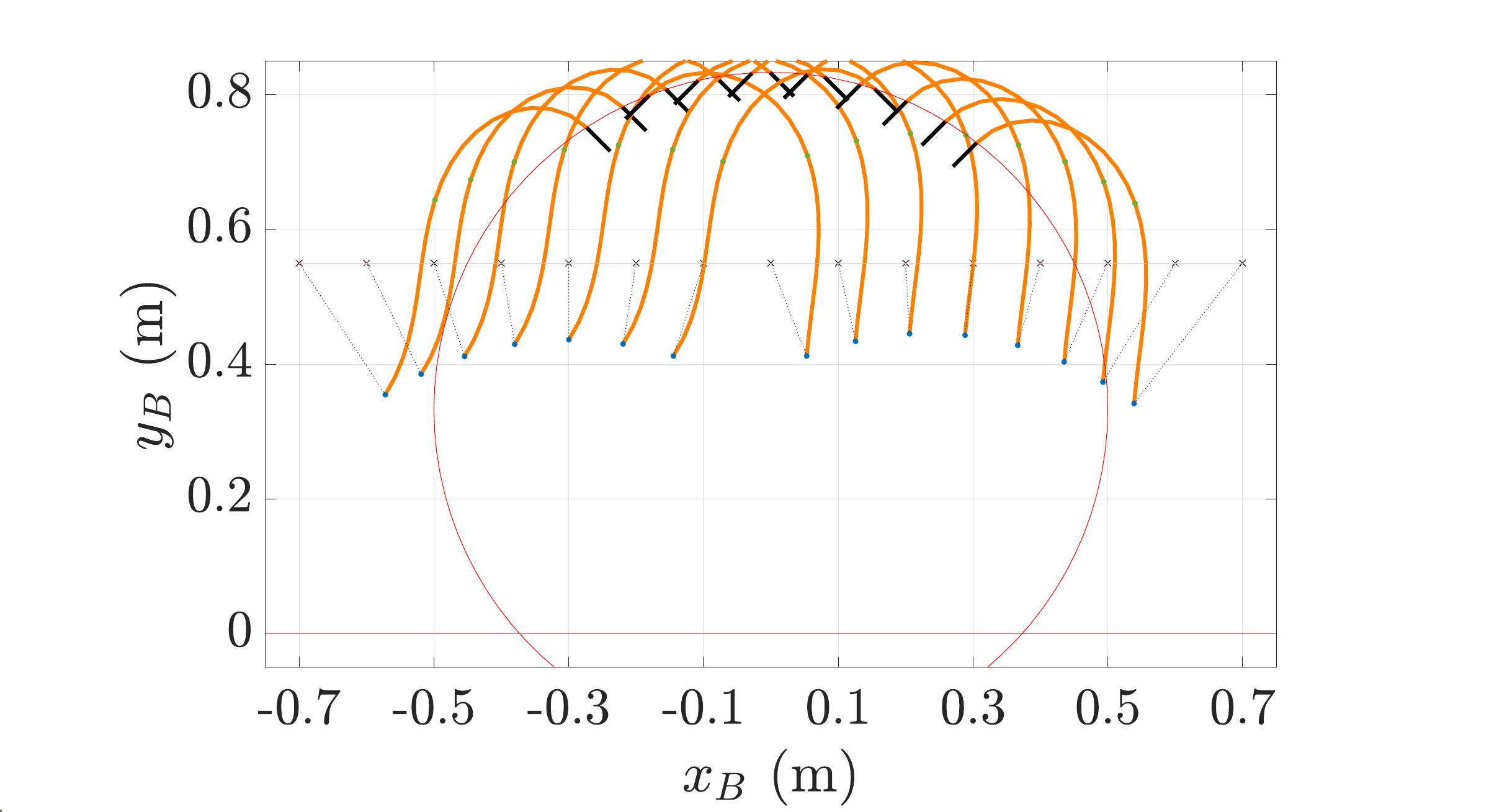}
\caption{Solutions generated by the optimization \eqref{eqn:position_NLO} 
for OB1 in the endpoint positioning experiment described in Section \ref{subsec:validation_position}. The model-free reference case is shown on the left, and the model-based case on the right. The endpoint goal positions are marked with small crosses, and lines are drawn to the achievable optimized solution.}
\label{fig:opt_sols}
\end{figure}

Based on these solutions, we assess the controller experimentally by measuring $p_e$ after moving the manipulator to each configuration, as demonstrated in the example image sequences in Fig. \ref{fig:series}. 

The results of these experiments carried out throughout the whole grid are summarized in Fig. \ref{fig:heatmaps}, where the grid colors correspond to the error between the goal $p^*$, located at the center of each cell, and $p_e$.

\begin{figure*}[t]
\centering
\begin{subfigure}{\textwidth}
    \includegraphics[width=0.195\textwidth,clip]{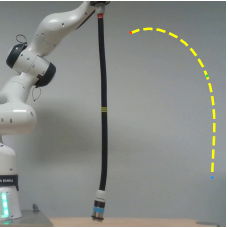}
    \includegraphics[width=0.195\textwidth,clip]{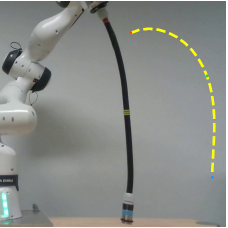}
    \includegraphics[width=0.195\textwidth,clip]{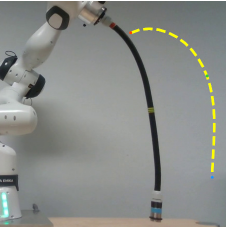}
    \includegraphics[width=0.195\textwidth,clip]{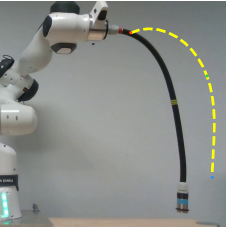}
    \includegraphics[width=0.195\textwidth,clip]{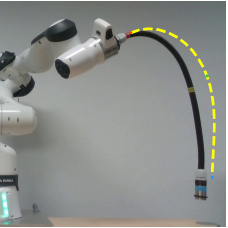}
    \caption{OB1}
\end{subfigure}
\begin{subfigure}{\textwidth}
    \includegraphics[width=0.195\textwidth,clip]{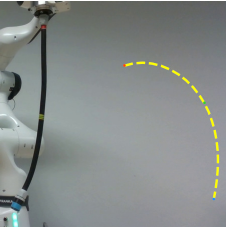}
    \includegraphics[width=0.195\textwidth,clip]{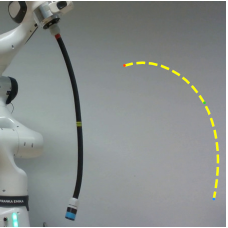}
    \includegraphics[width=0.195\textwidth,clip]{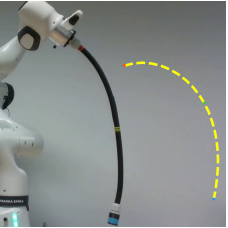}
    \includegraphics[width=0.195\textwidth,clip]{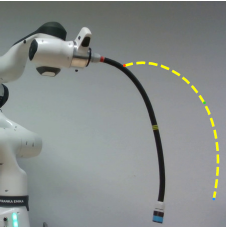}
    \includegraphics[width=0.195\textwidth,clip]{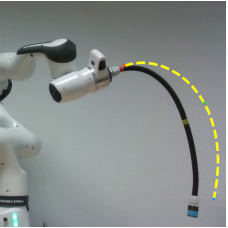}
    \caption{OB2}
\end{subfigure}
\begin{subfigure}{\textwidth}
    \includegraphics[width=0.195\textwidth,clip]{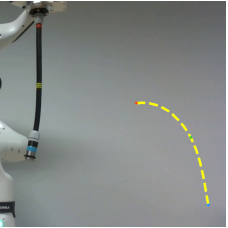}
    \includegraphics[width=0.195\textwidth,clip]{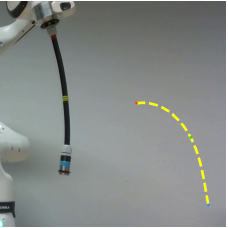}
    \includegraphics[width=0.195\textwidth,clip]{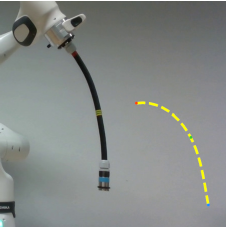}
    \includegraphics[width=0.195\textwidth,clip]{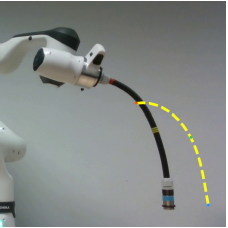}
    \includegraphics[width=0.195\textwidth,clip]{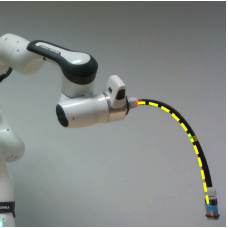}
    \caption{OB3}
\end{subfigure}
\begin{subfigure}{\textwidth}
    \includegraphics[width=0.195\textwidth,clip]{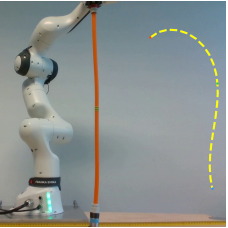}
    \includegraphics[width=0.195\textwidth,clip]{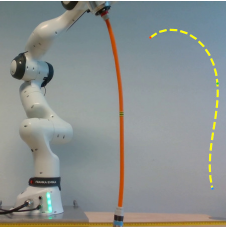}
    \includegraphics[width=0.195\textwidth,clip]{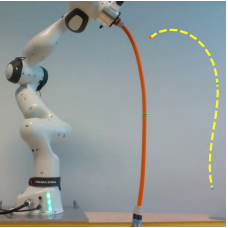}
    \includegraphics[width=0.195\textwidth,clip]{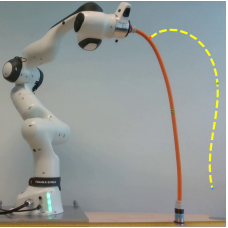}
    \includegraphics[width=0.195\textwidth,clip]{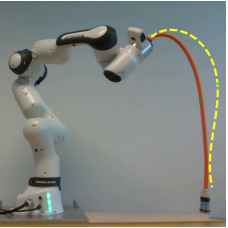}
    \caption{OB5}
\end{subfigure}
\caption{Example image sequences from the endpoint position control validation experiment in Section \ref{subsec:validation_position}, showing the manipulator moving to the desired object configurations corresponding to target endpoint goal locations. The dashed overlays indicate the intended final object goal states generated by the model-constrained optimization \eqref{eqn:position_NLO}.}
\label{fig:series}
\end{figure*}

For all objects it is clear that the endpoint reachability is significantly expanded compared to the 'model-free' reference case, as expected. The measured positioning accuracy in this increased workspace varies significantly, generally corresponding to decreased accuracy with the use of larger $|\phi|$. 
The model-based case is clearly an improvement based on these demonstrations. Measuring the mean endpoint error over all measured grid points, there is a reduction in the measured mean error compared to the 'model-free' reference case of 63.0\%, 68.1\%, 78.5\% and 57.7\% for the four objects respectively.

\begin{figure}%
\centering
    \begin{subfigure}{\columnwidth}
        \includegraphics[width=\textwidth,clip]{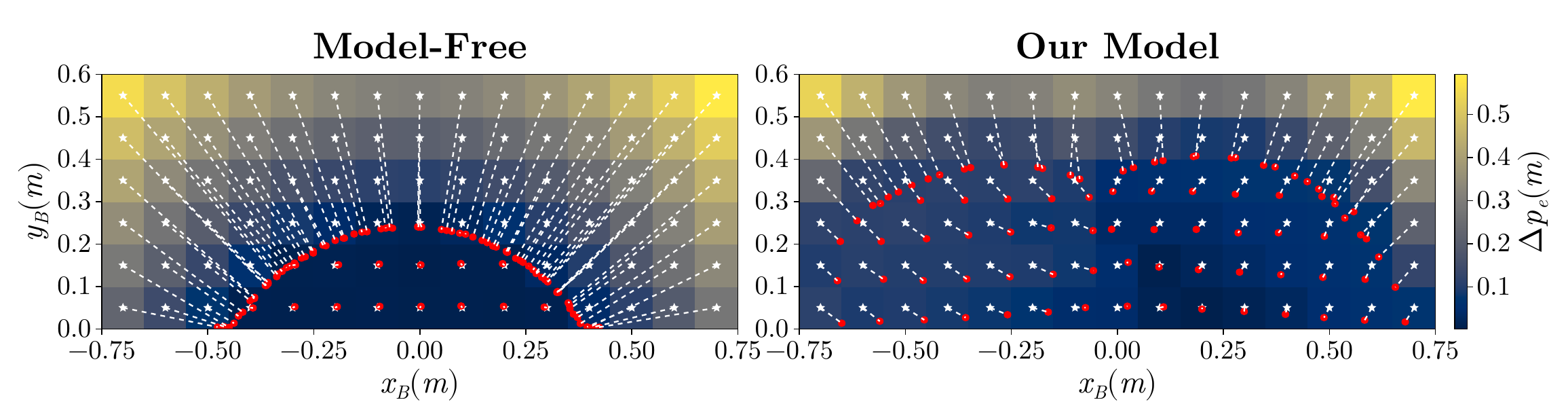}
    \caption{OB1}
    \end{subfigure}
    \begin{subfigure}{\columnwidth}
        \includegraphics[width=\textwidth,clip]{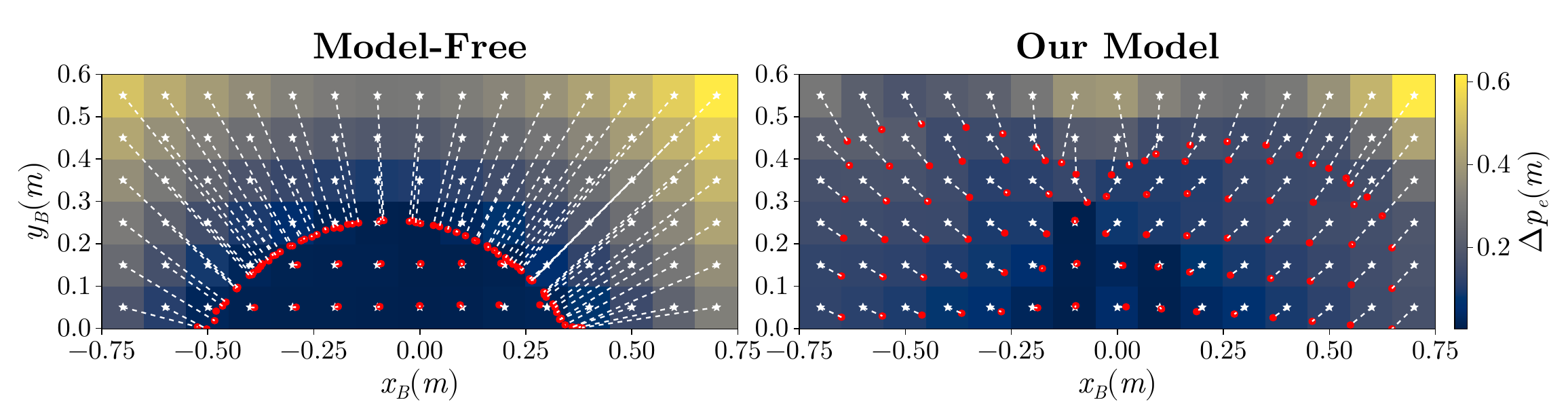}
    \caption{OB2}
    \end{subfigure}
    \begin{subfigure}{\columnwidth}
        \includegraphics[width=\textwidth,clip]{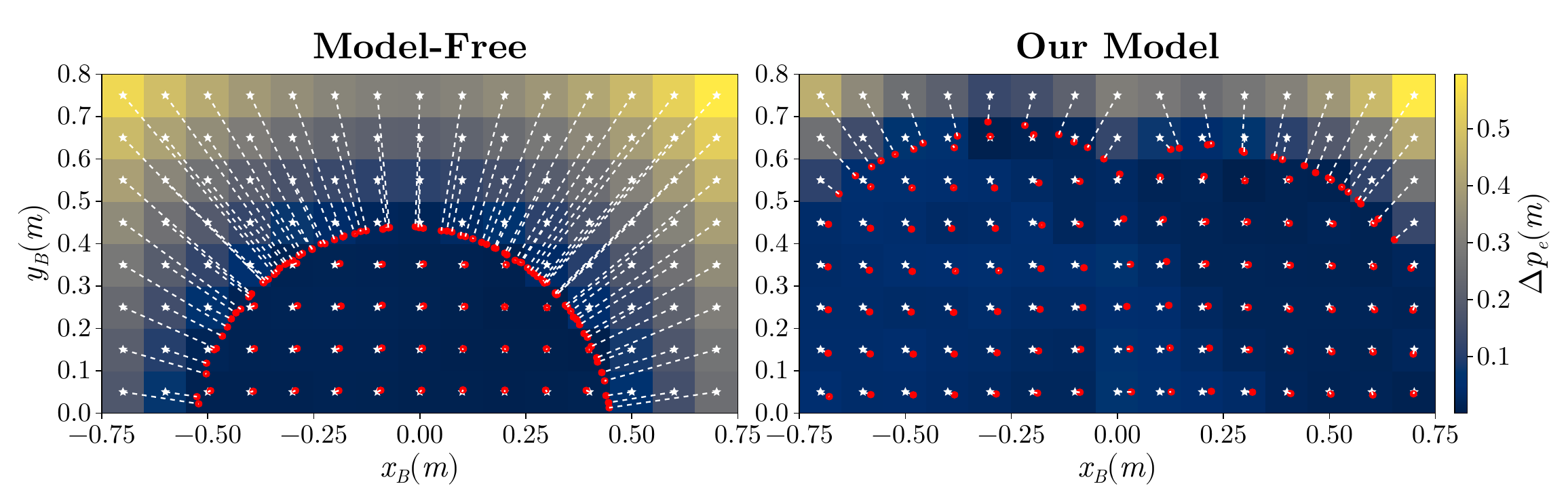}
    \caption{OB3}
    \end{subfigure}
    \begin{subfigure}{\columnwidth}
        \includegraphics[width=\textwidth,clip]{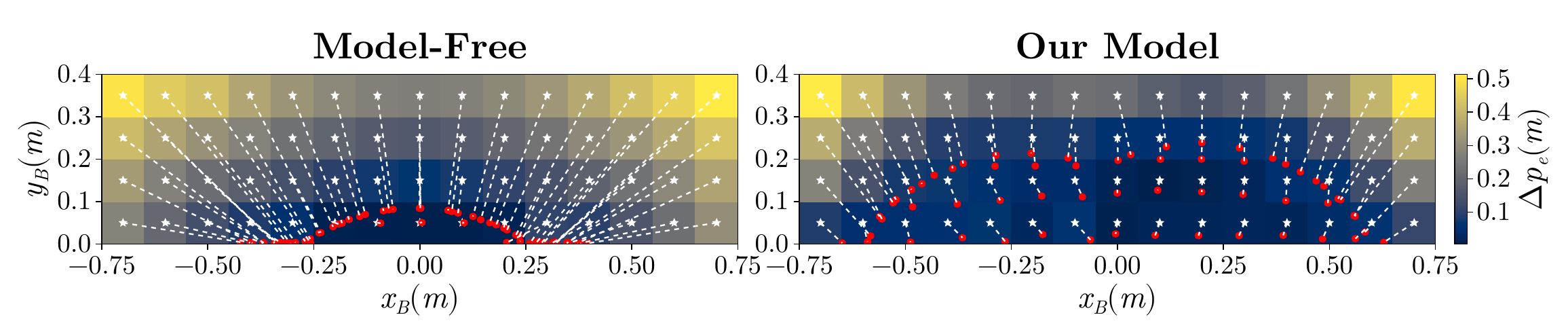}
    \caption{OB5}
    \end{subfigure}
    \caption{Visualization of the endpoint positioning accuracy over the workspace for OB1, OB2, OB3 and OB5. The plots (a) - (d), show the results of experiments to compare the model-free strategy and our model-based controller with shape regulation. The centre of each cell was the goal of the controller $p^*$ (white star). The intensity of the cell color indicates the error for that goal $p_e$. The red dot shows the measured position of the objects's endpoint. A white dashed line is drawn between each goal and the measured point
}
\label{fig:heatmaps}
\end{figure}

\begin{figure}[p]
\centering
    \begin{subfigure}{\columnwidth}
        \includegraphics[trim={8.7cm 0 8cm 0.8cm},width=0.525\textwidth,clip]{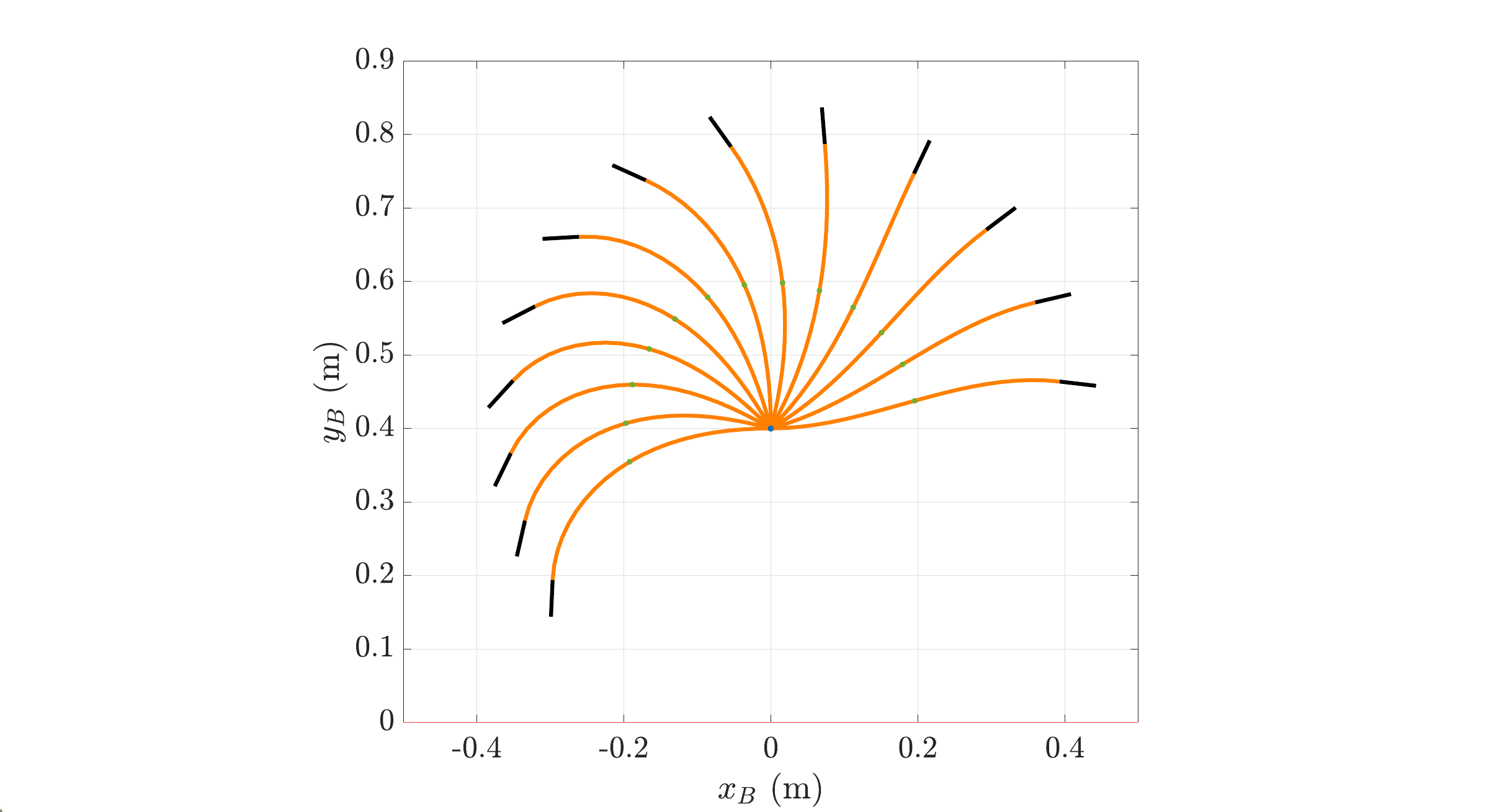}
        \hfill
        \includegraphics[trim={0cm 2cm 0cm 0cm},width=0.455\textwidth,clip]{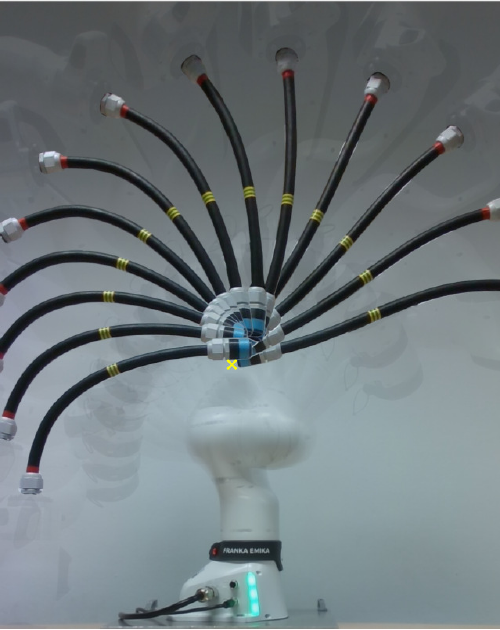}
        \includegraphics[trim={8.7cm 0 8cm 0.8cm},width=0.525\textwidth,clip]{images/orientation_analysis/black_orientation_solutions-eps-converted-to.pdf}
        \hfill
        \includegraphics[trim={0cm 2cm 0cm 0cm},width=0.455\textwidth,clip]{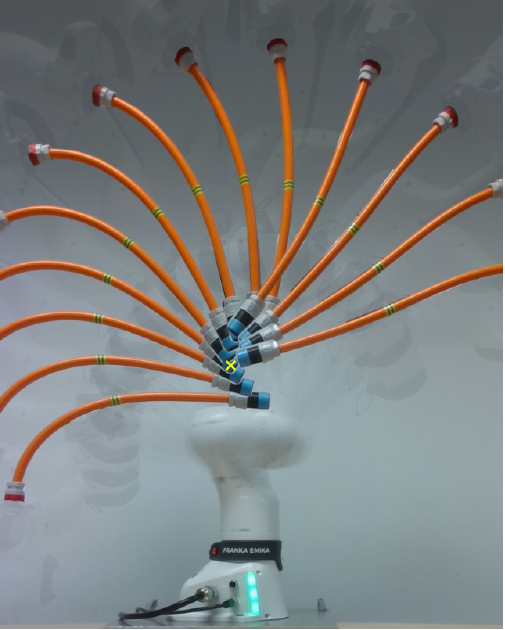}
    \caption{OB4 top, OB6 bottom}
    \end{subfigure}
    \begin{subfigure}{\columnwidth}
        \includegraphics[trim={1cm 3.5cm 3cm 0},width=\textwidth,clip]{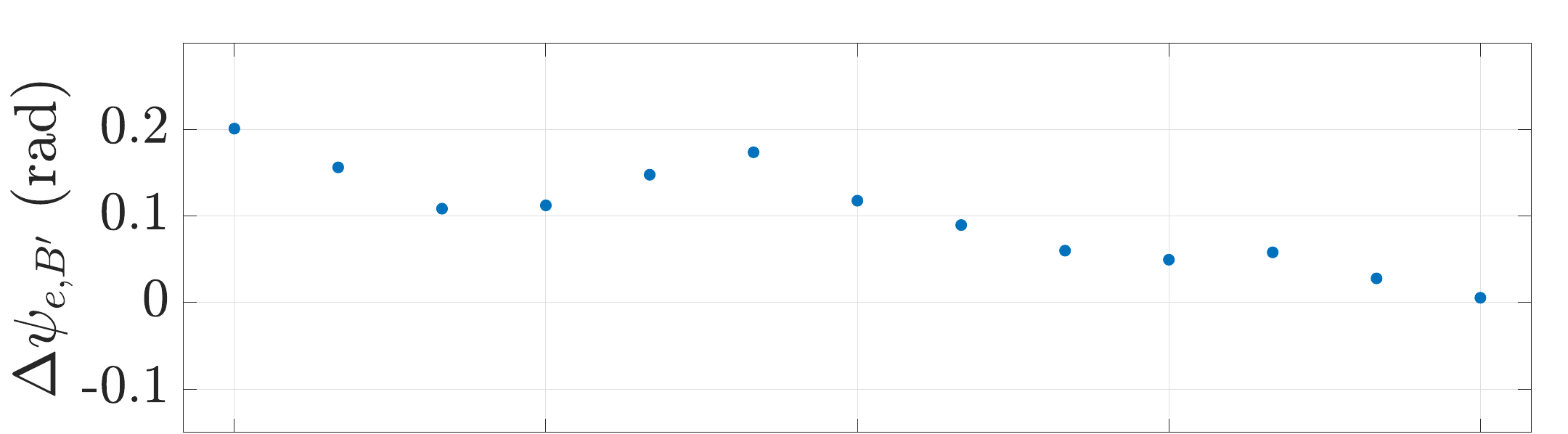}
        \includegraphics[trim={1cm 0 3cm 0},width=\textwidth,clip]{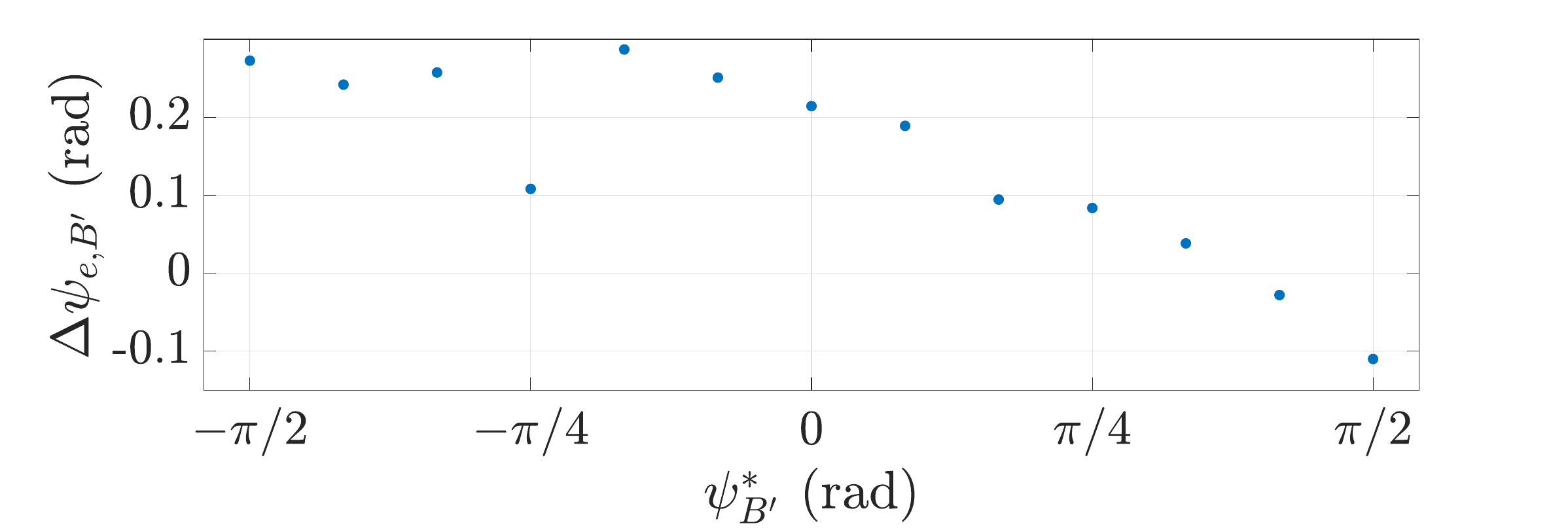}
    \caption{OB4 top, OB6 bottom}
    \end{subfigure}
    \begin{subfigure}{\columnwidth}
        \includegraphics[trim={1cm 0cm 3cm 0},width=\textwidth,clip]{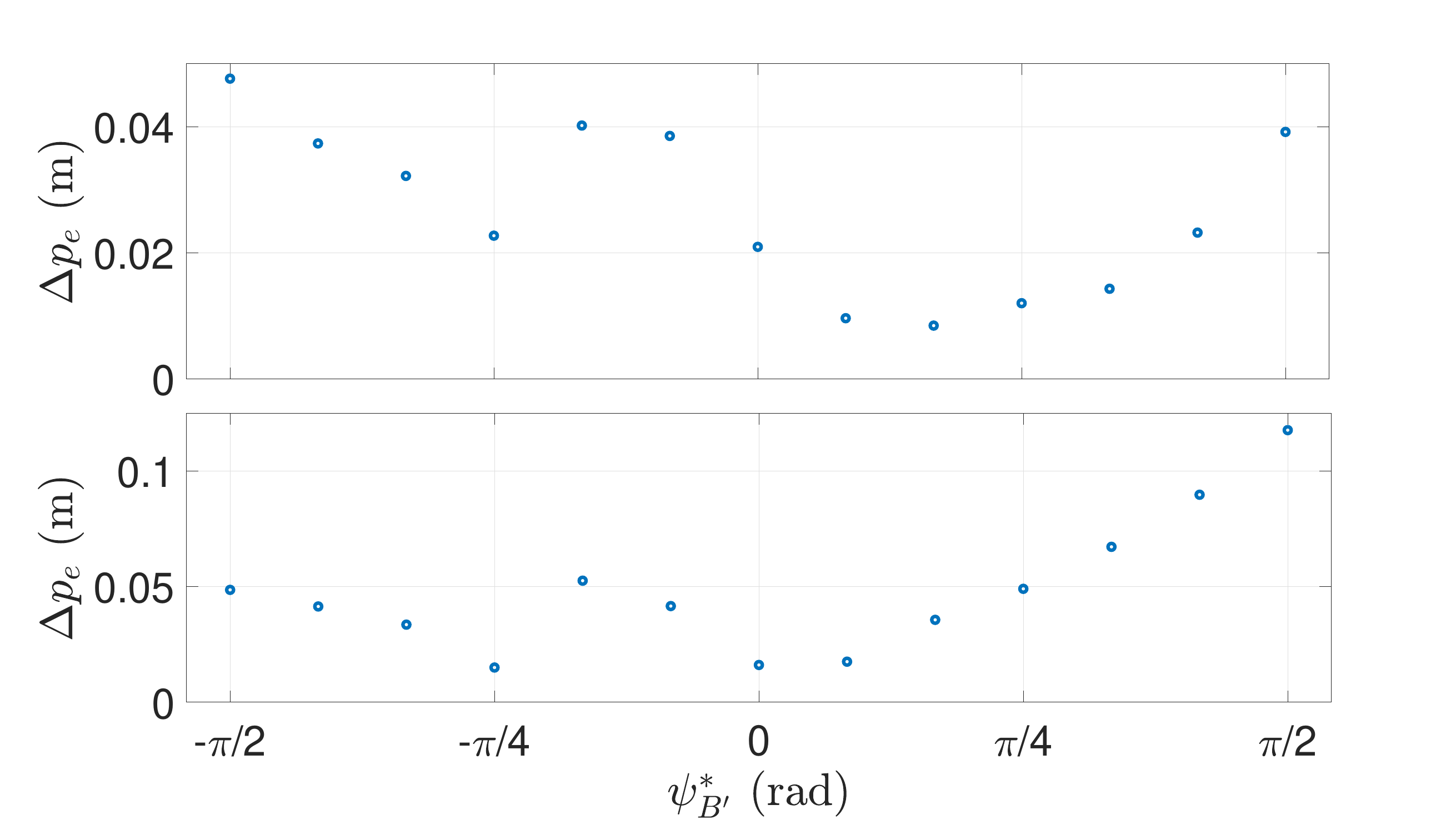}
    \caption{OB4 top, OB6 bottom}
    \end{subfigure}
    
\caption{(a) Solutions generated by the optimization \eqref{eqn:orientation_NLO} for the endpoint orientation control experiment in Section \ref{subsec:validation_orientation}, and composites of the corresponding real object states. The yellow cross indicates the intended endpoint position. (b) Error between the goal and measured object tip angles. (c) Position errors between the goal and measured object tip position.}
\label{fig:black_orientation_solutions_and_composite}
\end{figure}

\subsection{Orientation}\label{subsec:validation_orientation}

Here we continue assessing the reachable workspace of the object's endpoint, now focussing only on the orientation, ie. the accuracy of positioning $p_e$ at a single location in $\{S_B\}$ but with the tip pointed in a specified angle. We will refer to the angle of the tip as $\psi_{e,B'} = \phi + \alpha_e$, note that this is in the $\{S_{B'}\}$ frame so that pointing directly downwards corresponds to 0. For this validation we only investigate OB4 and OB6, as inspection of the static identification results indicated a very limited achievable tip angle range for the other objects, due to their length and weight.

The experimental process is similar to that in Section \ref{subsec:validation_position}, now with $\zeta = (p_e,\psi_{e,B'})$ requiring the addition of the endpoint orientation error into the optimization cost. We simplify by ensuring the endpoint position goal $p^*$ is always exactly reachable by removing workspace position constraints on the manipulator, and only target endpoint orientation goals $\psi_{B'}^*$ away from  $\pm \pi$, so it is sufficient to define $d$ as the 2-norm of the combined position and angle errors. We relax the constraint on the manipulator angle to $\frac{-3\pi}{4} \leq \phi < \pi$, which allows a goal range of $\frac{-\pi}{2} \leq \psi_{B'}^* \leq \frac{\pi}{2}$. Hence the optimization problem here is:

\begin{equation}\label{eqn:orientation_NLO}
\begin{split}
    \min_{\Theta^* \in \mathbb{R}^{n+1},(x^*,y^*,\phi^*)\in\mathbb{F}}
    \quad 
    &\begin{Vmatrix}
        p^*-p_e(x^*,y^*,\phi^*,\Theta^*) \\
        \psi_{B'}^*-\psi_{e,B'}(\phi^*,\Theta^*)
    \end{Vmatrix}_2\\
    \mathrm{s.t.} \quad G_{\Theta}(\Theta^*,\phi^*) + &kH{(\Theta^*-\bar{\Theta})} = 0\\
    \mathbb{F} = \{\frac{-3\pi}{4} &\leq \phi \leq \pi\}.
\end{split}
\end{equation}

\begin{figure*}[t]
\centering
\begin{subfigure}{0.195\textwidth}
    \includegraphics[trim={9.3cm -0.5cm 10.8cm 0cm},width=\textwidth,clip]{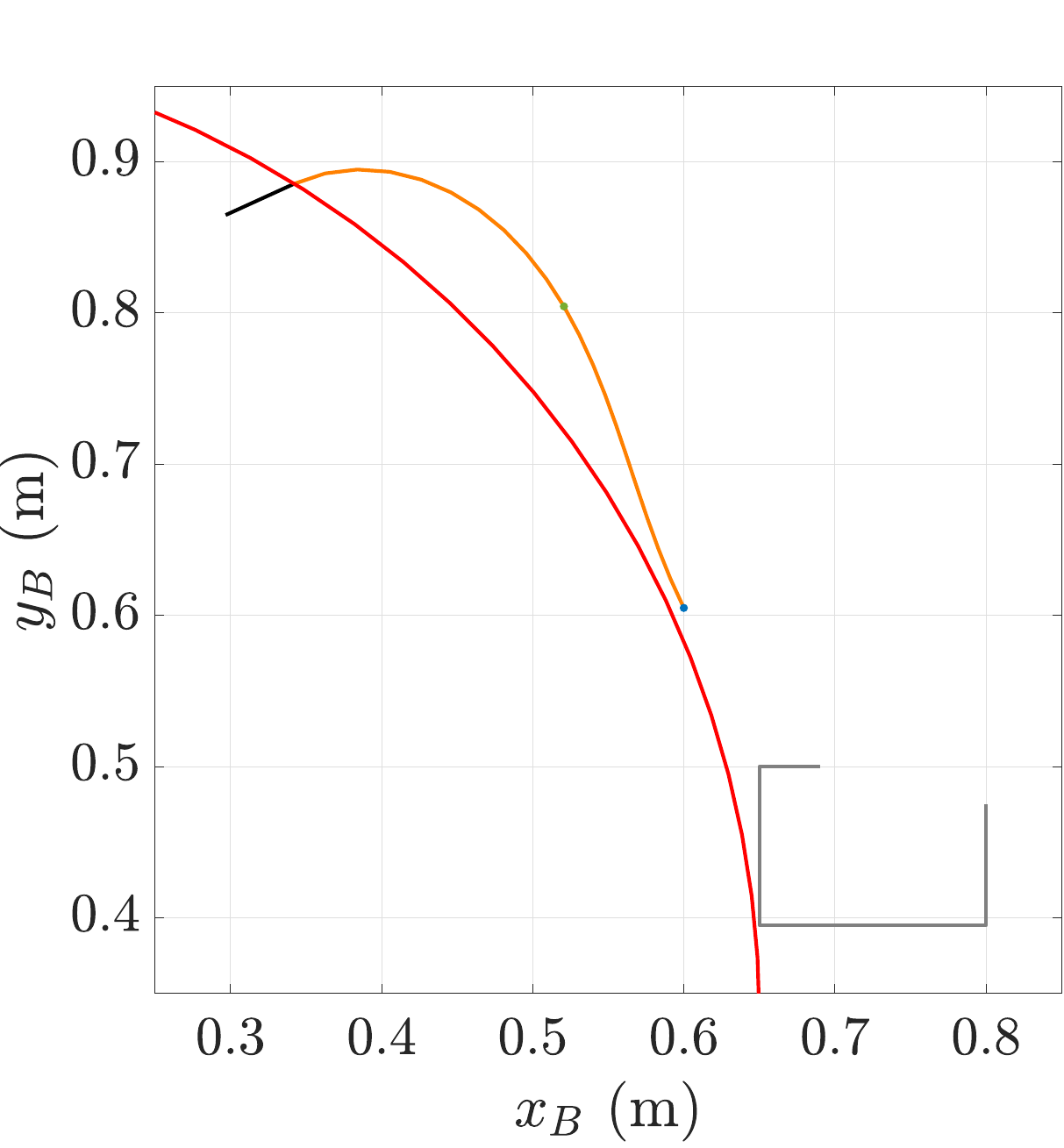}
\end{subfigure}
\begin{subfigure}{0.195\textwidth}
    \includegraphics[trim={9.3cm -0.5cm 10.8cm 0cm},width=\textwidth,clip]{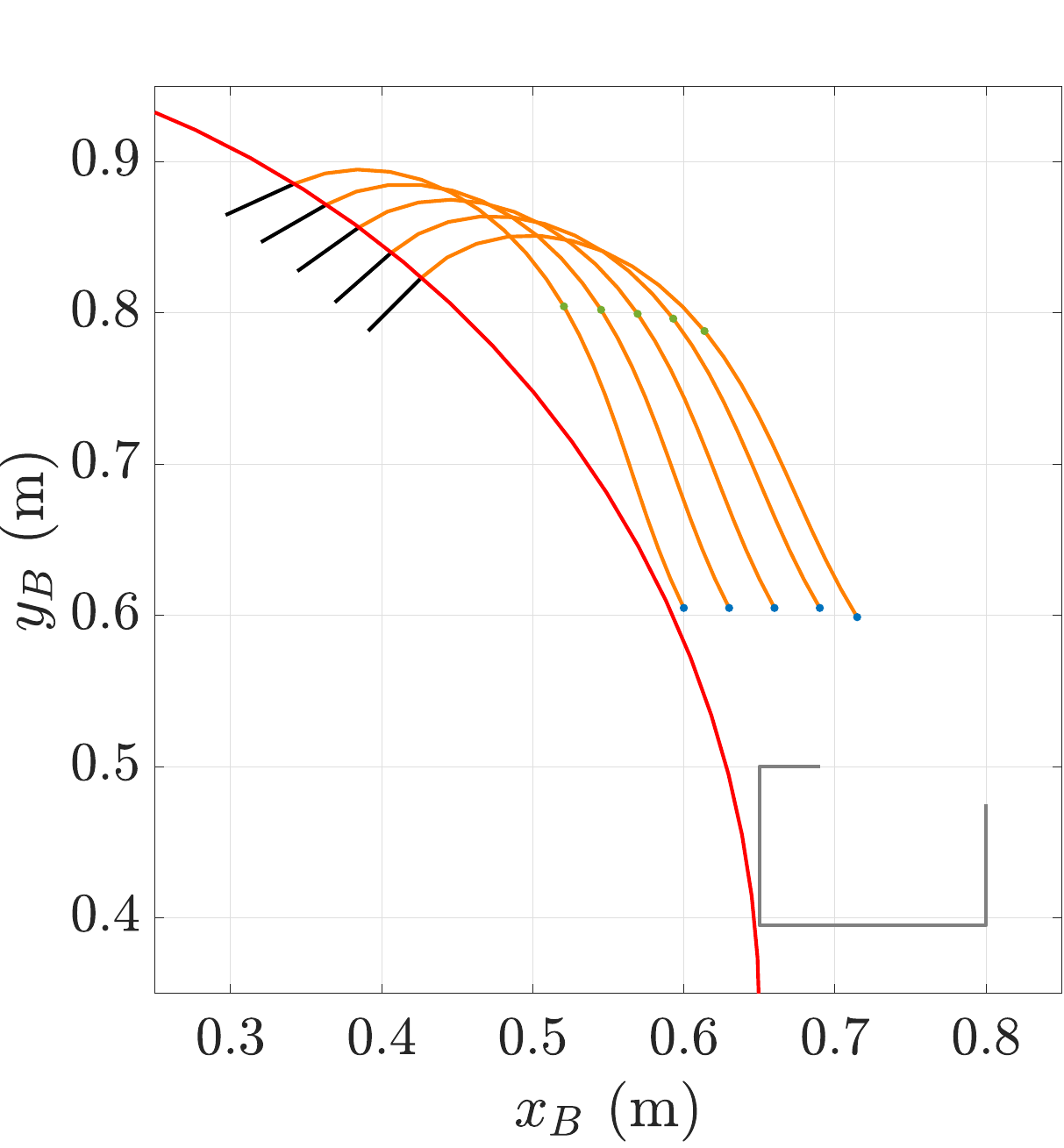}
\end{subfigure}
\begin{subfigure}{0.195\textwidth}
    \includegraphics[trim={9.3cm -0.5cm 10.8cm 0cm},width=\textwidth,clip]{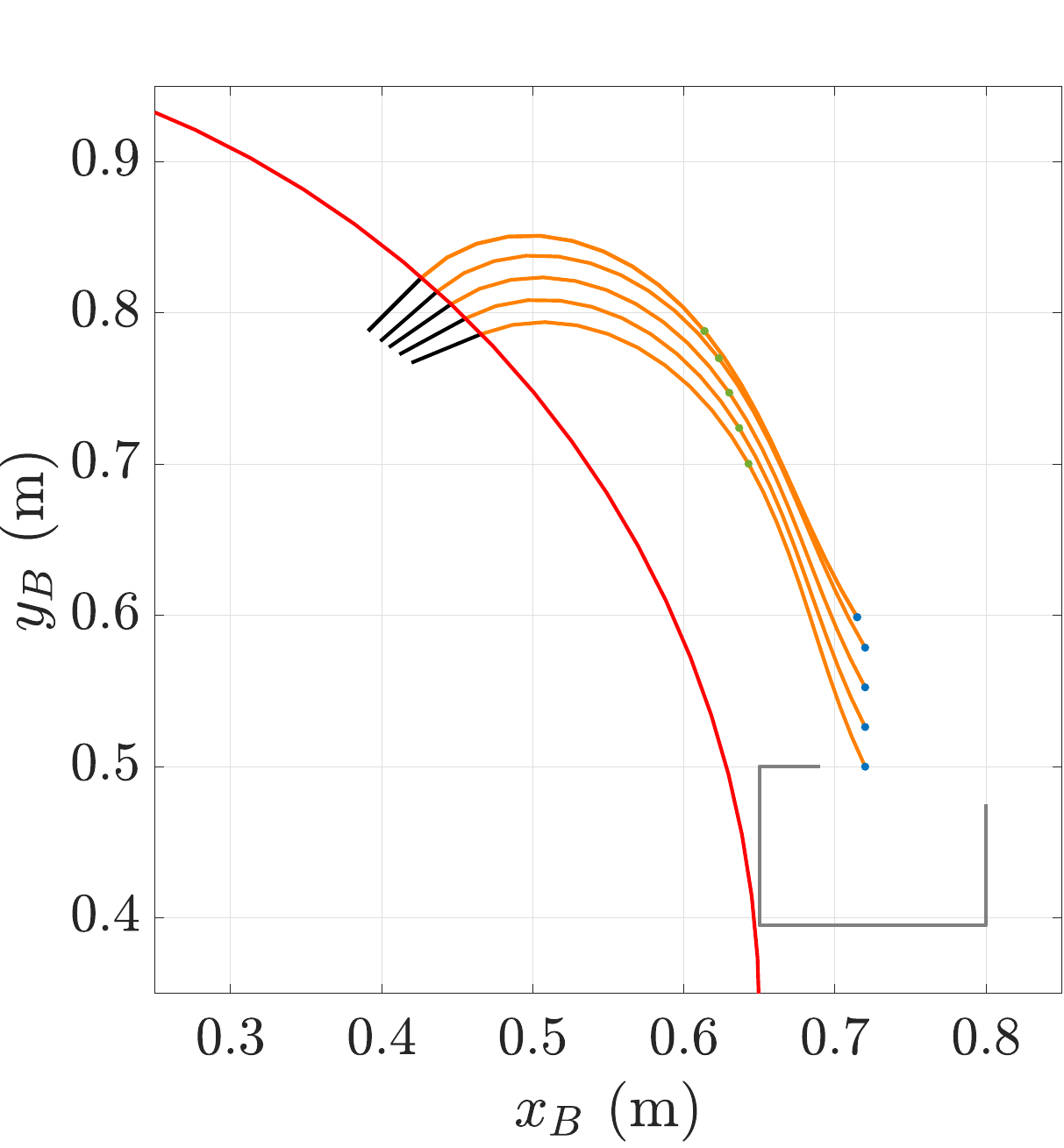}
\end{subfigure}
\begin{subfigure}{0.195\textwidth}
    \includegraphics[trim={9.3cm -0.5cm 10.8cm 0cm},width=\textwidth,clip]{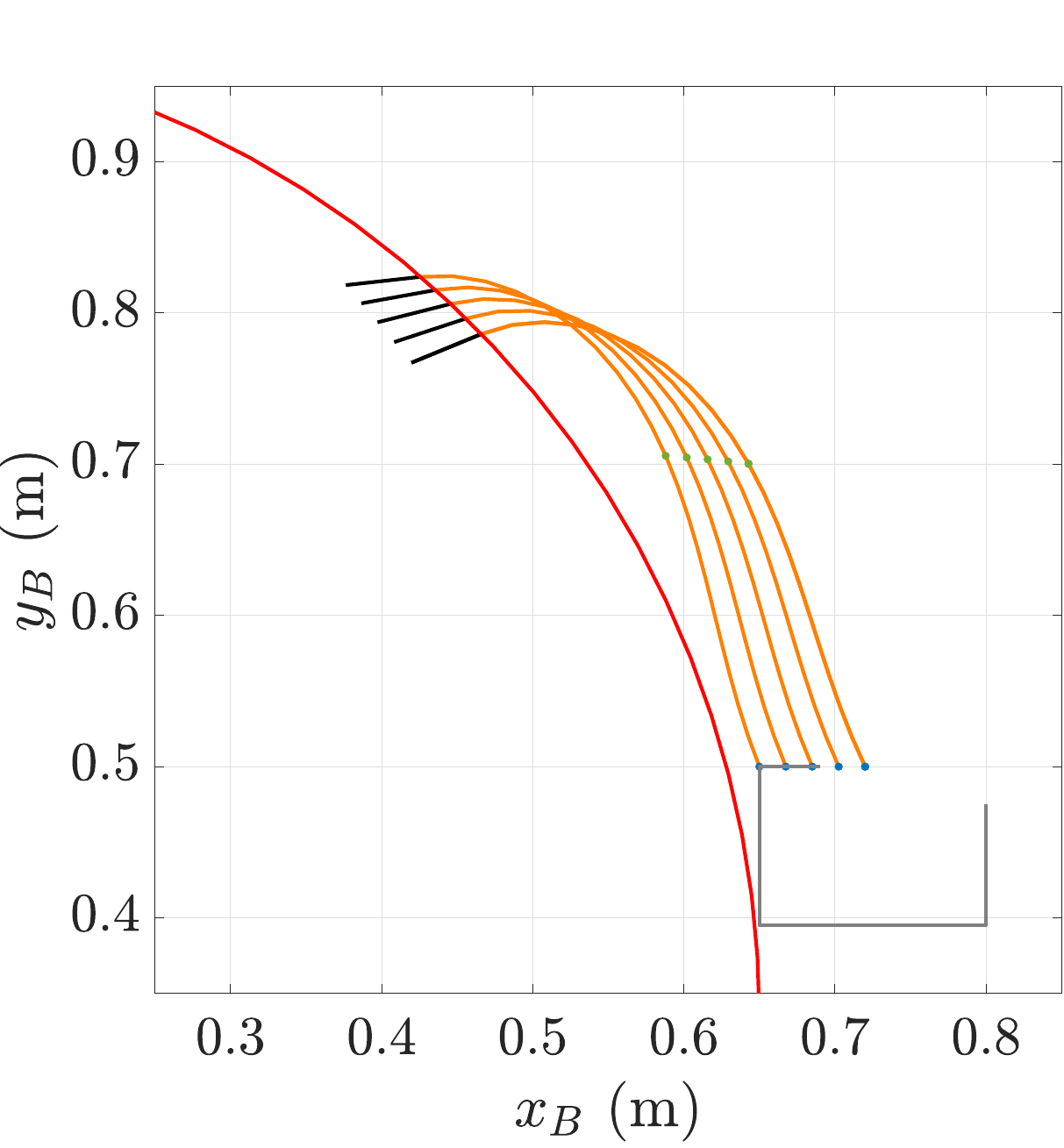}
\end{subfigure}
\begin{subfigure}{0.195\textwidth}
    \includegraphics[trim={9.3cm -0.5cm 10.8cm 0cm},width=\textwidth,clip]{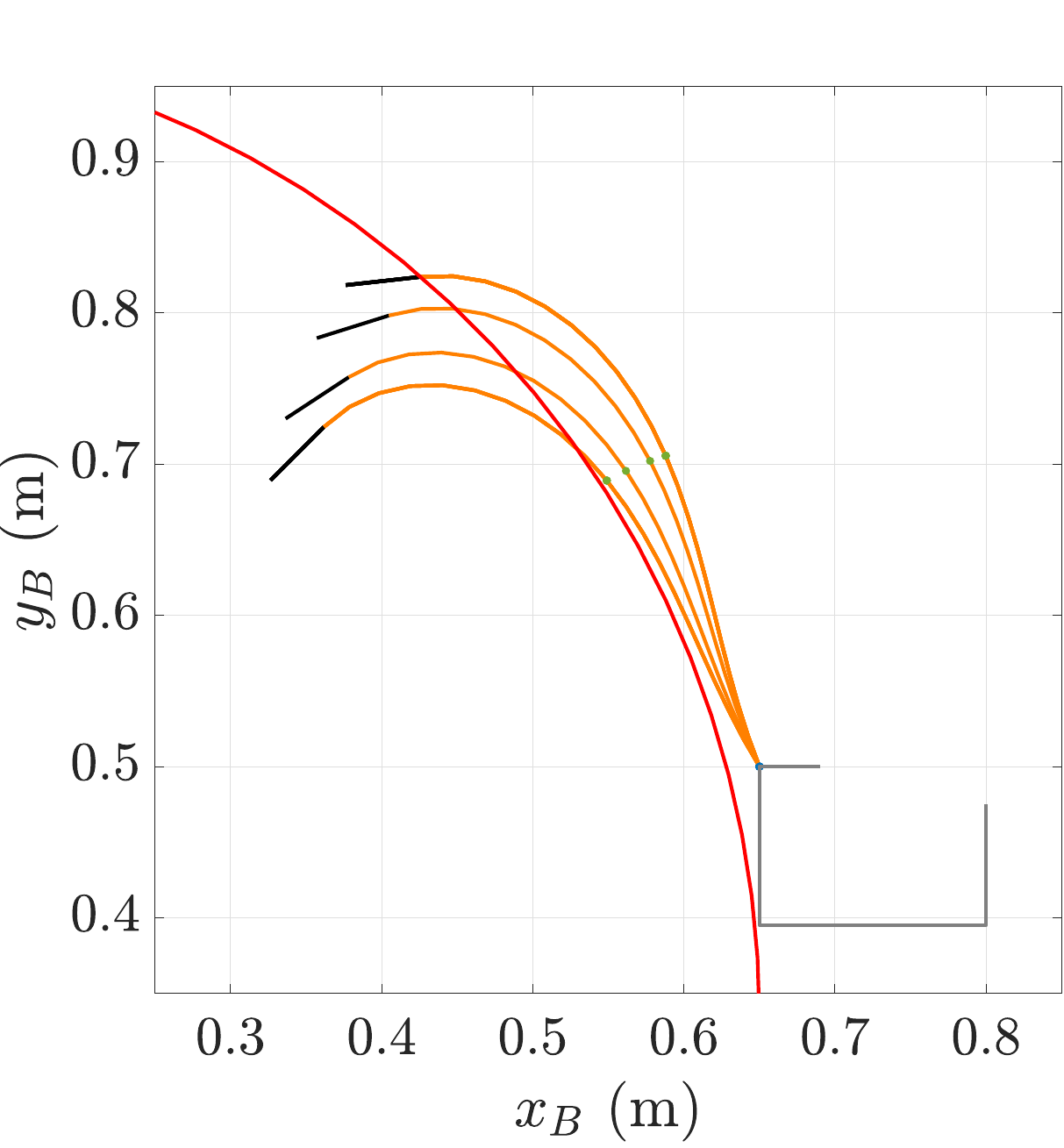}
\end{subfigure}

\begin{subfigure}{0.195\textwidth}
    \includegraphics[width=\textwidth,clip]{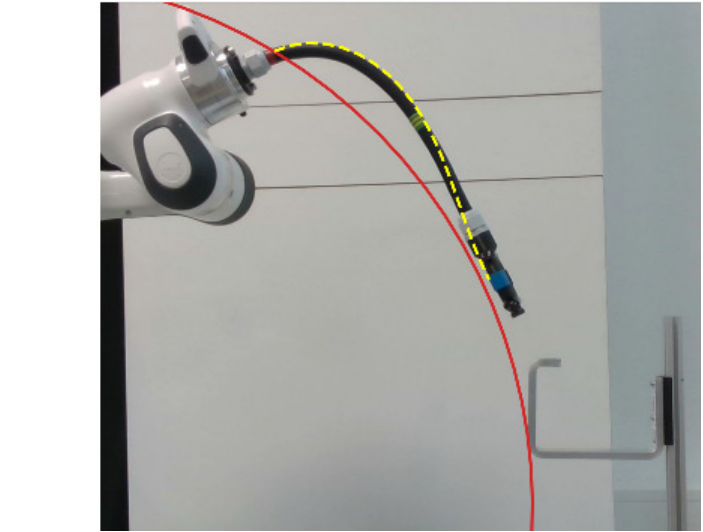}
\end{subfigure}
\begin{subfigure}{0.195\textwidth}
    \includegraphics[width=\textwidth,clip]{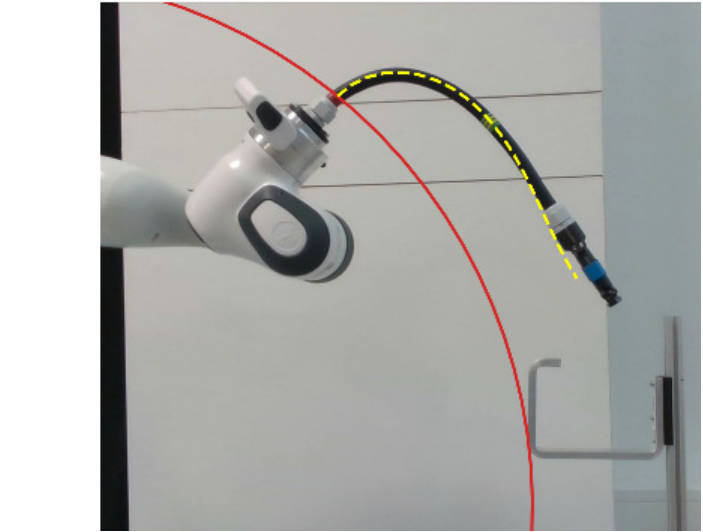}
\end{subfigure}
\begin{subfigure}{0.195\textwidth}
    \includegraphics[width=\textwidth,clip]{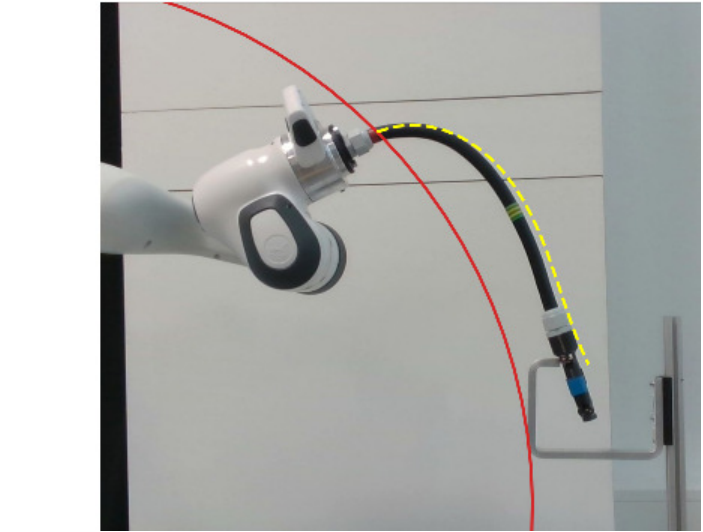}
\end{subfigure}
\begin{subfigure}{0.195\textwidth}
    \includegraphics[width=\textwidth,clip]{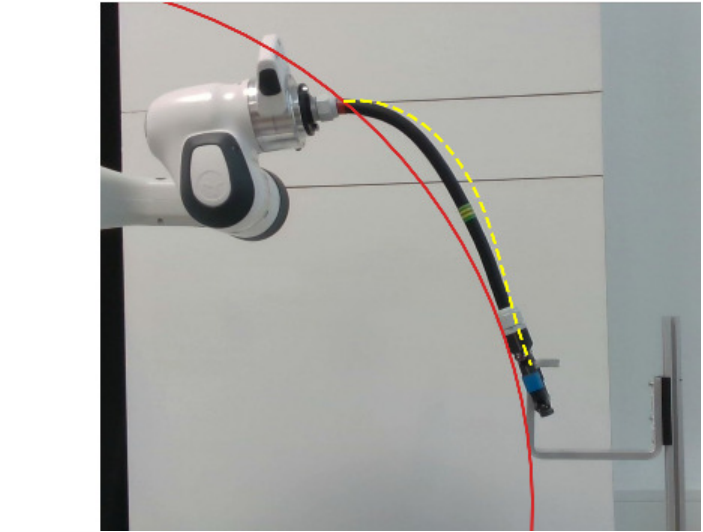}
\end{subfigure}
\begin{subfigure}{0.195\textwidth}
    \includegraphics[width=\textwidth,clip]{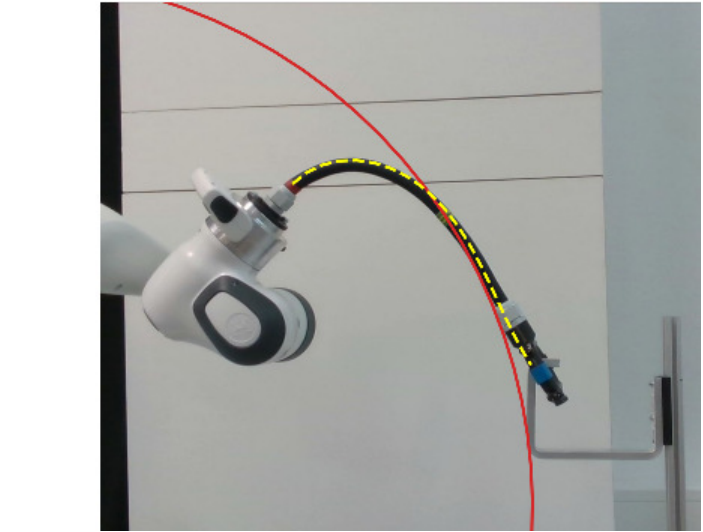}
\end{subfigure}

\caption{Results from the object hooking task demonstration. The base position and object configurations generated for a series of endpoint goal locations are shown above, with corresponding images from the experiment below, overlaid with the intended object configuration at that time, and the radial constraint on the manipulator endpoint.}
\label{fig:hook_demo}
\end{figure*}

\begin{figure*}[t]
\centering

\begin{subfigure}{0.215\textwidth}
    \includegraphics[trim={9cm 0cm 10.5cm 0cm},width=\textwidth,clip]{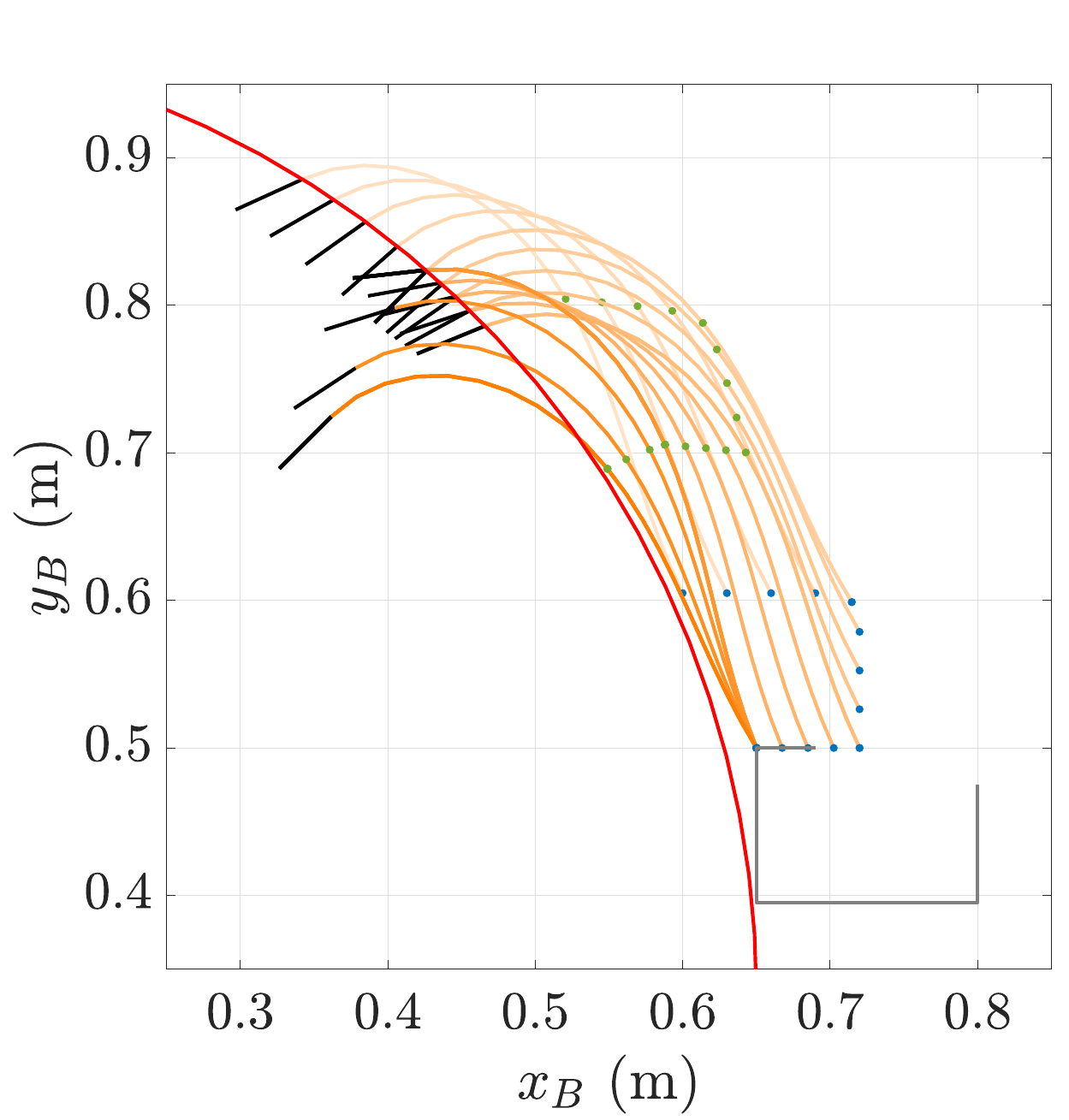}
    \caption{}
\end{subfigure}
\begin{subfigure}{0.4\textwidth}
    \includegraphics[trim={0cm 0cm 0cm 0cm},width=\textwidth,clip]{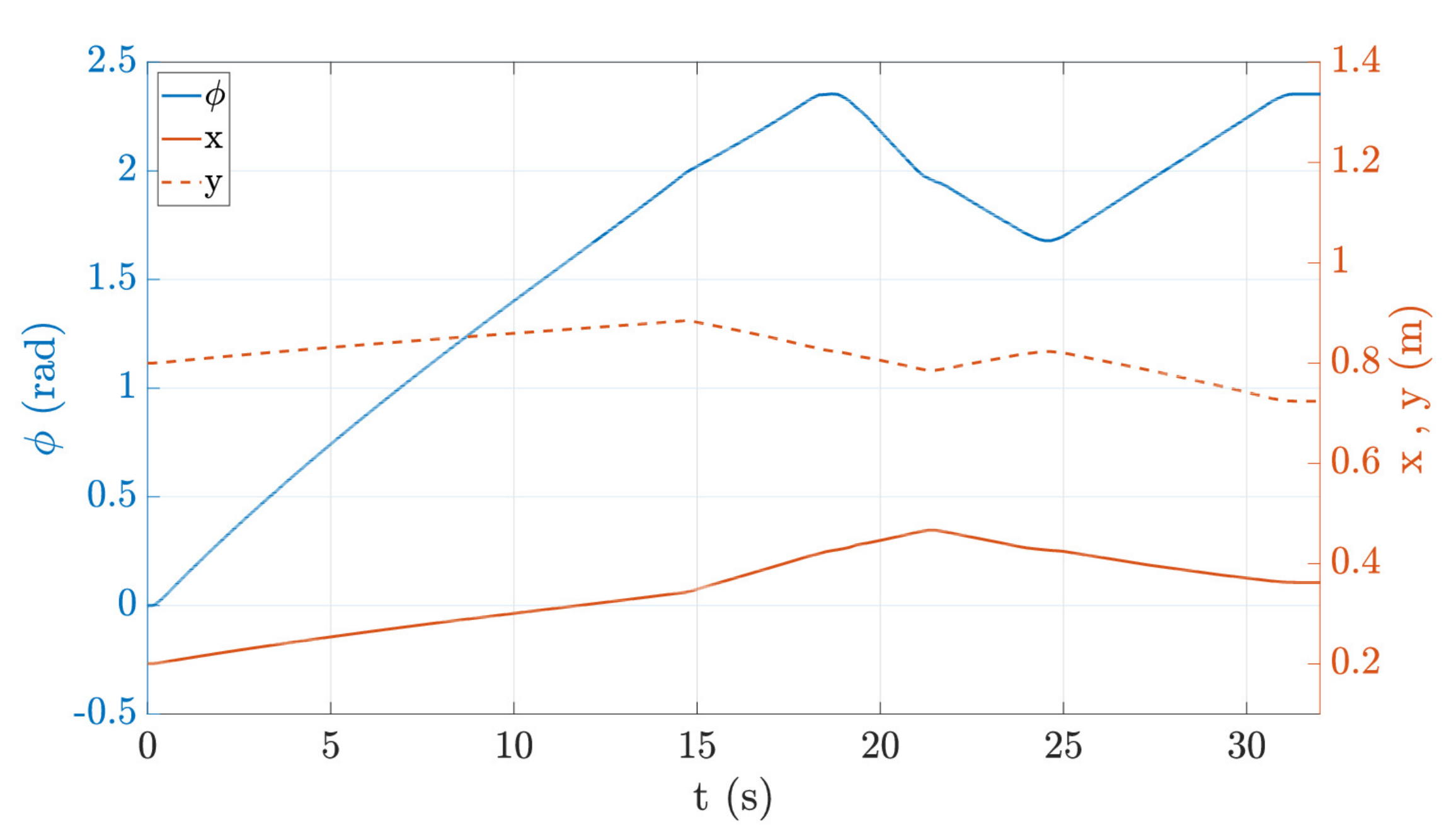}
    \caption{}
\end{subfigure}
\begin{subfigure}{0.37\textwidth}
    \includegraphics[trim={0cm 0cm 0cm 0cm},width=\textwidth,clip]{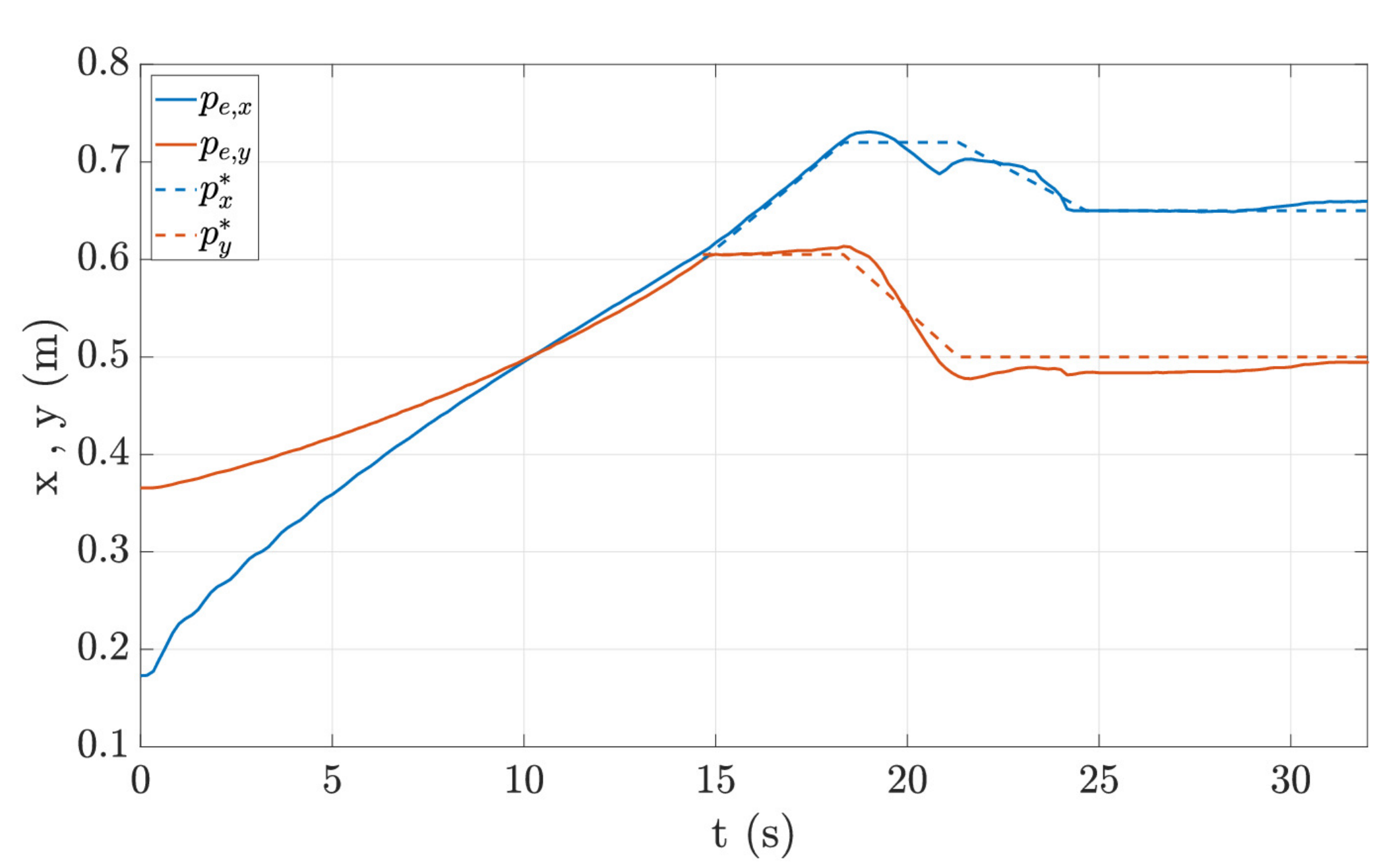}
    \caption{}
\end{subfigure}

\caption{(a) The complete series of object configurations generated for the hooking task. (b) The measured trajectory of the object base during the experiment. Note that until $t=15s$ the manipulator is moving to the first generated trajectory point in the series. (c) The measured trajectory of the object endpoint coordinates, as well as the reference endpoint goal trajectory.}
\label{fig:hook_demo_plots}
\end{figure*}

Solving the problem for $\psi_{B'}^* \in [\frac{-\pi}{2},\frac{\pi}{2}]$ at $\frac{\pi}{12}$ rad increments and $p^* = (0,0.4)$ produced the configurations shown in the plots on the left of Fig. \ref{fig:black_orientation_solutions_and_composite}a, with composites of the real object state at each of these configurations displayed on the right.

To quantitatively assess the accuracy, the angle of the tip was determined by measuring the points at the start and end of the endpoint weight to obtain a direction vector in the $xy$ plane. The error between this and the goal angle is plotted in Fig. \ref{fig:black_orientation_solutions_and_composite}b. The position errors are also plotted in Fig. \ref{fig:black_orientation_solutions_and_composite}c. This investigation also reveals considerable ability to reach specified orientations in the range $\frac{-\pi}{2} \leq \psi_{B'}^* \leq \frac{\pi}{2}$, with a recorded maximum absolute angle error of around 0.29 rad or 17\textdegree. Particularly in the range of $0 \leq \psi_{B'}^* \leq \frac{\pi}{2}$ the error is maintained below 0.12 rad (7\textdegree) and 0.21 rad (12\textdegree) for OB4 and OB6 respectively. The position errors are also small, with most cases being below $10 \%$ of the object's length.

\subsection{Task Demonstration}\label{subsec:task_demo}

As a further demonstration using the controller in a task oriented manner, we provide an example where the robot needs to hang the end of the object over a hook which lies outside the reachable workspace of both the robot end effector, and the object endpoint when in the reference vertical configuration. Five main goal locations for the endpoint were defined to accomplish this task, with additional intermediate waypoints interpolated in between them. The first four locations involve varying the position of the object endpoint while keeping its orientation constant. The final location maintains a fixed position while rotating the end effector by $\pi/4$ radians.  We then use the optimisation procedure for this series of goals to generate a trajectory for the object base pose, and execute this with the robot manipulator. 

The upper row of Fig. \ref{fig:hook_demo} shows the generated configurations at the main goals, as well as the intermediate points, and images captured at these locations during the experiment are shown below, with the modeled object state and manipulator endpoint constraint overlaid (a video of the complete experiment is shown in the multimedia attachment). The complete set of configurations are shown together in Fig. \ref{fig:hook_demo_plots}a, and the measured trajectory of the object base and the x- and y-coordinates of the object endpoint along with its intended goal path are plotted in Fig. \ref{fig:hook_demo_plots}b and Fig. \ref{fig:hook_demo_plots}c, respectively.

\section{Conclusion}\label{VI-conclusion}

We presented a strain-based modeling approach for manipulating deformable linear objects, using a polynomial curvature parametrization developed within the context of soft robotics to enable compact, model-based control. The manipulation task was framed as a control problem, and controllers with provable guarantees were introduced. We focused in this work on planar bending (i.e., curvature only), but additional strains such as torsion can be added to demonstrate 3D manipulation.
Results in simulations show that choosing a higher-order polynomial to describe the system’s curvature improves accuracy but comes at the cost of increased computational time when solving the optimization problems. Thus, for this paper, we opted for the linear curvature model as a compromise between accuracy and simplicity when it came to experimental validation.
Experiments showed that even minimal models can achieve steady-state accuracy across various object properties, outperforming model-free baselines by 57.7–78.5\% in positioning, with maximum orientation errors of 12\textdegree and 17\textdegree. We also demonstrated task-level applications and assessed dynamic model accuracy.
Future work will focus on validating experimentally higher-order strain models, non-collocated feedback, dynamic motions involving trajectory tracking, and fast actions such as throwing or whipping.

\bibliographystyle{IEEEtran}
\bibliography{main}

\end{document}